\documentclass{article}

\usepackage{microtype}
\usepackage{graphicx}
\usepackage{booktabs} %
\usepackage{multirow}
\usepackage{spverbatim} %
\usepackage{adjustbox}
\usepackage{tabularx}
\usepackage{ragged2e}
\usepackage{array}
\usepackage{hyperref}

\usepackage{url}

\usepackage[accepted]{icml2024}

\usepackage{amsmath}
\usepackage{amssymb}
\usepackage{mathtools}
\usepackage{amsthm}

\usepackage{enumitem}

\usepackage{placeins}

\usepackage[capitalize,noabbrev]{cleveref}

\usepackage[many]{tcolorbox}

\theoremstyle{plain}

\theoremstyle{definition}

\theoremstyle{remark}

\providecommand{\edit}[1]{{{#1}}}
\providecommand{\textedit}[1]{{{#1}}}

\providecommand{\danqi}[1]{
    {\protect\color{purple}{}}
}
\providecommand{\alex}[1]{
    {\protect\color{blue}{}}
}
\providecommand{\aatmik}[1]{
    {\protect\color{orange}{}}
}
\providecommand{\saumya}[1]{
    {\protect\color{orange}{}}
}

\DeclareMathOperator*{\argmax}{arg\,max}

\renewcommand{\paragraph}[1]{\vspace{0.2cm}\noindent\textbf{#1}}

\newcommand{\ourmethod}{QuRating}
\newcommand{\ourdata}{QuRatedPajama}

\newtcbox{\hlprimarytab}{on line, rounded corners, box align=base, colback=green!10,colframe=white,size=fbox,arc=3pt, before upper=\strut, top=-2pt, bottom=-4pt, left=-2pt, right=-2pt, boxrule=0pt}
\newtcbox{\hlsecondarytab}{on line, box align=base, colback=red!10,colframe=white,size=fbox,arc=3pt, before upper=\strut, top=-2pt, bottom=-4pt, left=-2pt, right=-2pt, boxrule=0pt}
\newcommand{\dashifted}{\raisebox{0.5\depth}{\tiny$\downarrow$}}
\newcommand{\uashifted}{\raisebox{0.5\depth}{\tiny$\uparrow$}}

\newcommand{\da}[1]{{\raisebox{0.6ex}{\tiny\hlsecondarytab{\dashifted{#1}}}}}
\newcommand{\ua}[1]{{\raisebox{0.6ex}{\tiny\hlprimarytab{\uashifted{#1}}}}}

\newcommand{\dar}[1]{{\raisebox{0.6ex}{\tiny\hlprimarytab{\dashifted{#1}}}}}
\newcommand{\uar}[1]{{\raisebox{0.6ex}{\tiny\hlsecondarytab{\uashifted{#1}}}}}

\icmltitlerunning{QuRating: Selecting High-Quality Data for Training Language Models}

\begin{document}

\twocolumn[
\icmltitle{\ourmethod{}: Selecting High-Quality Data for Training Language Models}

\begin{icmlauthorlist}
\icmlauthor{Alexander Wettig}{pli}
\icmlauthor{Aatmik Gupta}{pli}
\icmlauthor{Saumya Malik}{pli}
\icmlauthor{Danqi Chen}{pli}
\end{icmlauthorlist}

\icmlaffiliation{pli}{Department of Computer Science \& Princeton Language and Intelligence (PLI), Princeton University}

\icmlcorrespondingauthor{Alexander Wettig}{awettig@cs.princeton.edu}

\icmlkeywords{Language Models, Data Selection}

\vskip 0.3in
]

\printAffiliationsAndNotice{} %
\setcounter{footnote}{1}%
\begin{abstract}

Selecting high-quality pre-training data is important for creating capable language models, but existing methods rely on simple heuristics. We introduce \ourmethod{}, a method for selecting pre-training data that can capture human intuitions about data quality.
In this paper, we investigate four qualities---\textit{writing style}, \textit{required expertise}, \textit{facts \& trivia}, and \textit{educational value}---and find that LLMs are able to discern these qualities, especially when making pairwise judgments of texts.
We train a QuRater model to learn scalar ratings from pairwise judgments, and use it to annotate a 260B training corpus with quality ratings for each of the four criteria.
In our experiments, we select 30B tokens according to the different quality ratings and train 1.3B-parameter language models on the selected data. 
We find that it is important to balance quality and diversity.
When we sample using quality ratings as logits over documents,
our models obtain lower perplexity and stronger in-context learning performance than baselines.
Our best model is based on educational value and performs similarly to a model trained with uniform sampling for 50\% more steps.
Beyond data selection, we use the quality ratings to construct a training curriculum which improves performance without changing the training dataset.
We extensively analyze the quality ratings and discuss their characteristics, biases, and wider implications.%
\footnotemark{}

\end{abstract}

\section{Introduction}

There is increasing evidence that choosing the right training data is essential for producing state-of-the-art large language models (LLMs) \citep{brown2020language, chowdhery2023palm, rae2021scaling}.
Researchers have found that model performance can be improved by deduplicating training data \citep{lee2022deduplicating, abbas2023semdedup}, finding the right balance of domains \citep{touvron2023llama, xie2023doremi}, or selecting data that resembles a high-quality corpus \citep{brown2020language, chowdhery2023palm, xie2023data}. However, the ideal properties of training data remain hard to characterize; an improved understanding could enable open research to train stronger models under resource constraints.

\footnotetext{
To encourage further research, we release our prompts, \mbox{models}, and data at \href{https://github.com/princeton-nlp/QuRating}{https://github.com/princeton-nlp/QuRating}.}

In this work, we aim to capture the abstract qualities of texts which humans intuitively perceive.
As part of our approach, which we call QuRating (\textbf{qu}ality \textbf{rating}),\footnote{QuRating/QuRater can be pronounced like curating/curator.} (1) we compare pairs of texts along a quality criterion, (2) we train a model that translates the resulting judgments into scalar quality ratings, (3) we use these ratings to select pre-training data, and (4) we identify which abstract qualities are valuable by training language models on the selected subsets and evaluating their performance.

\begin{figure}[!t]
    \centering
    \vskip 0.05in
    \centerline{\includegraphics[width=\linewidth]{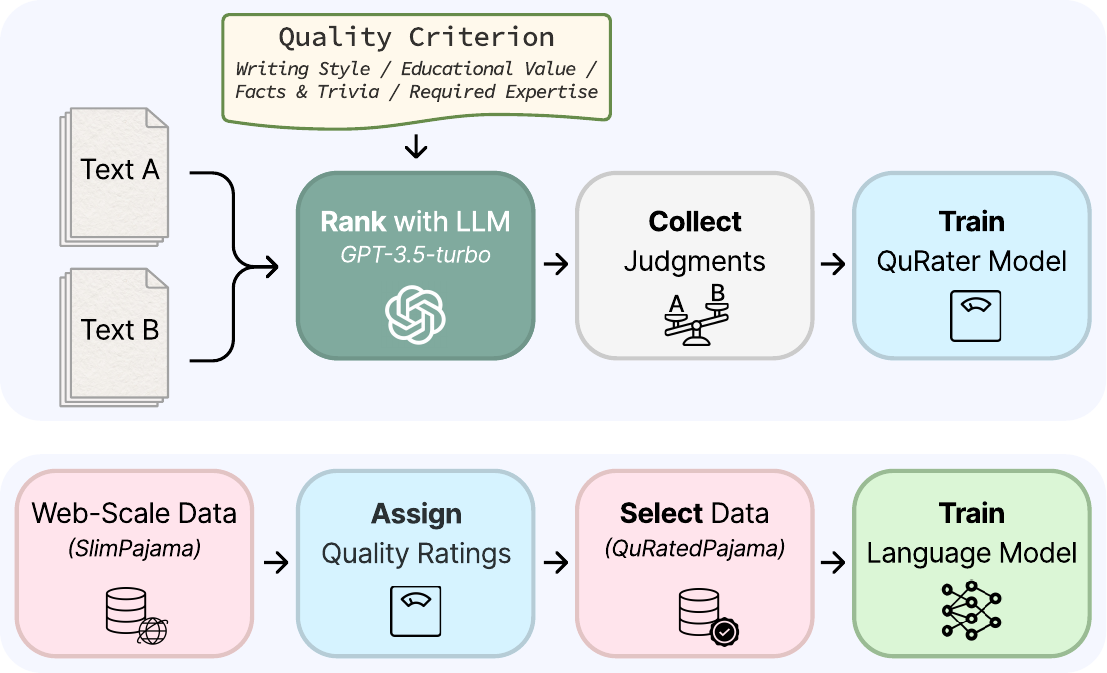}}
    \caption{In QuRating, we obtain comparative judgements from an LLM to train a QuRater model, which assigns scalar quality ratings to the documents in a language model training corpus.}
    \label{fig:overview}
    \vskip -0.15in
\end{figure}

Throughout the paper, we focus on four particular qualities of a text as our criteria for data selection: the text's \textit{writing style}, the amount of \textit{facts \& trivia} it contains, its \textit{educational value}, and the \textit{required expertise} needed to understand it.
These qualities are necessarily subjective. However, we verify that GPT-3.5-turbo can discern their presence in clear-cut cases. In particular, we find that LLMs produce more stable and accurate judgments when prompted to compare documents in a pairwise setting.

We use the Bradley-Terry model \citep{bradley1952rank} to translate the LLM-derived pairwise judgments into quantitative ratings for each piece of text.
We treat the ratings as logits over documents, from which we sample without replacement for subset selection. We add a temperature $\tau$ to trade off quality and diversity by interpolating between top-$k$ selection ($\tau \to 0$) and uniform random sampling ($\tau \to \infty$). We show that in this formulation, quality ratings are connected to rewards in RLHF \citep{ouyang2022training}.

For our experiments, we query GPT-3.5-turbo to judge 250K text pairs for each of the four quality criteria and fine-tune a Sheared-Llama-1.3B model \citep{xia2023sheared} to learn the implied quality ratings.
We then use this fine-tuned \textit{QuRater} model to predict quality ratings across 260B tokens from the SlimPajama corpus \citep{cerebras2023slimpajama} to produce the \textit{QuRatedPajama} dataset.
Finally, we train new 1.3B-parameter language models from scratch by selecting subsets of 30B out of 260B tokens.
We compare our method to uniform sampling, perplexity filtering \citep{wenzek2020ccnet, marion2023more}, and importance resampling \citep{xie2023data} with respect to high-quality domains. Our evaluation focuses on in-context learning on 10 diverse tasks as a measure of model capability \citep{eval-harness}. %

We find that selecting only the highest-rated documents produces models which excel at particular tasks but underperform on others.
Sampling with a temperature $\tau=2.0$ produces more consistent results across tasks and improves validation perplexity.
When we use our best selection criterion, \textit{educational value}, we improve in-context learning (ICL) performance on every tasks by an average of 1.8\% compared to uniform selection, similar to a baseline trained for 50\% more steps. 
Selecting based on \textit{writing style} leads to the best perplexity but surprisingly does not lead to substantial improvements in downstream task performance.

We also leverage the quality ratings to build a training curriculum. Our experiments show that models trained on data ordered based on \textit{required expertise} outperform models which are trained on the same data in a random order.

We perform an extensive analysis of the quality ratings. For each domain, we study the distribution of ratings and report insights from inspecting high and low-ranking documents.
Finally, we discuss the social impact of data selection and document the effect of QuRating on web pages from the AboutMe dataset \citep{lucy2024aboutme}.

Our work demonstrates how certain human notions of data quality are effective signals for scalable data selection.
We release our code, the GPT-3.5-turbo outputs, the fine-tuned QuRater model and the annotated QuRatedPajama dataset to encourage data exploration and efficient LLM training.

\section{Background}
We review some best practices in data engineering for language models and discuss them in relation to our approach.

\paragraph{Rule-based heuristics.} Data selection pipelines commonly include hand-crafted heuristics that filter out low-quality data \citep{rae2021scaling, laurencon2022roots, together2023redpajama, penedo2023the, dolma}. These typically involve thresholds on mean word length, stop word fraction, and word repetitions, e.g., the so-called C4 filters \citep{raffel2020exploring} and the Gopher rules \citep{rae2021scaling}. While binary rules are useful for excluding \textedit{noisy internet artifacts, we need more precise quality measures to identify the most desirable examples in a dataset.}

\paragraph{Model-based heuristics.}
In \textit{heuristic classification} \citep{brown2020language}, a bigram discriminator model selects data that resembles a high-quality target domain, such as Wikipedia articles. This paradigm has been widely adopted \citep{du2022glam, pile, chowdhery2023palm, touvron2023llama}.
In Data Selection with Importance Resampling (DSIR), \citet{xie2023data} sample from generative models instead of discriminators, and show that this improves performance.
We argue that an entire domain such as Wikipedia is an imprecise proxy for data quality.
Another popular method is perplexity filtering \citep{wenzek2020ccnet, muennighoff2023scaling, marion2023more}, i.e., choosing data that has high likelihood under a language model. However, we note that this includes data with simple and repetitive content, which is easy to predict for a model.

\paragraph{LLM quality signals.}
Most similar to our work, \citet{gunasekar2023textbooks} filter data by querying \mbox{GPT-4} to identify documents with ``educational value for a student whose goal is to learn basic coding concepts''.
In contrast to our work, they augment filtered web data with synthetic data generated by LLMs.
We study four quality criteria and use pairwise comparisons, which produce more stable rankings with \mbox{GPT-3.5-turbo}.
Our work is also connected to \citet{korbak2023pretraining}, who incorporate human preferences into language model training but with a focus on toxicity and privacy. We select data with the goal of teaching language models strong skills with fewer samples.

\begin{figure*}[t]
    \vskip 0.1in
    \centering
    \centerline{\includegraphics[width=\linewidth]{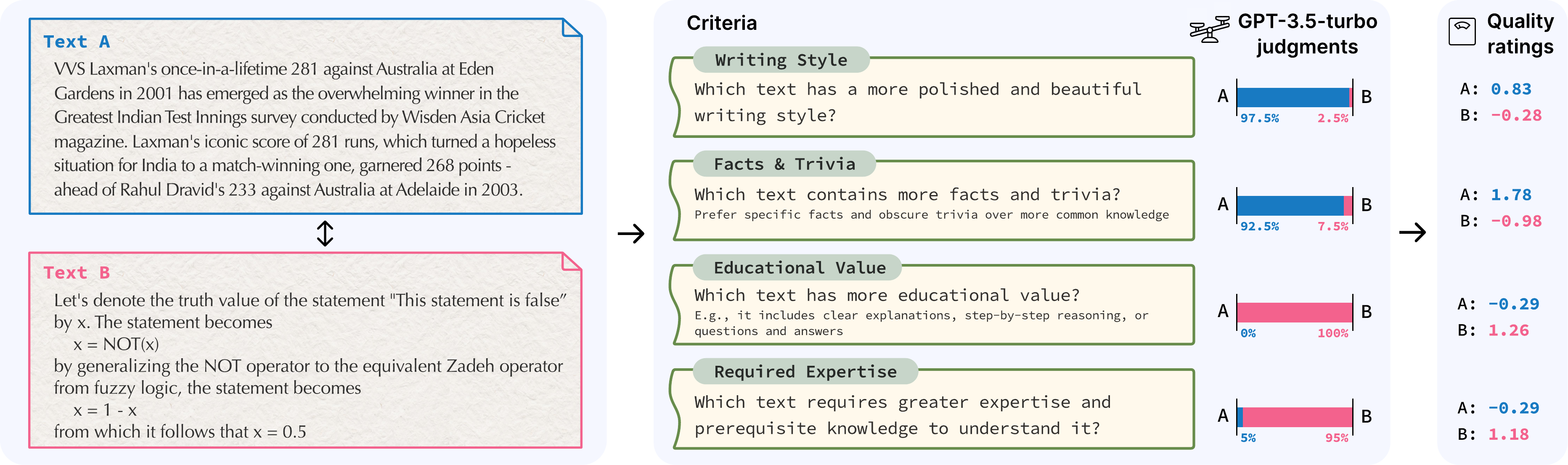}}
    \caption{We consider a pair of texts and report the judgments we elicit from GPT-3.5 as to which text it prefers, using different prompts that correspond to different qualitative criteria. We also report the quality ratings which we ultimately use for data selection. These ratings are given by our QuRater model, which is trained to assign ratings that best correspond to pairwise judgments.}
    \label{fig:prompts}
    \vskip -0.1in
\end{figure*}

\paragraph{Deduplication.}
Deduplicating training data has become standard practice \citep{lee2022deduplicating, anil2023palm2, touvron2023llama}, as it improves language models by removing repeated training data and boosting sample diversity.
Deduplication is usually done in a fuzzy or semantic manner \citep{jiang2023fuzzy, abbas2023semdedup, tirumala2023d4}.
While it reduces the number of documents in a corpus, it is not well-suited to sample a subset of a specific size. Rather, it should be run before selecting data based on quality.

\paragraph{Distillation.}
In knowledge distillation, a student model is trained to absorb the knowledge of a teacher model \cite{hinton2025distilling}.
This approach has been adapted to language models \cite{kim2016sequence, sanh2019distil, agarwal2024onpolicy},
where the student model receives rich feedback from the teacher at every token position.
Selecting data with QuRating can be seen as a much sparser form of distillation, providing as guidance only a single quality signal per sequence. After the training data has been selected, models acquire knowledge not from a teacher but from the raw documents in the training corpus.

\section{Quantifying Qualitative Aspects of Text}
We develop a method which can directly capture human intuition about data quality and leverage it for scalable data selection. We obtain fine-grained quality signals without having to craft heuristic rules or select proxy domains.

\subsection{Overview of the Method}
We associate a \textit{quality criterion} with a question that asks which of two pieces of text $(t_A,\;t_B)$ exhibits a certain quality to a higher degree.
We record the confidence $p_{B \succ A} \in [0, 1]$ with which a judge chooses $B$ over $A$.
In this paper, we sample pairs of texts from a vast collection of documents and use GPT-3.5-turbo to judge them based on the criterion.
Thus, we can rapidly create large datasets of judgments $\mathcal{J} = \{(t_i, t_j, p_{i \succ j})\}$.

We use the Bradley-Terry model \citep{bradley1952rank} to translate binary judgments into scalar quality ratings,
$$p_{B \succ A} = \sigma \left( s_B - s_A \right),$$
where $\sigma$ is the sigmoid function, and the ratings are estimated using maximum-likelihood estimation.
We parametrize the ratings with a so-called \textit{QuRater} model $s_\theta( t_i )$, which is trained with the binary cross-entropy loss,
\begin{align*}
\mathcal{L}_\theta = & \mathop{\mathbb{E}}_{(t_A, t_B, p_{B \succ A}) \in \mathcal{J}} \bigg[ -p_{B \succ A} \log \sigma \left(s_\theta(t_B) - s_\theta(t_A\right)) \\
    & \quad - \left(1-p_{B \succ A}\right) \log \sigma \left(s_\theta(t_A) - s_\theta(t_B\right))\bigg].
\end{align*}

This parallels the training of reward models in Reinforcement Learning from Human Feedback (RLHF) \citep{ouyang2022training} without conditioning the models on user input.

\subsection{Choice of Criteria and Prompts} \label{sec:prompts}
Which abstract qualities of text are most deserving of study?
We argue that the most interesting criteria (1) are applicable to a wide variety of text, (2) require a deeper understanding of a text's content, which cannot easily be derived from surface features, (3) result in fine-grained rankings with few ties, (4) are complementary to each other.
After exploring different candidates and iterating on their prompts, we choose the following four questions as our criteria:

\begin{itemize}[topsep=0pt,parsep=2pt,partopsep=2pt,leftmargin=1em]
    \item \textit{Which text has a more polished and beautiful \textbf{writing style}?}  We use ``style'' to emphasize the writing over the subject matter. In our exploration, ``beautiful'' and ``polished'' favor literary and academic writing, respectively.

    \item \textit{Which text contains more \textbf{facts and trivia}? Prefer specific facts and obscure trivia over more common knowledge.}
    Inspired by LLMs' potential applications as knowledge bases \citep{petroni2019language},
    we aim to identify texts that have a high density of long-tail factual knowledge.
    We find that adding ``trivia'' helps the LLM choose facts about niche topics and fictional worlds.

    \item \textit{Which text has more \textbf{educational value}? E.g., it includes clear explanations, step-by-step reasoning, or questions and answers.}
    Following \citet{gunasekar2023textbooks}, we prompt for educational value and specify some properties that are considered particularly valuable for inducing reasoning capabilities in LLMs, e.g., ``step-by-step reasoning'' resembles Chain-of-Thought prompting \citep{wei2022chain, kojima2022large}.

    \item \textit{Which text \textbf{requires greater expertise} and prerequisite knowledge to understand it?}
    It is interesting to study how the difficulty level of the training corpus impacts the capabilities of the model.
\end{itemize}

We show how these prompts act on two contrasting texts in Figure \ref{fig:prompts}. The full prompts can be found in \cref{app:detailed_prompts}.

\paragraph{Prompt validation.}
As quality judgments are ultimately subjective, we evaluate the prompts on clear-cut cases.
For each criterion, we curate two sets of 40 documents from the web with clear differences in quality.
We describe the specific constitution of this dataset in Table \ref{tab:validation_set_sources} in the appendix.
We choose GPT-3.5-turbo\footnote{Throughout the paper, we use \texttt{GPT-3.5-turbo-0613}.} as the LLM judge and evaluate the agreement with our preferences on 1600 document pairs.
Using our prompts, GPT-3.5-turbo agrees with our preferences over 97\% of the time on all the criteria except \textit{facts \& trivia}, on which we achieve 92\% agreement.

\subsection{Why Use Pairwise Comparisons?}\label{sec:why_pairwise}
Previous work explores using LLMs to judge individual texts \citep{gunasekar2023textbooks}.
We observe that LLMs are better at comparing texts than they are at judging individual texts; specifically, they produce more reliable judgments in pairwise settings, and can better discriminate between texts with fine variations in quality.

In a case study, we rank 10 documents (see Table \ref{tab:ten_wriqual_texts} in Appendix \ref{app:detailed_prompts}) based on the authors' collective perception of their \textit{writing style}. We use this ranking to study the LLM's ability to measure subtle gradations in quality. We ask GPT-3.5-turbo to (1) rate the documents' \textit{writing style} on a 1 to 10 scale or (2) make pairwise judgments for all pairs of documents.
We evaluate the Kendall tau rank coefficient with our human judgments on all 45 pairs and find that with GPT-3.5-turbo achieves $0.79 \pm 0.01$ in the pairwise setting, compared to $0.61 \pm 0.06$ for individual judgments, where we compute standard deviations over 3 runs.

Pairwise judgments are usually easy to verify for human annotators, whereas it is difficult to assess the correctness of the precise grade assigned to a document.
Research in psychology and education suggests that pairwise judgments improve self-consistency and inter-annotator agreement \citep{thurstone1927comparative, pollitt2012method}; and have been found to be useful in the specific setting of evaluating essays \citep{pollitt2004could, lesterhuis2022validity}.
In machine learning, pairwise comparisons have been used in assessing LM outputs \citep{ouyang2022training, dubois2023alpacafarm, zeng2023llmbar} and in information retrieval \citep{gienappsparse, sun-etal-2023-chatgpt, qin2023large}.

\subsection{Training the QuRater Model} \label{sec:data_quality_modeling}
Having settled on criteria and prompts, we produce a large-scale dataset of judgments by querying GPT-3.5-turbo on 250K text pairs for each criterion.
For each text pair, we prompt the LLM in both $(t_A, t_B)$ and $(t_B, t_A)$ order, and generate multiple continuations, which we average to compute the overall confidence $p_{B \succ A}$. This counteracts the positional bias observed in pairwise comparisons with LLMs \citep{wang2023large, zeng2023llmbar}.

The dataset is derived from 500K unique documents in SlimPajama \citep{cerebras2023slimpajama}, a web-scale pre-training corpus based on RedPajama \citep{together2023redpajama}. For each pair of documents, we extract snippets of at most 512 Llama tokens \citep{touvron2023llama}. Specifically, 200K pairs are sampled randomly across all domains, and an additional 10K pairs are sampled within each of the five specialist domains Wikipedia, Book, StackExchange, Github, ArXiv. In \cref{app:data_quality_modeling}, Table~\ref{tab:gpt_statistics} shows that we obtain many confident LLM judgments across domains, and Figure~\ref{fig:gpt_correlations} shows that the judgments are not strongly correlated between criteria.

We fine-tune a 1.3B parameter Sheared-Llama model \citep{xia2023sheared} on the dataset of pairwise LLM judgments. We add four linear heads to predict quality ratings across the four criteria. In \cref{app:data_quality_modeling}, we discuss the training setup in detail and show that the QuRater model has over 93\% accuracy on held-out judgments.

\section{Selecting Data By Quality Rating}
\label{sec:sampling_method}
Our goal is to select a subset of documents from a large-scale corpus. We introduce the following framework for sampling according to quality ratings. Let each document $d_i$ in a corpus $\mathcal{D}$ be annotated with a quality rating $s_i$, assigned by the QuRater model.
We sample documents without replacement according to the softmax probabilities,
$$p\left( d_i \right) \propto \exp \left( \frac{s_i}{\tau} \right),$$
normalized over the corpus $\mathcal{D}$.
The temperature term $\tau$ controls the sample diversity.
As $\tau \to 0$, this strategy becomes top-$k$ selection, the most straightforward approach of incorporating quality signals. At $\tau \to \infty$, it is equivalent to the uniform sampling baseline.
This sampling scheme implicitly changes the language modeling objective to reward-weighted regression \citep{korbak2023pretraining, peters2007reinforcement}.
We sample without replacement following \cite{xie2023data}, as it increases sample diversity and also allows for efficient sampling via the Gumbel top-$k$ trick \citep{kim2016exact, kool2019stochastic, vieira2014gumbel}.

\paragraph{Quality ratings as rewards.}
In \cref{app:rlhf}, we show the quality ratings are connected to rewards in Reinforcment Learning from Human Feedback (RLHF) \citep{ouyang2022training, rafailov2023direct} when using this sampling strategy.
For large enough sample sizes, our method approximates
pre-training a language model on a random subset of the dataset, and then using RLHF to steer the language model towards generating documents with higher quality ratings.
The temperature $\tau$ becomes the weight of the KL-divergence term in the RLHF objective, constraining the RLHF model to be similar to the pre-trained model.
Unlike the typical rewards used in RLHF, the quality ratings are not conditioned on user input and should serve a different purpose: to guide the model towards data from which it can learn generalizable skills \citep{arora2023theory, yu2023skill} and useful world knowledge \citep{li2023textbooks}.

\paragraph{Curriculum learning with quality ratings.}
Sampling without replacement naturally leads to a training curriculum.
In regular language model training, the examples are randomly shuffled after data selection.
However, if we train on examples in reverse order in which they were sampled,
examples with low quality ratings are more likely to be appear at the start of training and highly-rated examples towards the end of training. This is particularly interesting when there is sufficient budget to train on the entire corpus $\mathcal{D}$.

\section{Experiments}
We verify the QuRating approach in practice by training language models from scratch.

\subsection{Setup}

\paragraph{\ourdata{}.} We annotate a 260B token corpus with quality ratings across our four criteria---using the QuRater model from \cref{sec:data_quality_modeling}---to produce \ourdata{}.
The corpus is a subset of documents from SlimPajama~\citep{cerebras2023slimpajama}, an extensively deduplicated version of RedPajama~\citep{together2023redpajama}, and consists of sequences of 1024 tokens using the Llama tokenizer \citep{touvron2023llama}.
Since the QuRater model was only fine-tuned on short sequences, we compute the document-level quality rating by averaging over contiguous segments of up to 512 tokens weighted by their segment length. While the annotation process is expensive (equivalent to 520 NVIDIA H100 hours), it can be massively parallelized, and the resulting quality ratings can serve many purposes, e.g., data selection, curriculum training or data discovery.\footnote{We publicly release QuRatedPajama and make it available at \href{https://huggingface.co/datasets/princeton-nlp/QuRatedPajama-260B}{huggingface.co/datasets/princeton-nlp/QuRatedPajama-260B}.}

\paragraph{Training.}
Using different data selection methods, we select a subset of 30B tokens from \ourdata{} and
train a randomly initialized language model on this training set for one epoch in a randomly shuffled order.
The models have 1.3B parameters and use a transformer architecture \citep{vaswani} with RoPE embeddings \citep{su2024roformer}. Further details can be found in \cref{app:training_details}.
We train on slightly more data than the compute-optimal amount \citep{hoffmann2022training}. However, a larger number of training tokens should give a clearer signal regarding data quality.

\begin{table*}[t]
\centering
\vskip -0.05in
\caption{\ourmethod{} improves perplexity and average few-shot in-context learning (ICL) results when sampling with temperature $\tau = 2.0$.  We report validation perplexity and in-context learning task performance for 10 tasks. We highlight the best result in each column and improvement over uniform sampling with the same token budget. In \cref{app:training_details}, we report perplexity numbers for all models in Table~\ref{tab:ppl_results_domains} and detailed results for each ICL task in Table~\ref{tab:icl_results_full}.}
\label{tab:icl_results}
\vskip 0.05in
\resizebox{0.85\textwidth}{!}{
\begin{tabular}{ll@{\hspace{0.00    \textwidth}}>{\raggedleft}p{0.13\textwidth}@{\hspace{0.06\textwidth}}>{\centering}p{0.14\textwidth}@{\hspace{0.00\textwidth}}>{\centering}p{0.14\textwidth}@{\hspace{0.00\textwidth}}>{\centering}p{0.14\textwidth}@{\hspace{0.025\textwidth}}>{\centering\arraybackslash}p{0.10\textwidth}}
\toprule
 & &  & \textbf{Reading}  &  \textbf{Commonsense}  &  \textbf{World}  & \\
&&& \textbf{Comprehension}  &  \textbf{Reasoning}  &  \textbf{Knowledge}  &  \textbf{Average} \\
\multicolumn{2}{l}{\textbf{Selection Method}} & \textbf{Perplexity}  & \textit{(5 tasks)} & \textit{(3 tasks)} & \textit{(2 tasks)} & \textit{(10 tasks)} \\
\midrule
Uniform &  & 8.96\;\;\;\;\;\! & 50.9  & 55.0  & 14.9  & 44.9  \\
\midrule
DSIR & \textit{with Wiki} & 10.67 \uar{1.71} & 50.1 \da{0.8} & 49.8 \da{5.2} & 14.7 \da{0.2} & 42.9 \da{2.0} \\
 & \textit{with Book} & 11.00 \uar{2.04} & 47.9 \da{3.0} & \textbf{56.6 \ua{1.6}} & 14.1 \da{0.8} & 43.8 \da{1.1} \\
\cmidrule(lr){1-2}
Perplexity & \textit{lowest} & 11.92 \uar{2.96} & 48.3 \da{2.6} & 49.6 \da{5.4} & 13.7 \da{1.2} & 41.7 \da{3.2} \\ 
 & \textit{highest} & 9.97 \uar{1.01} & 49.6 \da{1.3} & 53.5 \da{1.5} & 13.4 \da{1.5} & 43.5 \da{1.4} \\
\cmidrule(lr){1-7}
Writing & \textit{top-$k$} & 10.53 \uar{1.57} & 49.3 \da{1.6} & 53.3 \da{1.7} & 13.5 \da{1.4} & 43.4 \da{1.5} \\
Style & $\tau=2.0$ & \textbf{8.90 \dar{0.06}} & 51.0 \ua{0.1} & 55.8 \ua{0.8} & 14.1 \da{0.8} & 45.0 \ua{0.1} \\
\cmidrule(lr){1-2}
Facts \& & \textit{top-$k$} & 10.56 \uar{1.60}&  54.3 \ua{3.4} & 51.7 \da{3.3} & 15.5 \ua{0.6} & 45.8 \ua{0.9} \\
Trivia & $\tau=2.0$ & 8.91 \dar{0.05} & 52.7 \ua{1.8} & 55.6 \ua{0.6} & 15.6 \ua{0.7} & 46.2 \ua{1.3} \\
\cmidrule(lr){1-2}
Educational & \textit{top-$k$} & 10.59 \uar{1.63} & \textbf{54.7 \ua{3.8}} & 54.9 \da{0.1} & 14.4 \da{0.5} & \textbf{46.7 \ua{1.8}} \\
Value & $\tau=2.0$ & 8.91 \dar{0.05} &  53.3 \ua{2.4} & 56.3 \ua{1.3} & \textbf{15.7 \ua{0.8}} & \textbf{46.7 \ua{1.8}} \\
\cmidrule(lr){1-2}
Required & \textit{top-$k$} & 11.54 \uar{2.58} & 52.8 \ua{1.9} & 48.7 \da{6.3} & 14.3 \da{0.6} & 43.9 \da{1.0} \\
Expertise & $\tau=2.0$ & 8.93 \dar{0.03} & 52.7 \ua{1.8} & 55.5 \ua{0.5} & 15.0 \ua{0.1} & 46.0 \ua{1.1} \\
\cmidrule(lr){1-2}
Criteria mix & $\tau=2.0$ &  \textbf{8.90 \dar{0.06}} & 52.1 \ua{1.2} & 55.5 \ua{0.5} & 15.2 \ua{0.3} & 45.7 \ua{0.8} \\
\midrule
\multicolumn{2}{l}{\textit{Uniform\;\;+50\% data}} & \textit{8.46 \dar{0.50}} & \textit{52.9 \ua{2.0}} & \textit{57.0 \ua{2.0}} & \textit{15.9 \ua{1.0}} & \textit{46.8 \ua{1.9}} \\
\bottomrule
\end{tabular}
}
\vskip -0.05in
\end{table*}

\paragraph{Evaluation.}
We aim to provide a holistic evaluation of the language models trained on 30B tokens:
\begin{itemize}[topsep=0pt,parsep=0pt,partopsep=0pt,leftmargin=1em]
    \item We measure the perplexity over 50M tokens from SlimPajama's held-out validation split.

    \item  We evaluate the in-context learning (ICL) performance using \texttt{lm-evaluation-harness} \citep{eval-harness}.
    We study 10 tasks, comprising 5 reading comprehension tasks (ARC-easy/challenge \citep{clark2018think}, SciQA \citep{welbl2017crowdsourcing}, LogiQA \citep{liu2020logiqa}, BoolQ \citep{clark2019boolq}), 3 commonsense reasoning tasks (HellaSwag \citep{zellers2019hellaswag}, PIQA \citep{bisk2020piqa}, WinoGrande \citep{sakaguchi2021winogrande}) and 2 knowledge-intensive tasks (NQ \citep{kwiatkowski2019natural}, MMLU \citep{hendrycks2021measuring}). We choose the number of few-shot examples for each task to ensure that all examples fit within the context window of 1024 tokens. We report the detailed settings in \cref{app:training_details}.

    \item We evaluate the instruction-following capabilities of our models, borrowing the setting used by \cite{xia2023sheared}. We perform supervised fine-tuning on 10,000 instruction-response pairs from the ShareGPT dataset. We evaluate on another 1,000 instructions and use the AlpacaFarm codebase \citep{dubois2023alpacafarm} to judge the responses from two models with \texttt{GPT-4-0314}.
\end{itemize}

\subsection{Data Selection Methods}
In each experiment, we select a 30B-token training dataset from \ourdata{} with one of the following methods, while retaining the same domain proportions as the overall dataset.
We leave it to future work to combine QuRating with methods that optimize the domain mixture.

\begin{itemize}[topsep=0pt,parsep=1pt,partopsep=1pt,leftmargin=1em]

\item \textit{Uniform}: We select randomly with a uniform probability across documents, equivalent to $\tau \to \infty$. For comparison's sake, we train an additional model on 45B tokens, requiring 50\% more compute.

\item \textit{Sample with \ourmethod{}}:
For each of the four criteria, we sample according to the quality ratings as described in \cref{sec:sampling_method}. We normalize the variance of the quality ratings to be $1$ and then sample with temperatures $\tau \in \{0.0 \text{ (i.e., top-$k$ selection)}, 1.0, 2.0\}$.

\item \textit{Inverse sampling}: As a control study, we repeat the above procedure with transformed quality ratings $s_i \to -s_i$ to select the lowest-rated documents.

\item \textit{Criteria mix}: We explore the setting of merging the \ourmethod{}-sampled data for $\tau=2.0$ for the four criteria, and subsampling it randomly to 30B tokens, while taking care to exclude duplicate documents.

\item \textit{DSIR}: We apply data selection with importance resampling (DSIR) \citep{xie2023data} and select examples that resemble either English Wikipedia or the Book domain \citep{together2023redpajama}---commonly used as proxies for quality \citep{brown2020language, touvron2023llama, xie2023data}. We follow \cite{xie2023data} and train hashed bigram models on \ourdata{} and the target data.

\item \textit{Perplexity Filtering}: We implement perplexity filtering \citep{wenzek2020ccnet, marion2023more} and select the documents with the lowest/highest perplexity scores, as computed by a pre-trained ShearedLlama-2.7B model \citep{xia2023sheared}---2$\times$ the size of our QuRater model.

\end{itemize}

\subsection{Results}
\begin{figure}[t]
    \centering
    \vskip 0.1in
    \centerline{\includegraphics[width=1.0\linewidth]{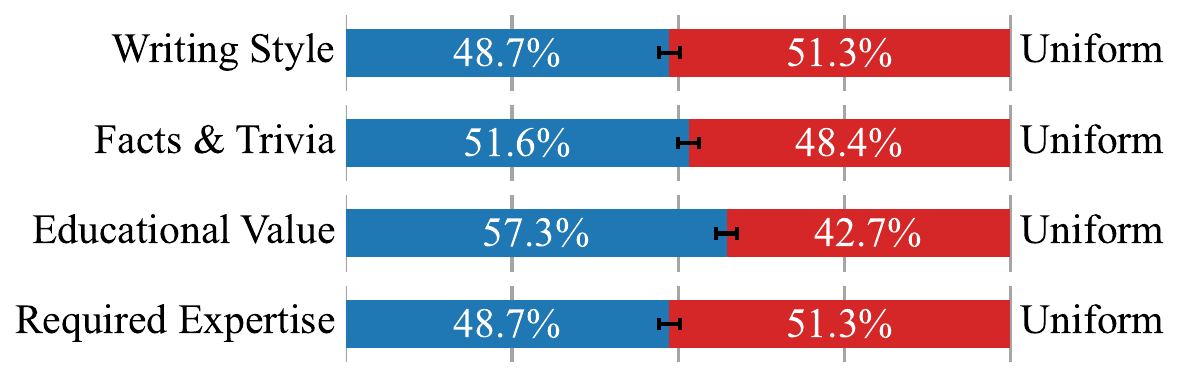}}
    \caption{Instruction following win rates of models trained with QuRating ($\tau = 2.0$) vs. uniform data selection after instruction fine-tuning on 10K ShareGPT examples.}
    \vskip -0.1in
    \label{fig:instruction_ft_winrates}
\end{figure}

We report perplexities and ICL results in of the models in Table~\ref{tab:icl_results} and the instruction following evaluation in Figure~\ref{fig:instruction_ft_winrates}. In \cref{app:training_details}, we provide comprehensive results for all models, namely, perplexity evaluation across domains in Table~\ref{tab:ppl_results_domains} and ICL results for individual tasks in Table~\ref{tab:icl_results_full}.

\paragraph{Baselines underperform uniform selection.}
Surprisingly, DSIR \citep{xie2023data} and perplexity filtering perform worse than random uniform sampling in our experiments. The perplexity evaluation in Table~\ref{tab:icl_results} suggests that these method introduce substantial bias to the training data, and we observe that this does not translate to better ICL results.

\paragraph{Sampling is better than top-$k$ selection.}
Selecting only the top-$k$ documents for training a language model results in substantially worse perplexity, suggesting that the best-rated documents do not have good coverage of the overall text distribution.
Our sampling strategy is effective at alleviating this, 
and increasing sample diversity further with a temperature of $\tau=2.0$ improves perplexity over uniform sampling, despite the shift between the train and test distribution.
In terms of in-context learning performance, top-$k$ selection achieves strong performance gains on individual tasks, but there is always a task where it performs worse than uniform selection. In contrast, sampling results in more balanced results across tasks, leading to better or equal average performance compared to top-$k$ selection.

\paragraph{Perplexity does not inform ICL performance.}
When varying the method for selecting training data, we cannot rely on perplexity as a proxy for model capabilities.
For example, the \textit{writing style} criterion yields the lowest perplexity, but surprisingly, only minor improvements in ICL performance.
In Figure~\ref{fig:icl_vs_ppl} in the appendix, we visualize the relationship across all tasks and models and observe no clear trends.

\paragraph{Educational value is the strongest criterion.}
Amongst our criteria, the models trained on texts with \textit{educational value} exhibit the strongest gains on ICL, and with \mbox{$\tau=2.0$} improve upon uniform sampling in all of the 10 tasks.
This model performs comparably to a uniform sampling baseline trained with +50\% more data and compute, highlighting the impact of selecting high-quality data.
It is also the only model that gives a clear win rate of 57.3\% against the uniform model after instruction tuning in Figure~\ref{fig:instruction_ft_winrates}.

\paragraph{Other criteria.}
We find that \textit{Facts \& trivia} and \textit{required expertise} improve ICL performance on average, and provide promising gains in reading comprehension and world knowledge tasks, but perform worse at commonsense reasoning.
In our control experiments, in which we sample from the lowest quality ratings, no criterion meaningfully improves overall ICL performance.
However, selecting documents low in \textit{facts \& trivia} or \textit{required expertise} benefits all 3 commonsense tasks, see Table~\ref{tab:icl_results_full} in the appendix.

\paragraph{Criteria mix.}
Mixing the selected subsets across criteria results in low perplexity and better average ICL performance than baselines, it does not outperform selection using only \textit{educational value} or \textit{facts \& trivia}. This may be since 25\% of the data is selected based on \textit{writing style}, our worst performing criterion.
We are optimistic that future work may find more effective ways of combining criteria.

\subsection{Curriculum Learning}
\paragraph{Setting.} We train two additional models on the 30B token dataset created with uniform selection. However, we change the order in which samples are seen during training based on \textit{required expertise}. Specifically, we use our sampling method (\cref{sec:sampling_method}) with temperature $\tau=2.0$, and train on data in the same order in which it was sampled (high $\to$ low expertise), or in the reverse order (low $\to$ high expertise). This explores whether quality ratings are useful in forming a curriculum without changing the set of training examples.

\paragraph{Results.}
The evaluation results for perplexity and ICL are included in the appendix in Tables~\ref{tab:ppl_results_domains} and \ref{tab:icl_results_full} respectively.
We find that, even when training on the same set of examples, quality ratings are still useful for improving performance, compared to training with a randomly permuted sample order.
Both the curriculum of low-to-high expertise and its reverse improve average ICL performance by 0.6\% and 0.5\% respectively, with strong performance in different tasks. 
However, only the curriculum of increasing expertise improves perplexity on held-out data.

\begin{figure*}[t]
    \centering
    \vskip 0.05in
    \centerline{\includegraphics[width=\linewidth]{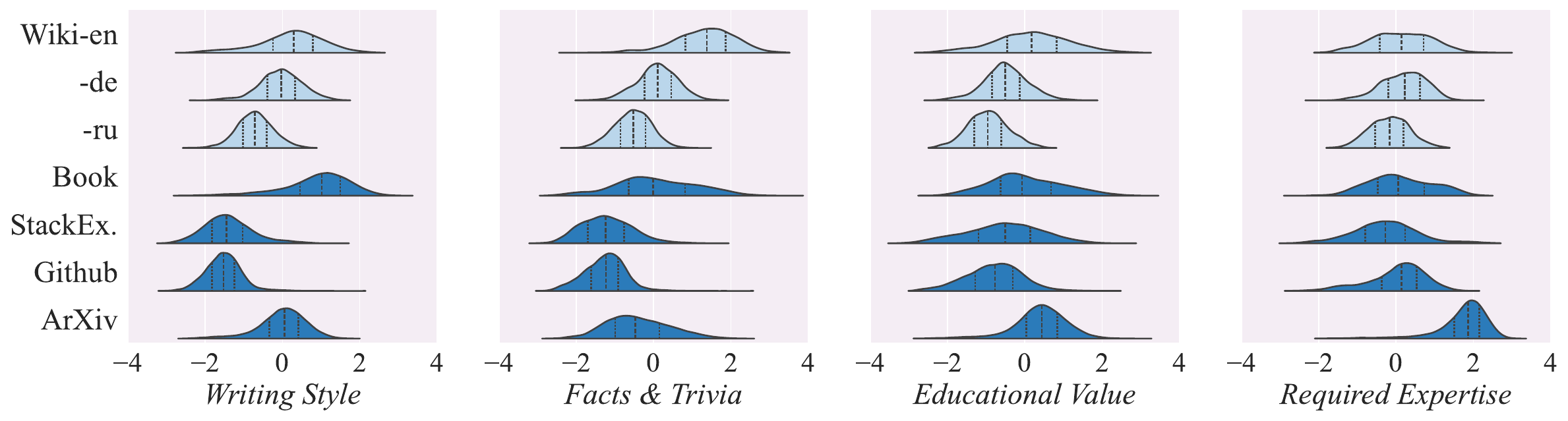}}
    \vskip -0.1in
    \caption{Distribution of quality ratings, normalized for each criterion to have zero mean and unit standard deviation across the corpus.}
    \label{fig:scores_domains}
\end{figure*}

\section{Analysis of Quality Ratings} \label{sec:analysis}
Understanding the nature of quality ratings on such a vast amount and variety of data is challenging. We begin by studying the distribution of ratings across domains, as well as across unsupervised clusters when domains are too coarse.
We then inspect raw documents across these distributions and discuss our observations. Lastly, we document the social, topical, and geographical biases of our approach by applying it to the AboutMe dataset \citep{lucy2024aboutme}.

\subsection{Distribution of Quality Ratings} \label{sec:distribution}
We sample 1M sequences from \ourdata{} and visualize the normalized quality ratings across the different RedPajama domains \citep{together2023redpajama}.
CommonCrawl and C4 constitute the majority of training data, but lack interpretable metadata. We therefore leverage techniques in unsupervised domain discovery.
We follow \citet{gururangan2023scaling} and implement $k$-Means clustering with $k=25$. We name the clusters by the most salient TF-IDF terms at the cluster centroids.

In Figure~\ref{fig:scores_domains}, we plot the distribution of quality ratings across Wikipedia, Book, StackExchange, Github and ArXiv domains.
The results align with expectations: The Book domain has high ratings for \textit{writing style}; a subset of Wikipedia is particularly rich in \textit{facts \& trivia}; ArXiv requires particularly high \textit{expertise}. 
However, each domain contains a wide range of ratings, suggesting that it would be sub-optimal to select data by simply picking domains, e.g., all domains contribute towards the overall top 5\% of documents in terms of \textit{educational value}.

We visualize the quality ratings for Wikipedia documents across different languages in Figure~\ref{fig:scores_domains}, and notice that the quality ratings exhibit a bias towards English.
Since we explicitly instruct GPT-3.5 to ignore language in its judgements in \cref{sec:prompts}, this highlights the need for more sophisticated approaches to de-bias model judgements.

We visualize the distributions for the cluster in CommonCrawl and C4 in Figure~\ref{fig:scores_clusters} in the appendix. They are similarly encouraging, e.g., the clusters associated with \textit{cells, protein, gene} and \textit{energy, climate, species} are rated highly on \textit{required expertise}, \textit{educational value}, and \textit{facts \& trivia}. Meanwhile, the \textit{book, author} cluster tends to obtain high ratings in \textit{writing style}. However, almost all clusters encompass a wide range of quality ratings.

\paragraph{Comparison to perplexity filtering.}
We compare sequence-level log-likelihood scores from Llama-2-7b \citep{touvron2023llama2} with the quality ratings across 1M training sequences and visualize the relationship in Figure~\ref{fig:nll_correlations} in the appendix. We observe that documents with low quality ratings have a wide range of likelihoods, and the Spearman correlation coefficient varies between 0.50 for \textit{writing style} to -0.02 for \textit{required expertise}.
Therefore, QuRating is meaningfully different from selecting texts based on perplexity scores from a strong LLM \citep{marion2023more}.

\subsection{Data Inspection} \label{sec:inspection}
We study raw documents from each of the domains and clusters discussed in \cref{sec:distribution}.
We select training examples at the 5th, 30th, 70th and 95th percentile for each criterion, and feature random extracts in \cref{app:raw_documents} without any cherry-picking.
While this is a minute sliver of the training data, the documents still exhibit clear qualitative differences and we invite the reader to inspect them in the appendix.

\paragraph{Behavior on code.}
Table \ref{fig:examples_Github} shows text extracts at different quality percentiles for each of the four criteria on documents from Github.
Although not designed for code, the quality ratings correlate with reasonable traits. We notice
\textit{writing style} and \textit{facts \& trivia} prefer code with comments and documentation; \textit{required expertise} ranks a CSS stylesheet lowest and embedded system code highest. The document with the 95th percentile \textit{educational value} rating is a markdown explaining Ruby string manipulation in Spanish. We also highlight that in the StackExchange domain (Table \ref{fig:examples_StackExchange}), the lower-ranked documents include convoluted stack traces, logs, XML, HTML, and CSS documents, while higher-ranked documents contain a mix of code and natural language.

\paragraph{Educational shortcut.}
We notice a potential shortcoming in the model's understanding of the \textit{educational value} criterion.
Some highly-rated documents are \textit{about} education-related topics (schools, universities, etc.) but not inherently educational; see Tables~\ref{fig:examples_cluster_13}, \ref{fig:examples_cluster_17} and \ref{fig:examples_cluster_25}. This may be remedied with a better prompt or a stronger judge like GPT-4.

\begin{table*}[t] %
\centering
\caption{We select 10\% of the webpages in the AboutMe dataset by sampling with different quality criteria and temperature $\tau=2.0$. We report the categories that are most/least retained (amplified/suppressed) in the selected data and report their retention rates in \%.}
\vskip 0.05in
\label{tab:cluster_retention} %
\tiny
{
\setlength{\tabcolsep}{0pt} %
\setlength{\cmidrulekern}{6pt} %

\begin{tabularx}{\textwidth}{@{\hspace{3pt}}*{4}{X@{}>{\raggedleft \arraybackslash}p{6pt}@{\hspace{6pt}}X@{}>{\raggedleft \arraybackslash}p{6pt}@{\hspace{6pt}}}@{\hspace{-3pt}}}
\toprule
\multicolumn{4}{c}{\textbf{Writing Style}} & \multicolumn{4}{c}{\textbf{Facts \& Trivia}} & \multicolumn{4}{c}{\textbf{Educational Value}} & \multicolumn{4}{c}{\textbf{Required Expertise}} \\
\cmidrule(l{3pt}r){1-4} \cmidrule(r){5-8} \cmidrule(r){9-12} \cmidrule(r{3pt}){13-16}
\multicolumn{2}{l}{\hspace{3pt}$\uparrow$ \textit{Topics: amplified}} & \multicolumn{2}{l}{$\downarrow$ \textit{Topics: suppressed}} & \multicolumn{2}{l}{$\uparrow$ \textit{Topics: amplified}} & \multicolumn{2}{l}{$\downarrow$ \textit{Topics: suppressed}} & \multicolumn{2}{l}{$\uparrow$ \textit{Topics: amplified}} & \multicolumn{2}{l}{$\downarrow$ \textit{Topics: suppressed}} & \multicolumn{2}{l}{$\uparrow$ \textit{Topics: amplified}} & \multicolumn{2}{l}{$\downarrow$ \textit{Topics: suppressed}} \\
\cmidrule(l{3pt}r){1-2} \cmidrule(r){3-4} \cmidrule(r){5-6} \cmidrule(r){7-8} \cmidrule(r){9-10} \cmidrule(r){11-12} \cmidrule(r){13-14} \cmidrule(r{3pt}){15-16}
art, gallery & 14 & quality, equipment & 7 & research, university & 16 & fashion, women & 7 & research, university & 16 & fashion, women & 6 & research, university & 18 & fashion, women & 7 \\
writing, books & 14 & car, vehicle & 7 & energy, water & 13 & hair, beauty & 7 & students, school & 15 & online, store & 7 & law, legal & 15 & online, store & 7 \\
design, designer & 13 & online, store & 7 & community, local & 12 & online, store & 7 & children, child & 14 & car, vehicle & 7 & solutions, tech. & 14 & event, events & 7 \\
photography & 13 & website, information & 8 & film, production & 12 & event, events & 8 & health, care & 14 & furniture, jewelry & 7 & dr, medical & 14 & food, restaurant & 7 \\
life, yoga & 13 & products, quality & 8 & art, gallery & 12 & car, vehicle & 8 & dr, medical & 14 & event, events & 8 & software, data & 13 & furniture, jewelry & 7 \\
\cmidrule(l{3pt}r){1-2} \cmidrule(r){3-4} \cmidrule(r){5-6} \cmidrule(r){7-8} \cmidrule(r){9-10} \cmidrule(r){11-12} \cmidrule(r){13-14} \cmidrule(r{3pt}){15-16}
\multicolumn{2}{l}{\hspace{3pt}$\uparrow$ \textit{Roles: amplified}} & \multicolumn{2}{l}{$\downarrow$ \textit{Roles: suppressed}} & \multicolumn{2}{l}{$\uparrow$ \textit{Roles: amplified}} & \multicolumn{2}{l}{$\downarrow$ \textit{Roles: suppressed}} & \multicolumn{2}{l}{$\uparrow$ \textit{Roles: amplified}} & \multicolumn{2}{l}{$\downarrow$ \textit{Roles: suppressed}} & \multicolumn{2}{l}{$\uparrow$ \textit{Roles: amplified}} & \multicolumn{2}{l}{$\downarrow$ \textit{Roles: suppressed}} \\
\cmidrule(l{3pt}r){1-2} \cmidrule(r){3-4} \cmidrule(r){5-6} \cmidrule(r){7-8} \cmidrule(r){9-10} \cmidrule(r){11-12} \cmidrule(r){13-14} \cmidrule(r{3pt}){15-16}
art therapist & 17 & mvp & 7 & postdoctoral fellow & 19 & manicurist & 6 & pathologist & 18 & band & 6 & postdoctoral fellow & 22 & mommy & 6 \\
celebrant & 16 & hacker & 8 & research associate & 19 & mummy & 6 & lang. pathologist & 18 & act & 6 & research associate & 21 & crafter & 6 \\
laureate & 16 & youtuber & 8 & research scientist & 18 & mama & 7 & postdoctoral fellow & 17 & bandleader & 7 & research fellow & 2 & momma & 6 \\
travel writer & 16 & breeder & 8 & research fellow & 17 & beauty therapist & 7 & classroom teacher & 17 & dj & 7 & research scientist & 18 & quilter & 6 \\
wedding planner & 16 & system administrator & 9 & geologist & 17 & seamstress & 7 & instruct. designer & 17 & drummer & 7 & associate professor & 18 & florist & 7 \\
\cmidrule(l{3pt}r){1-2} \cmidrule(r){3-4} \cmidrule(r){5-6} \cmidrule(r){7-8} \cmidrule(r){9-10} \cmidrule(r){11-12} \cmidrule(r){13-14} \cmidrule(r{3pt}){15-16}
\multicolumn{2}{l}{\hspace{3pt}$\uparrow$ \textit{Regions: amplified}} & \multicolumn{2}{l}{$\downarrow$ \textit{Regions: suppressed}} & \multicolumn{2}{l}{$\uparrow$ \textit{Regions: amplified}} & \multicolumn{2}{l}{$\downarrow$ \textit{Regions: suppressed}} & \multicolumn{2}{l}{$\uparrow$ \textit{Regions: amplified}} & \multicolumn{2}{l}{$\downarrow$ \textit{Regions: suppressed}} & \multicolumn{2}{l}{$\uparrow$ \textit{Regions: amplified}} & \multicolumn{2}{l}{$\downarrow$ \textit{Regions: suppressed}} \\
\cmidrule(l{3pt}r){1-2} \cmidrule(r){3-4} \cmidrule(r){5-6} \cmidrule(r){7-8} \cmidrule(r){9-10} \cmidrule(r){11-12} \cmidrule(r){13-14} \cmidrule(r{3pt}){15-16}
Southern Europe & 12 & Eastern Asia & 8 & Central Asia & 13 & Southern Asia & 10 & Sub-Sah. Africa & 10 & Eastern Asia & 8 & Eastern Europe & 12 & Pacific Islands & 10 \\
Western Europe & 12 & Southern Asia & 8 & Eastern Europe & 12 & South-East. Asia & 10 & Northern Europe & 10 & Pacific Islands & 9 & Western Europe & 12 & North America & 10 \\
Northern Europe & 11 & Central Asia & 9 & Northern Africa & 11 & Northern Europe & 10 & Australia \& N.Z. & 10 & Southern Asia & 9 & Central Asia & 11 & Australia \& N.Z. & 10 \\
Latin Am. \& Carr. & 11 & South-East. Asia & 9 & Western Europe & 11 & North America & 10 & North America & 10 & South-East. Asia & 9 & Northern Africa & 11 & South-East. Asia & 10 \\
Australia \& N.Z. & 11 & Pacific Islands & 9 & Pacific Islands & 11 & Eastern Asia & 10 & Western Europe & 10 & Southern Europe & 10 & Eastern Asia & 11 & Southern Asia & 10 \\
\bottomrule
\end{tabularx}
\par \vspace{3pt}
\tiny \RaggedRight \textbf{Abbreviations:} tech. = technology $\mid$ lang. pathologist = language pathologist $\mid$ instruct. designer = instructional designer $\mid$ Latin Am. \& Carr. = Latin America \& Carribean $\mid$ Sub-Sah. = Sub-Saharan 
}
\end{table*}

\subsection{Documenting Social Bias} \label{sec:aboutme}
We apply our data selection pipeline to the AboutMe dataset, \citep{lucy2024aboutme} which associates 10M webpages with geographic, topical, and social role metadata. Following a setting by the authors, we sample a 10\% subset using \ourmethod{} with temperature $\tau=2$.
We calculate retention rates by measuring what fraction of the total pages associated with a topic cluster, social role, or geographical region are retained in the 10\% selected subset, where a rate higher or lower than 10\% means that the attribute is \textit{amplified} or \textit{suppressed}, respectively.  We report the most amplified and suppressed topics, roles, and regions in Table~\ref{tab:cluster_retention}.

Compared to prior data selection methods studied by \citet{lucy2024aboutme}, we find that the retention rates for QuRating are slightly more balanced.
However, sampling is important: In Table~\ref{tab:cluster_retention_topk} in the appendix, we show that the retention rates are far exacerbated in all categories when using top-$k$ selection.
Our results share some common trends with prior methods, e.g., topics related to shopping websites---\textit{online, store} and \textit{fashion, women, (brand)}---are among the most suppressed across all quality criteria; \citet{lucy2024aboutme} make a similar observation.

We observe that the most amplified attributes reflect the quality criteria:
Topics selected with high expertise (research, law, technology, medical, software) are indeed widely considered to require specialized knowledge.
Roles associated with research are amplified when we sample based on \textit{required expertise}, \textit{facts \& trivia}, and \textit{educational value}; roles associated with art and writing are amplified if we use the \textit{writing style} ratings.
\textedit{We observe that documents associated with conventionally female roles (\textit{mommy}, \textit{manicurist}, \textit{beauty therapist}, \textit{quilter}) are suppressed across the \textit{facts \& trivia} and \textit{required expertise} methods. The roles of \textit{youtuber} and \textit{hacker} are suppressed when selecting for \textit{writing style}, suggesting a bias against ``internet" language, while documents linked with performing roles (\textit{band}, \textit{act}, \textit{dj}, \textit{drummer}) are suppressed when selecting for \textit{educational value}. }   
The geographical trends are less pronounced, but we observe that selection based on \textit{writing style} exhibits a mild preference towards websites from Europe.

We agree with \citet{lucy2024aboutme} and \citet{dodge2021documenting} that it is important to study the effect of data filtering on social and geographical representation.
The impact on the resulting language models is not well documented yet, but may be understood in terms of representational and allocative harms \cite{barocas2017problem, suresh2021framework}, and potential manifestations include stereotyping \cite{caliskan2017semantics, manzini2019black, tan2019assessing, abid2021persistent}, erasure \cite{dev2021harms}, or simply a lack of performance in relevant tasks not considered in traditional benchmarks.
We note that web-scraped datasets are already immensely skewed in terms of their social and geographical factors, e.g., by ease of internet access \cite{bender2021dangers}, and creating a taxonomy of social factors is difficult \cite{blodgett2020language}.
Given the broad coverage of web-scale data and the wide range of LLM applications, the question of what would constitute a fair pre-training distribution remains important and up for debate.

\section{Conclusion}
Training corpora for state-of-the-art language models are becoming increasingly large,
such that there are concerns that models may run out of data \citep{muennighoff2023scaling}.
However, under resource constraints, selecting data with QuRating is a promising avenue for improving language models.
To facilitate further research, we release the pairwise judgements, the resulting QuRater model, the language model checkpoints and the annotated \ourdata{}.

\paragraph{Limitations.} 
We note several limitations of our work.
QuRating relies on the ability of LLMs to discern text qualities, making it sensitive to biases and limitations of LLMs, and these are still not well understood.
The difference between pairwise judgments and scoring individual texts will also vary across LLMs and prompts.
Large-scale collection of human quality judgments is needed to better evaluate the robustness of automatic annotations. This will also elucidate the extent of subjective judgment in different qualities.
Our paper finds certain social and linguistic biases in the quality ratings, and future work is necessary to study and reduce these biases during data selection, and to investigate the effect on the resulting language models.
Finally, our experiments are at a relative small scale (1.3B parameters) and it is not certain whether results will transfer to larger models. 
We also note that that the best of our four quality criteria may not be optimal.
However, QuRating remains a useful framework for exploring other notions of data quality.

\section*{Impact Statement}
Language models are increasingly applied in real-world scenarios, and their behavior is inextricably linked to their training data.
Data selection may help produce models at lower computational costs, reducing the environmental footprint of model training \cite{strubell2019energy, lacoste2019quantifying, patterson2021carbon}, and allowing organizations with relatively fewer resources to train stronger models.
Our experiments are still expensive to reproduce, as each training run takes an equivalent of 200 NVIDIA H100 GPU hours.
By studying a set of intuitive qualities as the basis for data selection, our work also sheds light on the relationship between pre-training data and model capabilities.

A number of harmful behaviors of language models trained on large-scale web data are well documented, including exhibiting social biases \cite{nadeem2021stereoset, abid2021persistent} and producing toxic generations \cite{gehman2020realtoxicityprompts}.
We study which social, linguistic, and geographical biases are inherent in our data selection method in \cref{sec:aboutme} to promote transparency. 
Future research is necessary to study the effect of such biases. 
We recommend that in practice, QuRating should be combined with manual curation of certain languages and topics, and the resulting models should be carefully evaluated for biases before wider deployment. \textedit{We also emphasize that the quality ratings do not measure the social or literary value of a text and should not be used for textual or demographic studies.}

\section*{Acknowledgements}
We thank Luca Soldaini for their generous insights about evaluating data decisions.
We are grateful to Greg Durrett and members of the TAUR lab at UT Austin for noticing inconsistencies in the data presented in the first pre-print.
We thank Mengzhou Xia and Tianyu Gao for their advice on experimental details.
We also thank Carlos E. Jimenez, Tanya Goyal, Paul Röttger, Alexis Chevalier, Sanjeev Arora, Zirui Wang and Jiatong Yu for helpful discussions and feedback. We thank Shreyan Puri for contributing validation data. Finally, we thank the anonymous reviewers for their constructive feedback.
This research is supported by Microsoft Azure credits through the ``Accelerate Foundation Models
Academic Research'' Initiative. This research is also funded by the National Science Foundation (IIS-2211779).

\bibliography{custom}
\bibliographystyle{icml2024}

\appendix
\onecolumn

\section{Full Prompts}\label{app:detailed_prompts}
Our full prompt templates are shown below, where the criteria on the right are substituted for \texttt{\{criterion\}} on the left. We settled on these prompts via a heuristic and iterative process, in which we varied the prompt wording and observed trends on a few examples from SlimPajama \cite{cerebras2023slimpajama}. Throughout this project, we used GPT-3.5-turbo, as we decided that it was too expensive to collect large-scale annotations with GPT-4. We believe that future work should take a more principled approach and first curate a high-quality dataset for prompt refinement. 

We observe better performance with a short and generic system response, namely \texttt{You are a helpful assistant.} than describing a personality with expert skills in all subjects. The effect of different personalities on the subjective judgements in our data selection method is an interesting avenue for future work.

To validate our prompts, we handpick 40 documents from the web that correspond to what we believe should be either highly and poorly rated documents for each criterion. Note that we do not use these data points for prompt refinement. We report the data sources of this dataset in Table \ref{tab:validation_set_sources} below. Our prompts achieve 97.6\% agreement for \textit{writing style}, 91.9\% agreement on \textit{facts \& trivia}, 98.2\% agreement for \textit{educational value} and 98.5\% agreement for \textit{required expertise}.

We add additional instructions that the judgement should not be influenced by the languages present in the texts, the length of the texts---although for the final dataset we compare texts with the same number of tokens---and the order in which the texts are presented. However, these instructions do not suffice to overcome these biases. For example, we observe that GPT-3.5-turbo still exhibits positional bias.

\newtcolorbox{promptbox}[1]{
        boxrule = 1.5pt,
        fontupper = \small\tt,
        fonttitle = \bf\color{black},
        arc = 5pt,
        rounded corners,
        colframe = black,
        colbacktitle = white!97!black,
        colback = white!97!black,
        title = #1,
}

\newtcolorbox{criterionbox}[1]{
        boxrule = 1.5pt,
        fontupper = \small\tt,
        fonttitle = \bf\color{black},
        arc = 5pt,
        rounded corners,
        colframe = black,
        colbacktitle = white!97!blue,
        colback = white!97!blue,
        title = #1,
}

\begin{minipage}[t]{0.59\textwidth}
    \vspace{0pt} %
    \begin{promptbox}{Prompt Template}
    Compare two text excerpts and choose the text which \{criterion\}
    \bigskip
    
    Aspects that should NOT influence your judgement: \\
    1. Which language the text is written in \\
    2. The length of the text \\
    3. The order in which the texts are presented
    \bigskip
    
    Note that the texts are cut off, so you have to infer their contexts.
    The texts might have similar quality, but you should still make a relative judgement and choose the label of the preferred text.
    \bigskip
    
    [Option A] \\
    ... \{text\_a\} ...
    \bigskip
    
    [Option B] \\
    ... \{text\_b\} ...
    \bigskip
    
    Now you have to choose between either A or B. Respond only with a single word.
    \vspace{0.445cm}
    \end{promptbox}%
\end{minipage}%
\hspace{0.01\textwidth}%
\begin{minipage}[t]{0.4\textwidth}
    \vspace{0pt} %
    \begin{criterionbox}{Writing Style}
    has a more polished and beautiful writing style.
    \end{criterionbox}
    \begin{criterionbox}{Facts \& Trivia}
    contains more facts and trivia. Prefer specific facts and obscure trivia over more common knowledge.
    \end{criterionbox}
    \begin{criterionbox}{Educational Value}
    has more educational value, e.g., it includes clear explanations, step-by-step reasoning, or questions and answers.
    \end{criterionbox}
    \begin{criterionbox}{Required Expertise}
    requires greater expertise and prerequisite knowledge to understand it.
    \end{criterionbox}
\end{minipage}

\begin{table}[th]
\centering
\caption{For each of our criteria, we curate 40 documents that exhibit particularly strong or weak qualities. We use this data for prompt tuning and validating model performance. This table gives a description of the sources of documents in this validation set. Examples with citations come from existing NLP datasets.}
\label{tab:validation_set_sources}
\vskip 0.05in
\resizebox{\textwidth}{!}{
\begin{tabular}{lll}
\toprule
\multicolumn{2}{l}{Criterion} &  Sources\\
\midrule
Writing Style & \textit{High} & 11 featured Wikipedia articles, 10 fiction books, 8 academic papers, \\
& & 6 famous speeches, 3 Supreme Court decisions, 2 Shakespeare texts \\
\cmidrule(lr){2-3}
 & \textit{Low} & 10 Yelp reviews \citep{yelp_reviews}, 10 spam messages \citep{sms_spam}, \\
 & & 10 tables/lists, 5 Amazon product reviews, and 3 Enron spam e-mails \citep{enron}
 \\
\cmidrule(lr){1-3}
Facts \& Trivia & \textit{High} & 21 niche Wikipedia articles, 12 fun fact lists, 7 IMDb trivia sections \\
\cmidrule(lr){2-3}
 & \textit{Low} & 15 Wikipedia summaries of Pixar movies, 15 books \citep{pile}, 5 poems, 5 textbook explanations
 \\
\cmidrule(lr){1-3}
Educational Value & \textit{High} & 13 Khan Academy explanations (across subjects), 8 science textbooks, \\
& & 6 history textbooks, 5 high-level Wikipedia articles
\\
\cmidrule(lr){2-3}
 & \textit{Low} & 10 Reality TV transcripts, 10 fantasy/sci-fi books, 10 niche Wikipedia articles, 7 gossip news posts, 3 obscure WikiHow articles
 \\
\cmidrule(lr){1-3}
Required Expertise & \textit{High} & 15 Wikipedia articles, 12 technical academic papers, 9 advanced textbook excerpts, 4 patents/laws \\
\cmidrule(lr){2-3}
 & \textit{Low} & 10 WikiHow articles, 10 Children’s books, 5 nursery rhymes, 5 miscellaneous how-to articles
 \\
\bottomrule
\end{tabular}
}
\end{table}

\begin{table*}[hb]
\tiny
\caption{We handpick a list of 10 documents from various sources and present a ranking which, in the authors' view, reflect steady decreases in writing style, allowing us to test the nuance of LLM judgments.}
\label{tab:ten_wriqual_texts}
\vskip 0.05in
\begin{tabularx}{\textwidth}{lXp{18mm}}
    \toprule
    \textbf{Rank} & \textbf{Text} & \textbf{Description}  \\
    \midrule
    1 & \texttt{Amory Blaine inherited from his mother every trait, except the stray inexpressible few, that made him worth while. His father, an ineffectual, inarticulate man with a taste for Byron and a habit of drowsing over the Encyclopedia Britannica, grew wealthy at thirty through the death of two elder brothers, successful Chicago brokers, and in the first flush of feeling that the world was his, went to Bar Harbor and met Beatrice O'Hara. In consequence, Stephen Blaine handed down to posterity his height of ...} & F. Scott Fitzgerald's \textit{This Side of Paradise}\\
    \midrule
    2 & \texttt{Technologies for making and manipulating DNA have enabled advances in biology ever since the discovery of the DNA double helix. But introducing site-specific modifications in the genomes of cells and organisms remained elusive. Early approaches relied on the principle of site-specific recognition of DNA sequences by oligonucleotides, small molecules, or self-splicing introns. More recently, the site-directed zinc finger nucleases (ZFNs) and TAL effector nucleases (TALENs) using the principle of site-specific ...} & CRISPR-Cas9 paper abstract \citep{crispr}\\
    \midrule
    3 & \texttt{The winter of 1906-07 was the coldest in Alberta's history and was exacerbated by a shortage of coal. One cause of this shortage was the strained relationship between coal miners and mine operators in the province. At the beginning of April 1907, the Canada West Coal and Coke Company locked out the miners from its mine near Taber. The same company was also facing a work stoppage at its mine in the Crow's Nest Pass, where miners were refusing to sign a new contract. The problem spread until by April ...} & featured Wikipedia article\\
    \midrule
    4 & \texttt{On December 3, Venezuela held a controversial referendum over a claim to the oil-rich Essequibo region controlled by Guyana. That same day, the Vice President of Venezuela, Delcy Rodríguez, shared a video on X, formerly Twitter, showing a group of Indigenous people lowering a Guyanese flag and hoisting a Venezuelan flag in its stead over the territory, which is also known as Guayana Esequiba. 'Glory to the brave people!' she wrote, which is the first line of the country's national anthem. The post came ...} & Bellingcat news article \citep{bellingcat}\\
    \midrule
    5 & \texttt{The Godfather is one of the most praised movies in cinema history. It gives everything that critics and audiences alike ask for in movies. In my opinion it gets all the attention it gets for being one of, or the best movies ever. One of the best things The Godfather does is its incredible casting and its iconic performances from each and every one of its characters. The actors are so convincing that it won the movie several academy awards. It also jumpstarted several actors, acting careers, and gave an ...} & IMDb movie review\\
    \midrule
    6 & \texttt{The food is good, but not a great value. Up front, I will just say, do not waste your time getting traditional sushi here because tbh it's not really that much better. For example, we ordered some maki and nigiri and while it was good, it wasn't that much better than our fave sushi places.   Instead, come here for their signature dishes and you'll probably be happier. We really enjoyed some of their signature dishes.   We dined as a party of 4 and we had:  Spicy edamame: tasty and spicy!   Yellowtail ...} & yelp restaurant review \\
    \midrule
    7 & \texttt{My Father worked for a Forbes 500 company since the 70s. Moved up the ranks as a software engineer and management, has patents for the company that saved it millions of dollars. He's almost to pension age and suddenly HR starts making his life miserable. He noticed this trend was happening to some of his coworkers when they were getting close to age 60 as well.  HR Lady calls him into the office and says that he was not punching in and out at the correct time. My Father, an engineer, is very very ... %
    } & reddit post \\
    \midrule
    8 & \texttt{THE ADVENTURE OF LINA AND HER ADVENTUROUS DOG SHERU Lina was a normal girl like any girl.She lived in the hills.She went to the top of the hills and she looked behind a special bush under the rearest of pine trees.She saw many pines behind it,but when she moved the pines she found a large piece of paper in which something was writen.Lina, Lina said her mother.GET UP!!You're late for school!!Oh mom!I'm too tired.Come on you have to go,no arguements.Lina was from a rich family.She lived in Los Anjilous ...
    } & childhood composition by friend of author \\
    \midrule
    9 & \texttt{"Sunshine Quiz Wkly Q! Win a top Sony DVD player if u know which country the Algarve is in? Txt ansr to 82277. Â£1.50 SP:Tyrone Customer service annoncement. You have a New Years delivery waiting for you. Please call 07046744435 now to arrange delivery You are a winner U have been specially selected 2 receive Â£1000 cash or a 4* holiday (flights inc) speak to a live operator 2 claim 0871277810810 URGENT! We are trying to contact you. Last weekends draw shows that you have won a
    Â£900 prize ...
    } & concatenated spam messages \citep{sms_spam}\\
    \midrule
    10 & \texttt{cRjp7tQcwHoNERPRhj7HbiDuessoBAkl8uM0GMr3u8QsHfyGaK7x0vC3L0YGGLA7Gh240GKhDjNwoaBtQubP8tbwrKJCSmRkUbg9aHzO QA4SLWbKcEVAiTfcQ68eQtnIF1IhOoQXLM7RlSHBCqibUCY3Rd0ODHSvgiuMduMDLPwcOxxHCCc7yoQxXRr3qNJuROnWSuEHX5WkwNRS ef5ssqSPXauLOB95CcnWGwblooLGelodhlLEUGI5HeECFkfvtNBgNsn5En628MrUyyFhrqnuFNKiKkXA61oqaGe1zrO3cD0ttidD ...} & randomly generated alphanumeric string\\
    \bottomrule
\end{tabularx}
\vskip -0.05in
\end{table*}

\FloatBarrier
\section{QuRater Model} \label{app:data_quality_modeling}
\subsection{Judgment Dataset} 
We use GPT-3.5-turbo to generate 20 predictions of either ``A'' or ``B'' for a criterion and a pair of documents in either order. When we conducted this work, we did not have access to the logits of the model, and therefore reconstruct the model confidence through multiple generations.
We prompt the LLM for each criterion separately, as we observed in our exploration that performance deteriorated when querying for all criteria together.

We also observe that GPT-3.5-turbo struggles to be decisive when queried with a pair of long documents, i.e., it reverts to choosing the document purely based on positional bias.
We tackle this problem by only ranking pairs of short text snippets. Specifically, we randomly extract segments of $n$ tokens (with respect to the Llama tokenizer \citep{touvron2023llama}) for each pair, where $n$ is chosen randomly according to $n \sim \text{Uniform}[256, 512]$ in half of cases and $n=512$ otherwise.

Table~\ref{tab:gpt_statistics} shows the statistics after querying with 250K pairs.
Note that we only obtain judgements on the English subset of Wikipedia, instead of all the languages present in the Wikipedia split of RedPajama \citep{together2023redpajama}.
For a small subset of queries, we do not obtain predictions, as they are blocked by OpenAI and Azure content filters. The cost for creating this dataset was \$2820.

\begin{figure}[h]
    \centering
    \vskip 0.1in
    \centerline{\includegraphics[width=6.5cm]{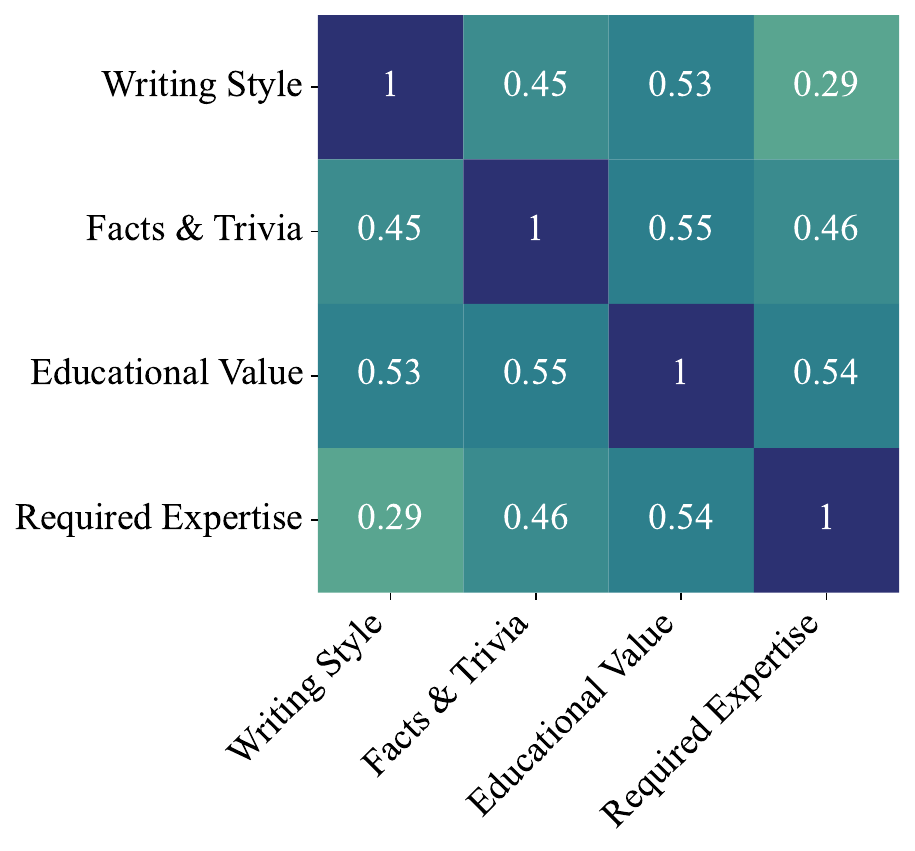}}
    \vskip -0.1in
    \caption{Pearson correlation coefficients between our criteria for predictions made by GPT-3.5-turbo.}
    \vskip -0.1in
    \label{fig:gpt_correlations}
\end{figure}

Figure~\ref{fig:gpt_correlations} shows that while GPT-3.5-turbo predictions are all positively correlated, the Pearson correlation coefficients are less than 0.6, and therefore will differ on many documents. The correlations are typically in the range of 0.45-0.55, except between \textit{required expertise} and \textit{writing style}, which have a correlation of 0.29.

\begin{table}[h]
    \centering
    \caption{Statistics of the qualitative choices generated by GPT-3.5-turbo. While we query with 250K pairs, we do not obtain labels for a small subset due to OpenAI and Azure content filters.
    The confidence margin for a prediction between $(t_A, t_B)$ is defined as $|p_A - p_B| = |2p_{B \succ A} - 1|$.}
    \label{tab:gpt_statistics}
    \vskip 0.05in
    \resizebox{0.85\linewidth}{!}{
    \begin{tabular}{l r rrrrr}
        \toprule
           & & \multicolumn{4}{c}{\# Pairs w/ confidence margin $\ge 50\%$}\\
         \cmidrule{3-6}
         Domains & \# Pairs \quad\quad & Writing Style & Facts \& Trivia & Educational Value & Required Expertise \\
         \midrule
         $\text{CommonCrawl} \cup \text{C4}$ & 124,731 & 95,739 & 88,954 & 92,658 & 96,633 \\
         $\text{Wikipedia (English)}$ & 10,104 & 7,093 & 6,398 & 7,418 & 7,136 \\
         $\text{Book}$ & 9,756 & 6,828 & 6,979 & 7,181 & 7,037 \\
         $\text{StackExchange}$ & 10,185 & 4,927 & 4,524 & 6,780 & 6,740 \\
         $\text{Github}$ & 10,514 & 2,802 & 2,901 & 7,408 & 6,553 \\
         $\text{ArXiv}$ & 10,401 & 5,019 & 5,042 & 6,261 & 5,637 \\
         \textit{other} & 69,966 & 54,578 & 50,200 & 46,059 & 54,390 \\
         \midrule
         Overall & 245,657 & 176,986 & 164,998 & 173,765 & 184,126 \\
         \bottomrule
    \end{tabular}
    }
\end{table}

\begin{table*}[ht]
\centering
\caption{We compare held-out accuracy of training a multi-task QuRater model vs. training separate QuRater models. The multi-task QuRater model is trained with separate prediction heads per criterion. We also show an ablation where we train the QuRater model from scratch instead of initializing with Sheared-Llama-1.3B \cite{xia2023sheared}. During the evaluation, we only predict on text pairs with confident judgement labels, i.e. a confidence (conf.) margin (defined as $|p_A - p_B| = |2p_{B \succ A} - 1|$) greater than 50\% or 80\%.}
\label{tab:data_quality_models}
\vskip 0.05in
\resizebox{0.6\linewidth}{!}{
\begin{tabular}{lc|cccc}
\toprule
Evaluation & Conf.& \textbf{Writing} & \textbf{Facts \&} & \textbf{Edu.} & \textbf{Required} \\
 dataset & margin & \textbf{Style} & \textbf{Trivia} & \textbf{Value} & \textbf{Expertise} \\
\midrule
\multicolumn{6}{c}{\textit{Multi-task model}} \\
\midrule
\multirow[t]{2}{*}{Validation} & 50\% & 94.5 & 93.5 & 93.6 & 95.1 \\
 & 80\% & 97.3 & 97.1 & 95.9 & 97.8 \\
\cmidrule(lr){1-6}
C4 & 50\% & 95.1 & 95.4 & 94.8 & 96.4 \\
Wikipedia (en) & 50\% & 94.4 & 90.1 & 93.3 & 95.6 \\
Book & 50\% & 92.3 & 93.4 & 94.9 & 94.6 \\
StackExchange & 50\% & 93.1 & 92.1 & 91.8 & 92.5 \\
Github & 50\% & 94.5 & 96.1 & 88.8 & 93.0 \\
ArXiv & 50\% & 92.9 & 94.3 & 88.9 & 93.9 \\
\midrule
\multicolumn{6}{c}{\textit{Separate models}} \\
\midrule
\multirow[t]{2}{*}{Validation} & 50\% & 94.4 & 93.4 & 93.1 & 94.9 \\
 & 80\% & 97.3 & 97.1 & 95.5 & 97.8 \\
\midrule
\multicolumn{6}{c}{\textit{Randomly initialized model}} \\
\midrule
\multirow[t]{2}{*}{Validation} & 50\% & 86.5 & 85.2 & 85.3 & 88.1 \\
 & 80\% & 90.5 & 90.1 & 88.3 & 92.5 \\
\bottomrule
\end{tabular}
}
\end{table*}

\begin{table*}[ht]
\centering
\caption{Number of sequences in the 260B token corpus from which we select data. The data is a subset of SlimPajama, where each document is processed into sequences of exactly 1024 tokens. Therefore, the proportion of domains is different from the raw SlimPajama.}
\label{tab:domain_stats}
\vskip 0.05in
\begin{tabular}{lrrr}
\toprule
Domain & \# Sequences & \# Tokens & Proportion \\
\midrule
CommonCrawl & 153,437,203 & 157,119,695,872 & 60.4 \\
C4 & 40,991,721 & 41,975,522,304 & 16.1 \\
ArXiv & 16,513,627 & 16,909,954,048 & 6.5 \\
Book & 15,676,440 & 16,052,674,560 & 6.2 \\
Github & 14,806,859 & 15,162,223,616 & 5.8 \\
Wikipedia & 7,741,248 & 7,927,037,952 & 3.0 \\
StackExchange & 4,974,184 & 5,093,564,416 & 2.0 \\
\midrule
Total & 254,141,282 & 260,240,672,768 & 100.0 \\
\bottomrule
\end{tabular}
\end{table*}

\FloatBarrier 
\subsection{QuRater Training}
We fine-tune QuRater models using Sheared-Llama-1.3B \citep{xia2023sheared}, a pruned version of Llama-2-7B \citep{touvron2023llama2}.
We add four linear regression heads to the transformer outputs at the last token of the sequence, which predict the quality ratings across the four criteria. This multi-task setup allows for fast inference of all criteria in one forward pass.
We only train and evaluate on judgements that have a confidence margin of at least 50\%, where the confidence margin is defined as $|p_A - p_B| = |2p_{B \succ A} - 1|$ for a prediction between $(t_A, t_B)$, since non-confident predictions contain little signal on data quality,
and can be caused by GPT-3.5-turbo's positional bias.
Table~\ref{tab:gpt_statistics} shows the effect of this filtering on dataset statistics.
We use a random 10\% of the dataset as a held-out validation split for early stopping and hyperparameter selection. For early stopping, we choose based on held-out accuracy averaged across all criteria.

We search over the following hyperparameter grid: learning rate $\in \{2\times 10^{-5}, 5\times 10^{-5}\}$, number of epochs $\in \{2, 4\}$, batch size of 512.
Model selection is based on which model achieves the best validation examples on the criterion with lowest performance overall. The selected model was trained with a learning rate of $5 \times 10^{-5}$ and is trained for 2 epochs.

In Table~\ref{tab:data_quality_models}, we report accuracy on the validation set, as well on specially procured test sets of 1,428 pairs of texts from specific domains.
We confirm that performance increases when evaluating only on confident GPT-3.5-turbo's judgements, where \textit{educational value} is the hardest category to predict.
Finally, we compare to separately fine-tuned models and model trained from a random initialization.
Multi-task fine-tuning usually gives comparable or better performance, and a pre-trained initialization helps substantially in this task.

\FloatBarrier

\section{Connection of Exponential Sampling to RLHF} \label{app:rlhf}
In RLHF \citep{ouyang2022training}, a language model $p(y|x)$ is fine-tuned to produce outputs $y$ given inputs $x$ that maximize a reward $r(x, y)$ subject to a relaxed KL constraint with respect to a reference langugage model $p_\text{ref}(y|x)$,
$$
p^*(y|x) = \argmax_p \mathop{\mathbb{E}}_{y \sim p(\cdot|x)} \left[ r(y) - \tau \log \frac{p(y|x)}{p_\text{ref}(y|x)} \right]
$$
Typically, the rewards encourage the model to act in a helpful and harmless manner \citep{bai2022training}.

It can be shown that this admits the closed-form solution \citep{korbak2022rl, rafailov2023direct, go2023aligning, korbak2022on}
$$
p^*(y|x) = \frac{1}{Z(x)} \; p_\text{ref} \left(y | x\right) \exp\left(\frac{r(x, y)}{\tau}\right).
$$

Consider the following setting: (1) we use the QuRater model $s(y)$ as reward model not conditioned on any user input, (2) the reference model is a language model $p_\mathcal{D}(y)$ pre-trained on a corpus $\mathcal{D}$. In that case, we write the optimal policy as:
$$
p^*(y) = \frac{1}{Z} \; p_\mathcal{D} \left(y\right) \exp\left(\frac{s(y)}{\tau}\right).
$$

We compare this optimal model with the model obtained from maximum log-likelihood optimization (i.e. language model training), where a document $y$ is resampled with a probability $\propto \exp \left(\frac{s(y)}{\tau} \right)$. Let $\hat p_\mathcal{D}(y)$ be the underlying data distribution of the training corpus $\mathcal{D}$, resulting in the weighted cross-entropy objective,
$$
\argmax_p \sum_y \exp\left( {s(y)} / {\tau}\right) p_\mathcal{D}(y)  \log p\left(y\right),
$$
which in practice is approximated via MonteCarlo sampling and importance resampling with the exponential quality ratings.
The optimal model to this objective is $~\hat p_\mathcal{D}(y) \exp\left( \frac{s(y)}{\tau}\right).$  Assuming $\mathcal{D}$ is large and $p_\mathcal{D}(y)$ approximates $\hat p_\mathcal{D}(y)$ sufficiently well, the maximum likelihood solution to the resampled distribution will approximate the optimal policy $p*(y)$.

In summary, our sampling strategy is equivalent to (1) training a language model on the entire dataset, and then (2) using RLHF to guide the language model towards generating documents with higher quality ratings.

\section{Experimental Details} \label{app:training_details}
Each data selection method retains the original domain proportions between the RedPajama subsets. Table~\ref{tab:domain_stats} shows the domain statistics of the 260B \ourdata{}, from which we select 30B tokens using different data selection methods.
\textit{\ourdata{} is a curated subset of SlimPajama, which is itself a subset of RedPajama. Both SlimPajama and RedPajama are released on HuggingFace under the Apache 2.0 License.}

We use a global batch size of 2048 sequences and a learning rate of $5\times10^{-4}$ with a cosine learning rate decay to $5\times10^{-5}$ and a linear warmup for the first $5\%$ of training steps.
Each model is trained on 8x NVIDIA H100, which costs 200 GPU hours for 30B tokens. We use a weight decay of $0.1$ and train with Adam \citep{Adam} with hyperparameters $\beta = (0.9, 0.95)$.
We train a 1.3B parameter transformer model with RoPE embedding \citep{su2024roformer} and SwiGLU activations \citep{shazeer2020glu}.

\paragraph{In-context learning settings.}
We choose a different number of few-shot examples per task to ensure that all demonstrations fit within the context window of 1024 tokens.
We use the following number of demonstrations (given in parentheses): ARC-easy (15), ARC-challenge (15), SciQA (2), LogiQA (2), BoolQ (0), HellaSwag (6), PIQA (6), WinoGrande (15), NQ (10), MMLU (10). We report accuracy for all tasks, except for NQ, where we report EM.
When available, we use the normalized accuracy metric provided by \texttt{lm-evaluation-harness}.

\begin{figure}[h]
    \centering
    \vskip 0.1in
    \centerline{\includegraphics[width=\linewidth]{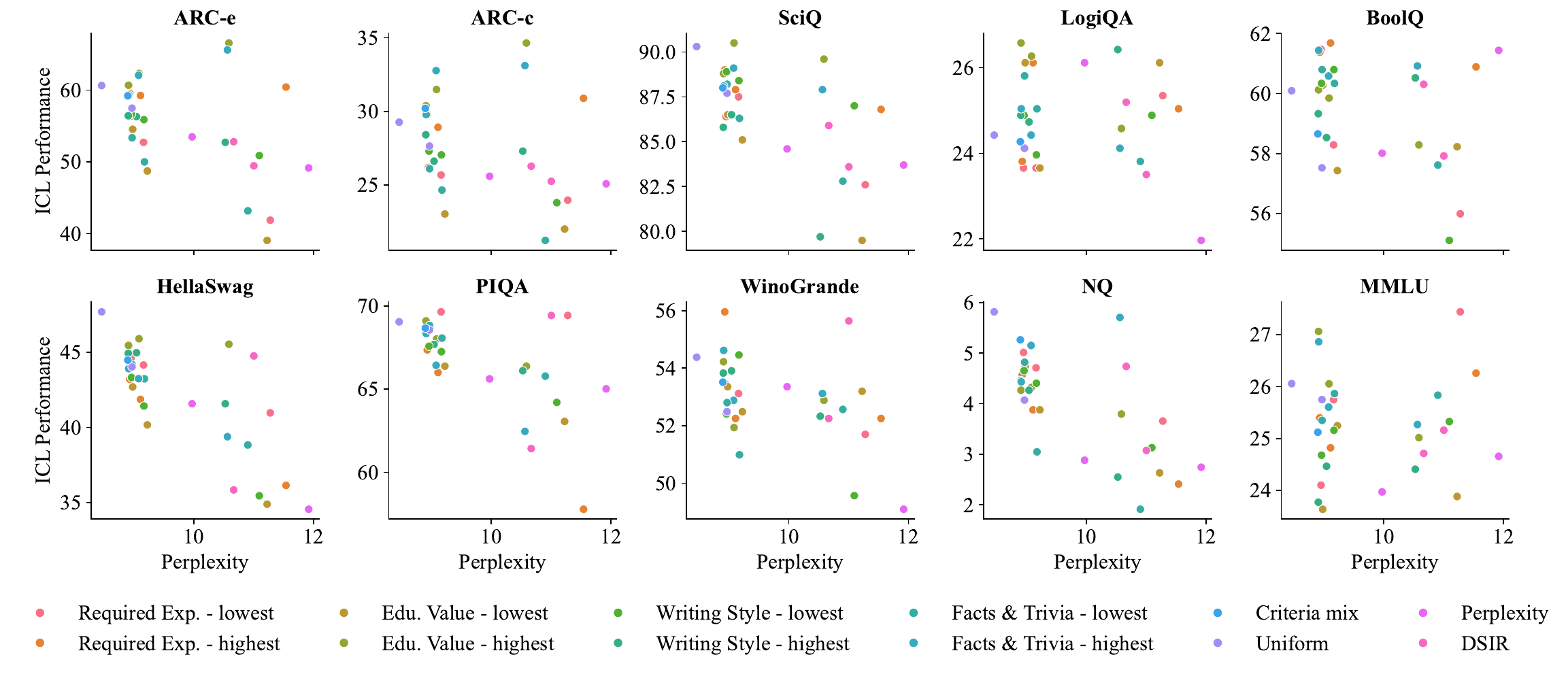}}
    \caption{We plot the relationship between the perplexity results and in-context learning performance of models in Tables~\ref{tab:ppl_results_domains} and~\ref{tab:icl_results_full}.
    While prior work has found perplexity to be a good predictor of downstream task performance when varying model parameters and number of training tokens \citep{xia2023training, du2024understanding}, we observe that this is not true when varying the training distribution.
    }
    \label{fig:icl_vs_ppl}
    \vskip -0.1in
\end{figure}

\paragraph{Detailed results.} We feature the full perplexity results in Table~\ref{tab:ppl_results_domains}, including the perplexity for each of the RedPajama subsets. Table~\ref{tab:icl_results_full} contains the ICL performance for all models.
The performance of the curriculum models is featured at the top of the tables. In Figure~\ref{fig:icl_vs_ppl}, we plot the relationship between perplexity and ICL task performance across models.
 
\begin{table}[ht]
\centering
\caption{Held-out per-token perplexity per RedPajama domain between language models trained on 30B tokens from different data selection methods. We highlight the best result in each column (before rounding). \textit{bottom-$k$} and \textit{inv.} denote inverse sampling, in which we sample documents with the lowest quality ratings. Abbreviations: HellaSw. = HellaSwag, W.G. = WinoGrande, exp. = expertise.}
\label{tab:ppl_results_domains}
\vskip 0.05in
\resizebox{\textwidth}{!}{
\begin{tabular}{llcccccccccc}
\toprule
\multicolumn{2}{l}{Selection Method} &  \textbf{CC}  &  \textbf{C4}  &  \textbf{Github}  &  \textbf{Wiki}  &  \textbf{ArXiv}  &  \textbf{StackEx}  &  \textbf{Book}  &  \textbf{Overall} \\
\midrule
\multicolumn{2}{l}{Uniform} & 9.81  & 11.66  & 2.58  & 9.46  & 5.21  & 4.31  & 12.04  & 8.96  \\
\multicolumn{2}{l}{+ curriculum: low-to-high exp.} &  9.76 \dar{0.05} & 11.70 \uar{0.04} & 2.56 \dar{0.02} & 9.30 \dar{0.16} & 5.11 \dar{0.10} & 4.28 \dar{0.03} & 11.92 \dar{0.12} & 8.92 \dar{0.04} \\
\multicolumn{2}{l}{+ curriculum: high-to-low exp.} & 9.80 \dar{0.01} & 11.58 \dar{0.08} & 2.56 \dar{0.02} & 9.37 \dar{0.09} & 5.40 \uar{0.19} & 4.26 \dar{0.05} & 12.08 \uar{0.04} & 8.96 \uar{0.00} \\
\cmidrule(lr){1-10}
DSIR & \textit{Wiki} & 11.10 \uar{1.29} & 15.19 \uar{3.53} & 3.07 \uar{0.49} & 18.26 \uar{8.80} & 6.24 \uar{1.03} & 5.09 \uar{0.78} & 14.47 \uar{2.43} & 10.67 \uar{1.71} \\
& \textit{Books} & 11.97 \uar{2.16} & 14.56 \uar{2.90} & 3.03 \uar{0.45} & 22.75 \uar{13.29} & 6.18 \uar{0.97} & 5.03 \uar{0.72} & 12.21 \uar{0.17} & 11.00 \uar{2.04} \\
\cmidrule(lr){1-2}
Perplexity & \textit{lowest} & 12.93 \uar{3.12} & 15.57 \uar{3.91} & 3.32 \uar{0.74} & 14.43 \uar{4.97} & 6.17 \uar{0.96} & 5.50 \uar{1.19} & 18.49 \uar{6.45} & 11.92 \uar{2.96} \\
& \textit{highest} & 11.06 \uar{1.25} & 13.06 \uar{1.40} & 2.90 \uar{0.32} & 10.58 \uar{1.12} & 5.56 \uar{0.35} & 4.64 \uar{0.33} & 12.14 \uar{0.10} & 9.97 \uar{1.01}  \\
\cmidrule(lr){1-10}
Writing & \textit{top-$k$} & \edit{11.50 \uar{1.69}} & \edit{14.56 \uar{2.90}} & \edit{2.98 \uar{0.40}} & \edit{16.52 \uar{7.06}} & \edit{5.37 \uar{0.16}} & \edit{5.04 \uar{0.73}} & \edit{12.03 \dar{0.01}} & \edit{10.53 \uar{1.57}} \\
Style & \textit{$\tau$=1.0} & 9.89 \uar{0.08} & 12.11 \uar{0.45} & 2.57 \dar{0.01} & 9.70 \uar{0.24} & 5.14 \dar{0.07} & 4.35 \uar{0.04} & \textbf{11.62 \dar{0.42}} & 9.04 \uar{0.08} \\
 & \textit{$\tau$=2.0} & 9.74 \dar{0.07} & 11.74 \uar{0.08} & 2.56 \dar{0.02} & 9.40 \dar{0.06} & 5.14 \dar{0.07} & 4.28 \dar{0.03} & 11.70 \dar{0.34} & 8.90 \dar{0.06} \\
\cmidrule(lr){1-2}
Facts \& & \textit{top-$k$} & \edit{10.92 \uar{1.11}} & \edit{14.68 \uar{3.02}} & \edit{2.99 \uar{0.41}} & \edit{32.15 \uar{22.69}} & \edit{5.53 \uar{0.32}} & \edit{5.15 \uar{0.84}} & \edit{13.92 \uar{1.88}} & \edit{10.56 \uar{1.60}} \\
Trivia & \textit{$\tau$=1.0} & 9.81 \uar{0.00} & 12.40 \uar{0.74} & 2.57 \dar{0.01} & 10.38 \uar{0.92} & 5.13 \dar{0.08} & 4.33 \uar{0.02} & 12.16 \uar{0.12} & 9.08 \uar{0.12} \\
 & \textit{$\tau$=2.0} & \textbf{9.70 \dar{0.11}} & 11.86 \uar{0.20} & 2.55 \dar{0.03} & 9.61 \uar{0.15} & 5.12 \dar{0.09} & 4.27 \dar{0.04} & 11.96 \dar{0.08} & 8.91 \dar{0.05} \\
\cmidrule(lr){1-2}
Educational & \textit{top-$k$} & \edit{11.41 \uar{1.60}} & \edit{14.30 \uar{2.64}} & \edit{2.94 \uar{0.36}} & \edit{18.66 \uar{9.20}} & \edit{5.26 \uar{0.05}} & \edit{4.90 \uar{0.59}} & \edit{14.22 \uar{2.18}} & \edit{10.59 \uar{1.63}} \\
Value & \textit{$\tau$=1.0} & 9.92 \uar{0.11} & 12.11 \uar{0.45} & 2.57 \dar{0.01} & 10.04 \uar{0.58} & 5.07 \dar{0.14} & 4.31 \uar{0.00} & 12.14 \uar{0.10} & 9.08 \uar{0.12} \\
 & \textit{$\tau$=2.0} & 9.74 \dar{0.07} & 11.71 \uar{0.05} & \textbf{2.55 \dar{0.03}} & 9.51 \uar{0.05} & 5.09 \dar{0.12} & 4.26 \dar{0.05} & 11.93 \dar{0.11} & 8.91 \dar{0.05} \\
\cmidrule(lr){1-2}
Required & \textit{top-$k$} & \edit{12.67 \uar{2.86}} & \edit{16.74 \uar{5.08}} & \edit{3.03 \uar{0.45}} & \edit{14.63 \uar{5.17}} & \edit{5.08 \dar{0.13}} & \edit{5.33 \uar{1.02}} & \edit{15.12 \uar{3.08}} & \edit{11.54 \uar{2.58}} \\
Expertise & \textit{$\tau$=1.0} & 9.95 \uar{0.14} & 12.35 \uar{0.69} & 2.57 \dar{0.01} & 9.50 \uar{0.04} & \textbf{5.03 \dar{0.18}} & 4.29 \dar{0.02} & 12.09 \uar{0.05} & 9.11 \uar{0.15} \\
 & \textit{$\tau$=2.0} & 9.77 \dar{0.04} & 11.83 \uar{0.17} & 2.55 \dar{0.03} & 9.31 \dar{0.15} & 5.09 \dar{0.12} & 4.25 \dar{0.06} & 11.92 \dar{0.12} & 8.93 \dar{0.03} \\
\cmidrule(lr){1-2}
Criteria mix & \textit{$\tau$=2.0} & 9.71 \dar{0.10} & 11.75 \uar{0.09} & 2.55 \dar{0.03} & 9.60 \uar{0.14} & 5.12 \dar{0.09} & 4.27 \dar{0.04} & 11.83 \dar{0.21} & \textbf{8.90 \dar{0.06}} \\
\cmidrule(lr){1-10}
Writing & \textit{bottom-$k$} & 12.16 \uar{2.35} & 14.31 \uar{2.65} & 3.07 \uar{0.49} & 12.71 \uar{3.25} & 6.00 \uar{0.79} & 4.91 \uar{0.60} & 16.29 \uar{4.25} & 11.10 \uar{2.14} \\
Style & \textit{inv. $\tau$=1.0} & 10.08 \uar{0.27} & 11.72 \uar{0.06} & 2.57 \dar{0.01} & 9.59 \uar{0.13} & 5.27 \uar{0.06} & 4.26 \dar{0.05} & 12.80 \uar{0.76} & 9.16 \uar{0.20} \\
 & \textit{inv. $\tau$=2.0} & 9.83 \uar{0.02} & 11.52 \dar{0.14} & 2.55 \dar{0.03} & 9.35 \dar{0.11} & 5.20 \dar{0.01} & \textbf{4.24 \dar{0.07}} & 12.27 \uar{0.23} & 8.96 \dar{0.00} \\
\cmidrule(lr){1-2}
Facts \& & \textit{bottom-$k$} & 12.10 \uar{2.29} & 13.30 \uar{1.64} & 3.27 \uar{0.69} & 11.48 \uar{2.02} & 6.57 \uar{1.36} & 5.22 \uar{0.91} & 14.20 \uar{2.16} & 10.90 \uar{1.94} \\
Trivia & \textit{inv. $\tau$=1.0} & 10.16 \uar{0.35} & 11.56 \dar{0.10} & 2.59 \uar{0.01} & 9.45 \dar{0.01} & 5.38 \uar{0.17} & 4.32 \uar{0.01} & 12.32 \uar{0.28} & 9.17 \uar{0.21} \\
 & \textit{inv. $\tau$=2.0} & 9.88 \uar{0.07} & \textbf{11.45 \dar{0.21}} & 2.56 \dar{0.02} & 9.28 \dar{0.18} & 5.24 \uar{0.03} & 4.27 \dar{0.04} & 12.05 \uar{0.01} & 8.97 \uar{0.01} \\
\cmidrule(lr){1-2}
Educational & \textit{bottom-$k$} & 12.29 \uar{2.48} & 14.37 \uar{2.71} & 3.45 \uar{0.87} & 11.99 \uar{2.53} & 6.33 \uar{1.12} & 5.61 \uar{1.30} & 14.90 \uar{2.86} & 11.23 \uar{2.27} \\
Value & \textit{inv. $\tau$=1.0} & 10.14 \uar{0.33} & 11.84 \uar{0.18} & 2.63 \uar{0.05} & 9.46 \uar{0.00} & 5.39 \uar{0.18} & 4.42 \uar{0.11} & 12.44 \uar{0.40} & 9.22 \uar{0.26} \\
 & \textit{inv. $\tau$=2.0} & 9.85 \uar{0.04} & 11.56 \dar{0.10} & 2.58 \dar{0.00} & \textbf{9.28 \dar{0.18}} & 5.25 \uar{0.04} & 4.31 \dar{0.00} & 12.09 \uar{0.05} & 8.98 \uar{0.02} \\
\cmidrule(lr){1-2}
Required & \textit{bottom-$k$} & 12.35 \uar{2.54} & 13.83 \uar{2.17} & 3.63 \uar{1.05} & 13.22 \uar{3.76} & 6.45 \uar{1.24} & 5.72 \uar{1.41} & 15.08 \uar{3.04} & 11.28 \uar{2.32} \\
Expertise & \textit{inv. $\tau$=1.0} & 10.07 \uar{0.26} & 11.60 \dar{0.06} & 2.64 \uar{0.06} & 9.70 \uar{0.24} & 5.38 \uar{0.17} & 4.42 \uar{0.11} & 12.39 \uar{0.35} & 9.16 \uar{0.20} \\
 & \textit{inv. $\tau$=2.0} & 9.82 \uar{0.01} & 11.46 \dar{0.20} & 2.58 \dar{0.00} & 9.43 \dar{0.03} & 5.25 \uar{0.04} & 4.31 \dar{0.00} & 12.07 \uar{0.03} & 8.95 \dar{0.01} \\
\midrule
\multicolumn{2}{l}{\textit{Uniform\;\;+50\% data}} & \textit{9.25 \dar{0.56}} & \textit{10.97 \dar{0.69}} & \textit{2.47 \dar{0.11}} & \textit{8.61 \dar{0.85}} & \textit{4.98 \dar{0.23}} & \textit{4.08 \dar{0.23}} & \textit{11.34 \dar{0.70}} & \textit{8.46 \dar{0.50}} \\
\bottomrule
\end{tabular}
}
\end{table}

\begin{table*}[ht]
\centering
\caption{The in-context learning performance across all our models. We report accuracy for all tasks, except for NQ, where we report EM, and highlight the best result in each column (before rounding). \textit{bottom-$k$} and \textit{inv.} denote inverse sampling, in which we sample documents with the lowest quality ratings. Abbreviations: HellaSw. = HellaSwag, W.G. = WinoGrande, exp. = expertise.}
\label{tab:icl_results_full}
\vskip 0.1in
\resizebox{\textwidth}{!}{
\begin{tabular}{llccccccccccc}
\toprule
& &  \multicolumn{5}{c}{Reading Comprehension} & \multicolumn{3}{c}{Commonsense Reasoning} & \multicolumn{2}{c}{World Knowledge} \\
\cmidrule(lr){3-7} \cmidrule(lr){8-10} \cmidrule(lr){11-12}
 & & \textbf{ARC-E}  &  \textbf{ARC-C}  &  \textbf{SciQ}  &  \textbf{LogiQA}  &  \textbf{BoolQ}  &  \textbf{HellaSw.}  &  \textbf{PIQA}  &  \textbf{W.G.}  & \textbf{NQ} & \textbf{MMLU}  &  \\
\multicolumn{2}{l}{\textbf{Selection Method}} & (15) & (15) & (2) & (2) & (0) & (6) & (6) & (15) & (10) & (5) & \textbf{Average} \\
\midrule
\multicolumn{2}{l}{Uniform}  & 57.5  & 27.6  & 87.7  & 24.1  & 57.5  & 44.0  & 68.6  & 52.5  & 4.1  & 25.7  & 44.9  \\
\multicolumn{2}{l}{+ curriculum: low-to-high exp.} & 58.0 \ua{0.5} & 28.1 \ua{0.5} & 87.0 \da{0.7} & 26.0 \ua{1.9} & 59.6 \ua{2.1} & 43.7 \da{0.3} & 67.5 \da{1.1} & 53.9 \ua{1.4} & 4.8 \ua{0.7} & 26.4 \ua{0.7} & 45.5 \ua{0.6} \\
\multicolumn{2}{l}{+ curriculum: high-to-low exp.} & 56.6 \da{0.9} & 28.8 \ua{1.2} & 89.7 \ua{2.0} & 24.3 \ua{0.2} & 55.2 \da{2.3} & 44.7 \ua{0.7} & 69.3 \ua{0.7} & 53.0 \ua{0.5} & 5.5 \ua{1.4} & 27.2 \ua{1.5} & 45.4 \ua{0.5} \\
\cmidrule(lr){1-13}
DSIR & \textit{Wiki} & 52.8 \da{4.7} & 26.3 \da{1.3} & 85.9 \da{1.8} & 25.2 \ua{1.1} & 60.3 \ua{2.8} & 35.8 \da{8.2} & 61.4 \da{7.2} & 52.2 \da{0.3} & 4.7 \ua{0.6} & 24.7 \da{1.0} & 42.9 \da{2.0} \\
&\textit{Books} & 49.5 \da{8.0} & 25.3 \da{2.3} & 83.6 \da{4.1} & 23.5 \da{0.6} & 57.9 \ua{0.4} & 44.8 \ua{0.8} & {69.4 \ua{0.8}} & 55.6 \ua{3.1} & 3.1 \da{1.0} & 25.2 \da{0.5} & 43.8 \da{1.1} \\
\cmidrule(lr){1-2}
Perplexity & \textit{lowest} & 49.2 \da{8.3} & 25.1 \da{2.5} & 83.7 \da{4.0} & 22.0 \da{2.1} & {61.4 \ua{3.9}} & 34.6 \da{9.4} & 65.0 \da{3.6} & 49.1 \da{3.4} & 2.7 \da{1.4} & 24.7 \da{1.0} & 41.7 \da{3.2} \\
& \textit{highest} & 53.5 \da{4.0} & 25.6 \da{2.0} & 84.6 \da{3.1} & 26.1 \ua{2.0} & 58.0 \ua{0.5} & 41.6 \da{2.4} & 65.6 \da{3.0} & 53.4 \ua{0.9} & 2.9 \da{1.2} & 24.0 \da{1.7} & 43.5 \da{1.4} \\
\cmidrule(lr){1-13}
Writing & \textit{top-$k$} & \edit{52.7 \da{4.8}} & \edit{27.3 \da{0.3}} & \edit{79.7 \da{8.0}} & \edit{26.4 \ua{2.3}} & \edit{60.5 \ua{3.0}} & \edit{41.6 \da{2.4}} & \edit{66.1 \da{2.5}} & \edit{52.3 \da{0.2}} & \edit{2.5 \da{1.6}} & \edit{24.4 \da{1.3}} & \edit{43.4 \da{1.5}} \\
Style & \textit{$\tau$=1.0} & 56.3 \da{1.2} & 26.6 \da{1.0} & 86.5 \da{1.2} & 24.7 \ua{0.6} & 58.5 \ua{1.0} & 45.0 \ua{1.0} & 67.7 \da{0.9} & 53.9 \ua{1.4} & 4.3 \ua{0.2} & 24.5 \da{1.2} & 44.8 \da{0.1} \\
 & \textit{$\tau$=2.0} & 56.4 \da{1.1} & 28.4 \ua{0.8} & 85.8 \da{1.9} & 24.9 \ua{0.8} & 59.3 \ua{1.8} & 44.9 \ua{0.9} & 68.6 \phantom{\da{0.0}}  & 53.8 \ua{1.3} & 4.5 \ua{0.4} & 23.8 \da{1.9} & 45.0 \ua{0.1} \\
\cmidrule(lr){1-2}
Facts \& & \textit{top-$k$} & \edit{65.6 \ua{8.1}} & \edit{33.1 \ua{5.5}} & \edit{87.9 \ua{0.2}} & \edit{24.1 \phantom{\ua{0.0}}}  & \edit{60.9 \ua{3.4}} & \edit{39.4 \da{4.6}} & \edit{62.5 \da{6.1}} & \edit{53.1 \ua{0.6}} & \edit{\textbf{5.7 \ua{1.6}}} & \edit{25.3 \da{0.4}} & \edit{45.8 \ua{0.9}} \\
Trivia & \textit{$\tau$=1.0} & 62.1 \ua{4.6} & 32.8 \ua{5.2} & 89.1 \ua{1.4} & 24.4 \ua{0.3} & 60.6 \ua{3.1} & 43.2 \da{0.8} & 66.4 \da{2.2} & 52.9 \ua{0.4} & 5.2 \ua{1.1} & 25.6 \da{0.1} & 46.2 \ua{1.3} \\
 & \textit{$\tau$=2.0} & 59.3 \ua{1.8} & 29.8 \ua{2.2} & 88.1 \ua{0.4} & 25.0 \ua{0.9} & 61.4 \ua{3.9} & 43.9 \da{0.1} & 68.3 \da{0.3} & 54.6 \ua{2.1} & 4.4 \ua{0.3} & 26.9 \ua{1.2} & 46.2 \ua{1.3} \\
\cmidrule(lr){1-2}
Educational & \textit{top-$k$} & \edit{\textbf{66.6 \ua{9.1}}} & \edit{\textbf{34.6 \ua{7.0}}} & \edit{89.6 \ua{1.9}} & \edit{24.6 \ua{0.5}} & \edit{58.3 \ua{0.8}} & \edit{45.5 \ua{1.5}} & \edit{66.4 \da{2.2}} & \edit{52.9 \ua{0.4}} & \edit{3.8 \da{0.3}} & \edit{25.0 \da{0.7}} & \edit{\textbf{46.7 \ua{1.8}}} \\
Value & \textit{$\tau$=1.0} & 62.3 \ua{4.8} & 31.5 \ua{3.9} & \textbf{90.5 \ua{2.8}} & 26.3 \ua{2.2} & 59.8 \ua{2.3} & \textbf{45.9 \ua{1.9}} & 68.0 \da{0.6} & 51.9 \da{0.6} & 4.3 \ua{0.2} & 26.1 \ua{0.4} & 46.7 \ua{1.8} \\
 & \textit{$\tau$=2.0} & 60.7 \ua{3.2} & 30.4 \ua{2.8} & 88.8 \ua{1.1} & \textbf{26.6 \ua{2.5}} & 60.1 \ua{2.6} & 45.4 \ua{1.4} & 69.1 \ua{0.5} & 54.2 \ua{1.7} & 4.3 \ua{0.2} & 27.1 \ua{1.4} & 46.7 \ua{1.8} \\
\cmidrule(lr){1-2}
Required & \textit{top-$k$} & \edit{60.4 \ua{2.9}} & \edit{30.9 \ua{3.3}} & \edit{86.8 \da{0.9}} & \edit{25.0 \ua{0.9}} & \edit{60.9 \ua{3.4}} & \edit{36.1 \da{7.9}} & \edit{57.8 \da{10.8}} & \edit{52.2 \da{0.3}} & \edit{2.4 \da{1.7}} & \edit{26.3 \ua{0.6}} & \edit{43.9 \da{1.0}} \\
Expertise & \textit{$\tau$=1.0} & 59.3 \ua{1.8} & 28.9 \ua{1.3} & 87.9 \ua{0.2} & 26.1 \ua{2.0} & \textbf{61.7 \ua{4.2}} & 41.9 \da{2.1} & 66.0 \da{2.6} & 52.2 \da{0.3} & 3.9 \da{0.2} & 24.8 \da{0.9} & 45.3 \ua{0.4} \\
 & \textit{$\tau$=2.0} & 59.6 \ua{2.1} & 29.8 \ua{2.2} & 89.0 \ua{1.3} & 23.8 \da{0.3} & 61.4 \ua{3.9} & 43.2 \da{0.8} & 67.4 \da{1.2} & \textbf{56.0 \ua{3.5}} & 4.6 \ua{0.5} & 25.4 \da{0.3} & 46.0 \ua{1.1} \\
\cmidrule(lr){1-2}
Criteria mix & $\tau=2.0$ & 59.2 \ua{1.7} & 30.2 \ua{2.6} & 88.0 \ua{0.3} & 24.3 \ua{0.2} & 58.7 \ua{1.2} & 44.5 \ua{0.5} & 68.7 \ua{0.1} & 53.5 \ua{1.0} & 5.3 \ua{1.2} & 25.1 \da{0.6} & 45.7 \ua{0.8} \\
\cmidrule(lr){1-13}
Writing & \textit{bottom-$k$} & 50.9 \da{6.6} & 23.8 \da{3.8} & 87.0 \da{0.7} & 24.9 \ua{0.8} & 55.1 \da{2.4} & 35.5 \da{8.5} & 64.2 \da{4.4} & 49.6 \da{2.9} & 3.1 \da{1.0} & 25.3 \da{0.4} & 41.9 \da{3.0} \\
Style & \textit{inv. $\tau$=1.0} & 55.9 \da{1.6} & 27.0 \da{0.6} & 88.4 \ua{0.7} & 24.0 \da{0.1} & 60.8 \ua{3.3} & 41.4 \da{2.6} & 67.2 \da{1.4} & 54.5 \ua{2.0} & 4.4 \ua{0.3} & 25.2 \da{0.5} & 44.9 \da{0.0} \\
 & \textit{inv. $\tau$=2.0} & 56.6 \da{0.9} & 27.3 \da{0.3} & 88.9 \ua{1.2} & 24.9 \ua{0.8} & 60.3 \ua{2.8} & 43.3 \da{0.7} & 67.6 \da{1.0} & 52.4 \da{0.1} & 4.7 \ua{0.6} & 24.7 \da{1.0} & 45.1 \ua{0.2} \\
\cmidrule(lr){1-2}
Facts \& & \textit{bottom-$k$} & 43.2 \da{14.3} & 21.2 \da{6.4} & 82.8 \da{4.9} & 23.8 \da{0.3} & 57.6 \ua{0.1} & 38.8 \da{5.2} & 65.8 \da{2.8} & 52.6 \ua{0.1} & 1.9 \da{2.2} & 25.8 \ua{0.1} & 41.4 \da{3.5} \\
Trivia & \textit{inv. $\tau$=1.0} & 50.0 \da{7.5} & 24.7 \da{2.9} & 86.3 \da{1.4} & 25.0 \ua{0.9} & 60.3 \ua{2.8} & 43.2 \da{0.8} & 68.1 \da{0.5} & 51.0 \da{1.5} & 3.0 \da{1.1} & 25.9 \ua{0.2} & 43.8 \da{1.1} \\
 & \textit{inv. $\tau$=2.0} & 53.4 \da{4.1} & 26.1 \da{1.5} & 88.2 \ua{0.5} & 25.8 \ua{1.7} & 60.8 \ua{3.3} & 44.2 \ua{0.2} & 68.8 \ua{0.2} & 52.8 \ua{0.3} & 4.8 \ua{0.7} & 25.4 \da{0.3} & 45.0 \ua{0.1} \\
\cmidrule(lr){1-2}
Educational & \textit{bottom-$k$} & 39.1 \da{18.4} & 22.0 \da{5.6} & 79.5 \da{8.2} & 26.1 \ua{2.0} & 58.2 \ua{0.7} & 34.9 \da{9.1} & 63.1 \da{5.5} & 53.2 \ua{0.7} & 2.6 \da{1.5} & 23.9 \da{1.8} & 40.3 \da{4.6} \\
Value & \textit{inv. $\tau$=1.0} & 48.7 \da{8.8} & 23.0 \da{4.6} & 85.1 \da{2.6} & 23.7 \da{0.4} & 57.4 \da{0.1} & 40.2 \da{3.8} & 66.4 \da{2.2} & 52.5 \phantom{\ua{0.0}}  & 3.9 \da{0.2} & 25.3 \da{0.4} & 42.6 \da{2.3} \\
 & \textit{inv. $\tau$=2.0} & 54.5 \da{3.0} & 26.1 \da{1.5} & 86.5 \da{1.2} & 26.1 \ua{2.0} & 60.3 \ua{2.8} & 42.7 \da{1.3} & 68.7 \ua{0.1} & 53.4 \ua{0.9} & 4.7 \ua{0.6} & 23.7 \da{2.0} & 44.7 \da{0.2} \\
\cmidrule(lr){1-2}
Required & \textit{bottom-$k$} & 41.9 \da{15.6} & 24.0 \da{3.6} & 82.6 \da{5.1} & 25.3 \ua{1.2} & 56.0 \da{1.5} & 41.0 \da{3.0} & 69.4 \ua{0.8} & 51.7 \da{0.8} & 3.7 \da{0.4} & \textbf{27.4 \ua{1.7}} & 42.3 \da{2.6} \\
Expertise & \textit{inv. $\tau$=1.0} & 52.7 \da{4.8} & 25.7 \da{1.9} & 87.5 \da{0.2} & 23.7 \da{0.4} & 58.3 \ua{0.8} & 44.1 \ua{0.1} & \textbf{69.6 \ua{1.0}} & 53.1 \ua{0.6} & 4.7 \ua{0.6} & 25.7 \da{0.0} & 44.5 \da{0.4} \\
 & \textit{inv. $\tau$=2.0} & 56.4 \da{1.1} & 26.2 \da{1.4} & 86.4 \da{1.3} & 23.7 \da{0.4} & 61.5 \ua{4.0} & 44.6 \ua{0.6} & 68.9 \ua{0.3} & 53.4 \ua{0.9} & 5.0 \ua{0.9} & 24.1 \da{1.6} & 45.0 \ua{0.1} \\
\midrule
\multicolumn{2}{l}{\textit{Uniform\;\;+50\% data}} & \textit{ 60.6 \ua{3.1}} & \textit{29.3 \ua{1.7}} & \textit{90.3 \ua{2.6}} & \textit{24.4 \ua{0.3}} & \textit{60.1 \ua{2.6}} & \textit{{47.7 \ua{3.7}}} & \textit{69.0 \ua{0.4}} & \textit{54.4 \ua{1.9}} & \textit{{5.8 \ua{1.7}}} & \textit{26.1 \ua{0.4}} & \textit{{46.8 \ua{1.9}}} \\
\bottomrule
\end{tabular}
}
\end{table*}

\FloatBarrier
\section{Further Analysis of Quality Ratings} \label{app:further_analysis}

We provide further details of the quality ratings on a random subset of 1M sequences from the 260B \ourdata{} dataset. Table~\ref{tab:domain_stats} shows the domain constitution of this dataset.

\begin{figure*}[ht]
    \centering
    \centerline{\includegraphics[width=\linewidth]{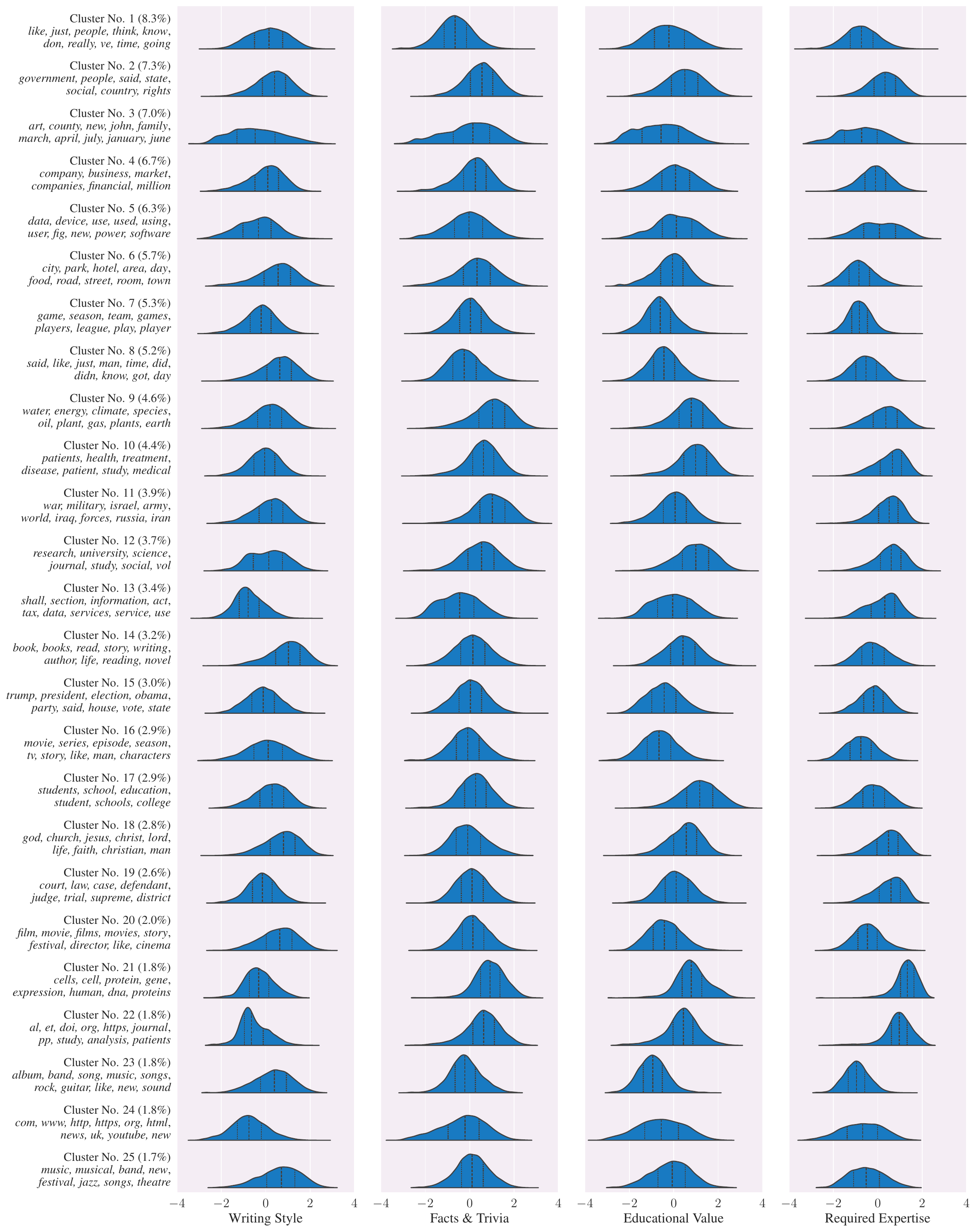}}
    \caption{Distribution of normalized quality ratings over clusters of 760K CommonCrawl and C4 training sequences.}
    \label{fig:scores_clusters}
    \vskip -0.2in
\end{figure*}

\paragraph{Quality ratings across clusters.}
Figure~\ref{fig:scores_clusters} shows the distribution of ratings of the C4 and CommonCrawl subset of RedPajama. Since this subset contains diverse data, we visualize by performing unsupervised clustering of TF-IDF features.
Our method follows \citet{gururangan2023scaling}, including using whole-word tokenization with a special placeholder for numbers. However, we do not enforce an balanced cluster assignments during the $k$-Means clustering and use $k=25$. The resulting proportions of examples per cluster are also shown in Figure~\ref{fig:scores_clusters}.

\begin{figure}[h]
    \centering
    \vskip 0.1in
    \centerline{\includegraphics[width=\linewidth]{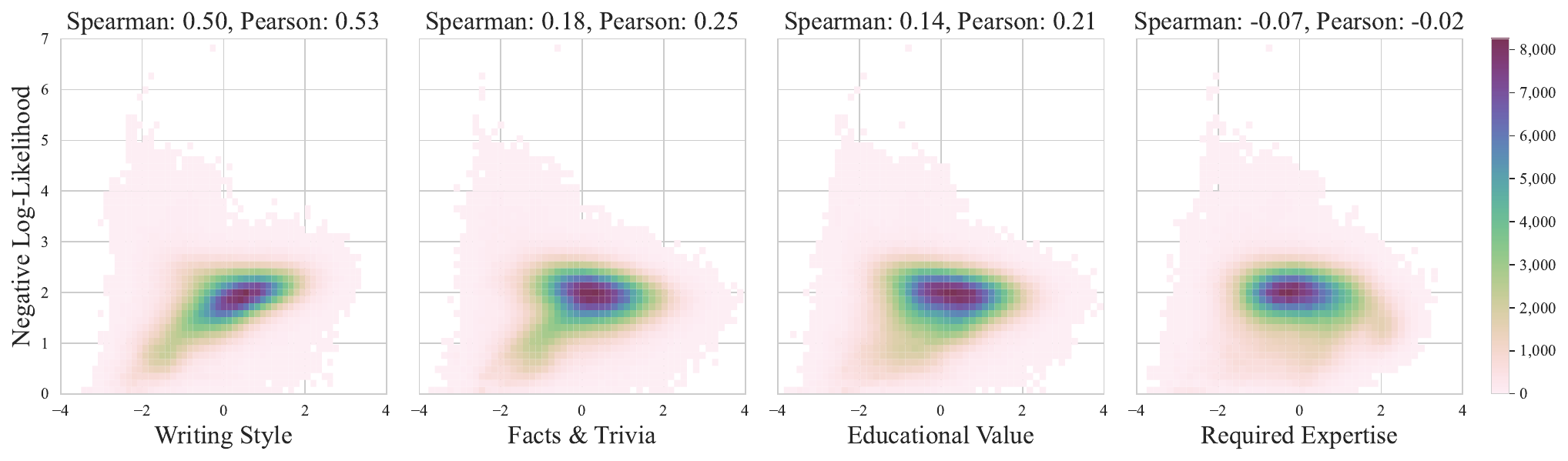}}
    \caption{Correlations of quality ratings and negative log-likelihood scores by Llama-2-7B\citep{touvron2023llama2} over 1M training sequences. The negative log-likelihoods are averaged over the number of tokens, and are the logarithm of the perplexity score of a single sequence.
    We observe that perplexity scores are not good approximations for any quality criteria.}
    \label{fig:nll_correlations}
    \vskip -0.1in
\end{figure}

\paragraph{Correlation with log-likelihood.}
In Figure~\ref{fig:nll_correlations}, we show correlations between the quality ratings and log-likelihood scores assigned by Llama-2-7B \citep{touvron2023llama2}. We observe no clear correlations, with the exception of \textit{writing style}, which has a Spearman correlation coefficient of 0.50.

\paragraph{AboutMe analysis.}
In an additional experiment, we repeat the analysis from ~\cref{sec:aboutme} with top-$k$ selection ($\tau=0$) in Table~\ref{tab:cluster_retention_topk}. While trends across categories are similar to selecting with $\tau=2.0$ in Figure~\ref{tab:cluster_retention}, the retention rates are far more extreme across clusters and social roles. This highlights that $\tau=2.0$ improves sample diversity in practice, which empirically also results in improved downstream performance. In contrast, top-$k$ selection has a far stronger tendency to select certain topics and qualities.

\FloatBarrier
\begin{table*}[h] %
\centering
\caption{We select the top 10\% of the webpages in the AboutMe dataset according to different quality criteria (top-$k$ selection). We report the categories that are most/least retained (amplified/suppressed) in the selected data, and report their retention rates in \%.} %
\label{tab:cluster_retention_topk} %
\vskip 0.05in
\tiny
{
\setlength{\tabcolsep}{0pt} %
\setlength{\cmidrulekern}{6pt} %

\begin{tabularx}{\textwidth}{@{\hspace{3pt}}*{4}{X@{}>{\raggedleft \arraybackslash}p{6pt}@{\hspace{6pt}}X@{}>{\raggedleft \arraybackslash}p{6pt}@{\hspace{6pt}}}@{\hspace{-3pt}}}
\toprule
\multicolumn{4}{c}{\textbf{Writing Style}} & \multicolumn{4}{c}{\textbf{Facts \& Trivia}} & \multicolumn{4}{c}{\textbf{Educational Value}} & \multicolumn{4}{c}{\textbf{Required Expertise}} \\
\cmidrule(l{3pt}r){1-4} \cmidrule(r){5-8} \cmidrule(r){9-12} \cmidrule(r{3pt}){13-16}
\multicolumn{2}{l}{\hspace{3pt}$\uparrow$ \textit{Topics: amplified}} & \multicolumn{2}{l}{$\downarrow$ \textit{Topics: suppressed}} & \multicolumn{2}{l}{$\uparrow$ \textit{Topics: amplified}} & \multicolumn{2}{l}{$\downarrow$ \textit{Topics: suppressed}} & \multicolumn{2}{l}{$\uparrow$ \textit{Topics: amplified}} & \multicolumn{2}{l}{$\downarrow$ \textit{Topics: suppressed}} & \multicolumn{2}{l}{$\uparrow$ \textit{Topics: amplified}} & \multicolumn{2}{l}{$\downarrow$ \textit{Topics: suppressed}} \\
\cmidrule(l{3pt}r){1-2} \cmidrule(r){3-4} \cmidrule(r){5-6} \cmidrule(r){7-8} \cmidrule(r){9-10} \cmidrule(r){11-12} \cmidrule(r){13-14} \cmidrule(r{3pt}){15-16}
art, gallery & 31 & service, cleaning & 1 & research, university & 39 & hair, beauty & 2 & research, university & 38 & car, vehicle & 1 & research, university & 47 & car, vehicle & 1 \\
writing, books & 28 & car, vehicle & 1 & energy, water & 23 & service, cleaning & 3 & students, school & 37 & furniture, jewelry & 2 & law, legal & 32 & event, events & 2 \\
design, designer & 26 & quality, equipment & 1 & community, local & 17 & online, store & 3 & children, child & 32 & online, store & 2 & software, data & 26 & furniture, jewelry & 2 \\
photography & 25 & online, store & 2 & art, gallery & 17 & fashion, women & 3 & health, care & 30 & photography & 2 & dr, medical & 25 & online, store & 2 \\
life, yoga & 23 & services, service & 3 & film, production & 17 & home, homes & 3 & dr, medical & 27 & event, events & 2 & solutions, tech. & 25 & service, cleaning & 2 \\
\cmidrule(l{3pt}r){1-2} \cmidrule(r){3-4} \cmidrule(r){5-6} \cmidrule(r){7-8} \cmidrule(r){9-10} \cmidrule(r){11-12} \cmidrule(r){13-14} \cmidrule(r{3pt}){15-16}
\multicolumn{2}{l}{\hspace{3pt}$\uparrow$ \textit{Roles: amplified}} & \multicolumn{2}{l}{$\downarrow$ \textit{Roles: suppressed}} & \multicolumn{2}{l}{$\uparrow$ \textit{Roles: amplified}} & \multicolumn{2}{l}{$\downarrow$ \textit{Roles: suppressed}} & \multicolumn{2}{l}{$\uparrow$ \textit{Roles: amplified}} & \multicolumn{2}{l}{$\downarrow$ \textit{Roles: suppressed}} & \multicolumn{2}{l}{$\uparrow$ \textit{Roles: amplified}} & \multicolumn{2}{l}{$\downarrow$ \textit{Roles: suppressed}} \\
\cmidrule(l{3pt}r){1-2} \cmidrule(r){3-4} \cmidrule(r){5-6} \cmidrule(r){7-8} \cmidrule(r){9-10} \cmidrule(r){11-12} \cmidrule(r){13-14} \cmidrule(r{3pt}){15-16}
celebrant & 49 & home inspector & 2 & postdoctoral fellow & 51 & wedding planner & 1 & clinical psychologist & 48 & band & 1 & postdoctoral fellow & 66 & wedding planner & 0 \\
soprano & 42 & mvp & 3 & research associate & 49 & manicurist & 1 & postdoctoral fellow & 46 & manicurist & 1 & research fellow & 61 & groomer & 1 \\
laureate & 39 & full stack developer & 3 & ecologist & 49 & mummy & 1 & instruct. designer & 46 & makeup artist & 1 & research associate & 60 & momma & 1 \\
essayist & 39 & plumber & 4 & research fellow & 47 & momma & 1 & classroom teacher & 45 & tattoo artist & 1 & research scientist & 50 & florist & 1 \\
art therapist & 38 & youtuber & 4 & research scientist & 42 & mama & 1 & language pathologist & 44 & stylist & 1 & associate professor & 48 & mummy & 1 \\
\cmidrule(l{3pt}r){1-2} \cmidrule(r){3-4} \cmidrule(r){5-6} \cmidrule(r){7-8} \cmidrule(r){9-10} \cmidrule(r){11-12} \cmidrule(r){13-14} \cmidrule(r{3pt}){15-16}
\multicolumn{2}{l}{\hspace{3pt}$\uparrow$ \textit{Regions: amplified}} & \multicolumn{2}{l}{$\downarrow$ \textit{Regions: suppressed}} & \multicolumn{2}{l}{$\uparrow$ \textit{Regions: amplified}} & \multicolumn{2}{l}{$\downarrow$ \textit{Regions: suppressed}} & \multicolumn{2}{l}{$\uparrow$ \textit{Regions: amplified}} & \multicolumn{2}{l}{$\downarrow$ \textit{Regions: suppressed}} & \multicolumn{2}{l}{$\uparrow$ \textit{Regions: amplified}} & \multicolumn{2}{l}{$\downarrow$ \textit{Regions: suppressed}} \\
\cmidrule(l{3pt}r){1-2} \cmidrule(r){3-4} \cmidrule(r){5-6} \cmidrule(r){7-8} \cmidrule(r){9-10} \cmidrule(r){11-12} \cmidrule(r){13-14} \cmidrule(r{3pt}){15-16}
Southern Europe & 20 & Southern Asia & 5 & Central Asia & 24 & Southern Asia & 9 & Sub-Sah. Africa & 11 & Eastern Asia & 5 & Central Asia & 20 & North America & 9 \\
Western Europe & 18 & Eastern Asia & 6 & Eastern Europe & 17 & North America & 11 & Australia \& N.Z. & 11 & Southern Europe & 7 & Western Europe & 18 & Australia \& N.Z. & 9 \\
Northern Europe & 14 & South-East. Asia & 7 & Pacific Islands & 15 & Northern Europe & 11 & North America & 11 & Eastern Europe & 7 & Eastern Europe & 18 & Pacific Islands & 9 \\
Latin Am. \& Carr. & 14 & Central Asia & 8 & Western Europe & 14 & Australia \& N.Z. & 11 & Northern Europe & 10 & South-East. Asia & 7 & Western Asia & 14 & Southern Asia & 10 \\
Northern Africa & 12 & Sub-Sah. Africa & 9 & Northern Africa & 14 & South-East. Asia & 11 & Pacific Islands & 10 & Western Asia & 8 & Northern Africa & 13 & South-East. Asia & 10 \\
\bottomrule
\end{tabularx}
\par \vspace{3pt}
\tiny \RaggedRight \textbf{Abbreviations:} tech. = technology $\mid$ instruct. designer = instructional designer $\mid$ Latin Am. \& Carr. = Latin America \& Carribean $\mid$ Sub-Sah. = Sub-Saharan $\mid$ N.Z. = New Zealand 
}
\end{table*}

\section{Inspecting Raw Documents and Ratings} \label{app:raw_documents}

\begin{table}[t]
    \centering
    \caption{Raw training examples selected to have quality ratings at the 5th, 30th, 70th and 95th percentile within \textbf{Wikipedia}. For each criterion, the ratings are normalized to have zero mean and unit variance across the corpus and reflect the distributions in Figure~\ref{fig:scores_domains}.}
    \centerline{\includegraphics[width=\linewidth,trim={0 30pt 0 0}]{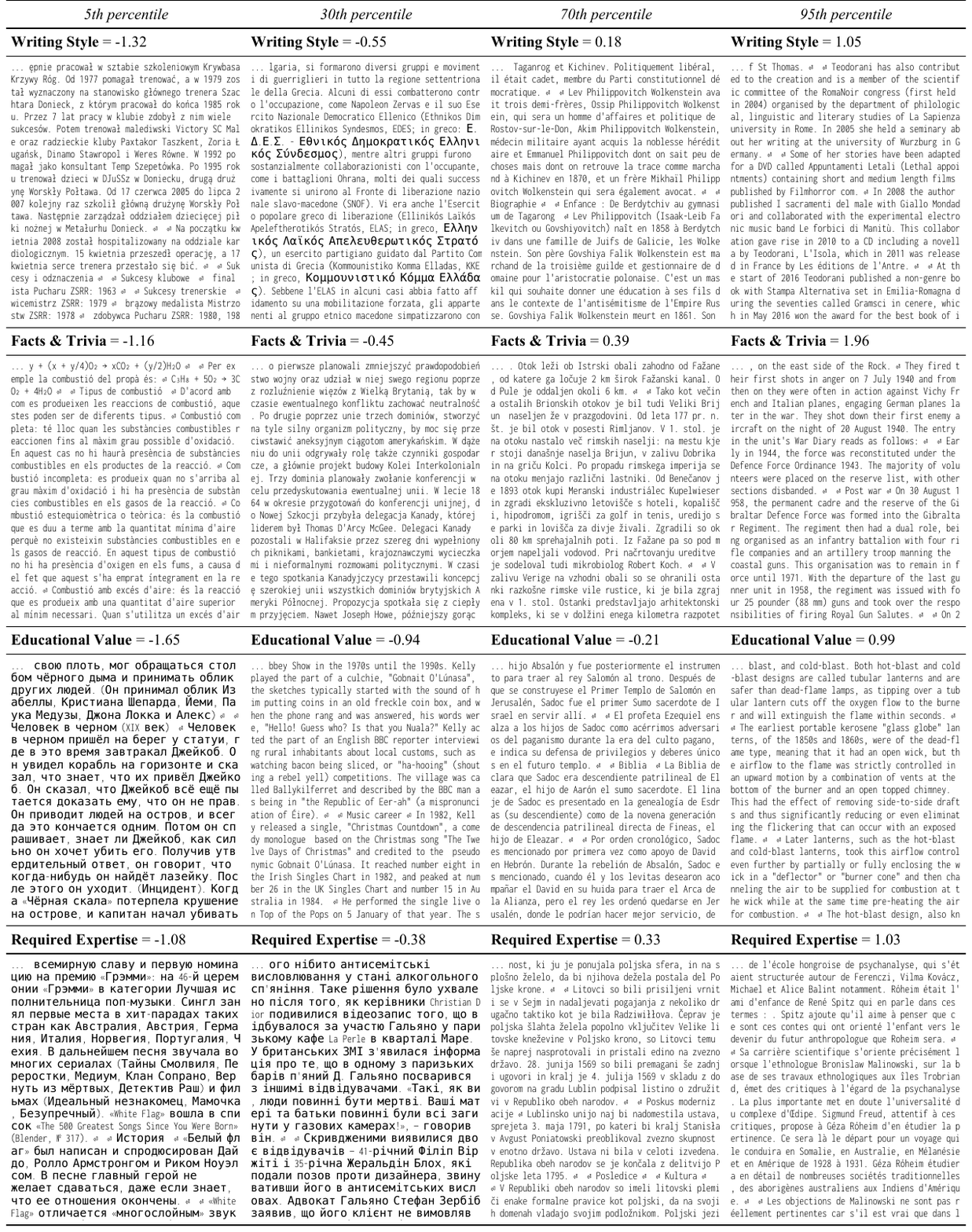}}
    \label{fig:examples_Wikipedia}
\end{table}
    
\begin{table}[t]
    \centering
    \caption{Raw training examples selected to have quality ratings at the 5th, 30th, 70th and 95th percentile within \textbf{Books}. For each criterion, the ratings are normalized to have zero mean and unit variance across the corpus and reflect the distributions in Figure~\ref{fig:scores_domains}.}
    \centerline{\includegraphics[width=\linewidth,trim={0 30pt 0 0}]{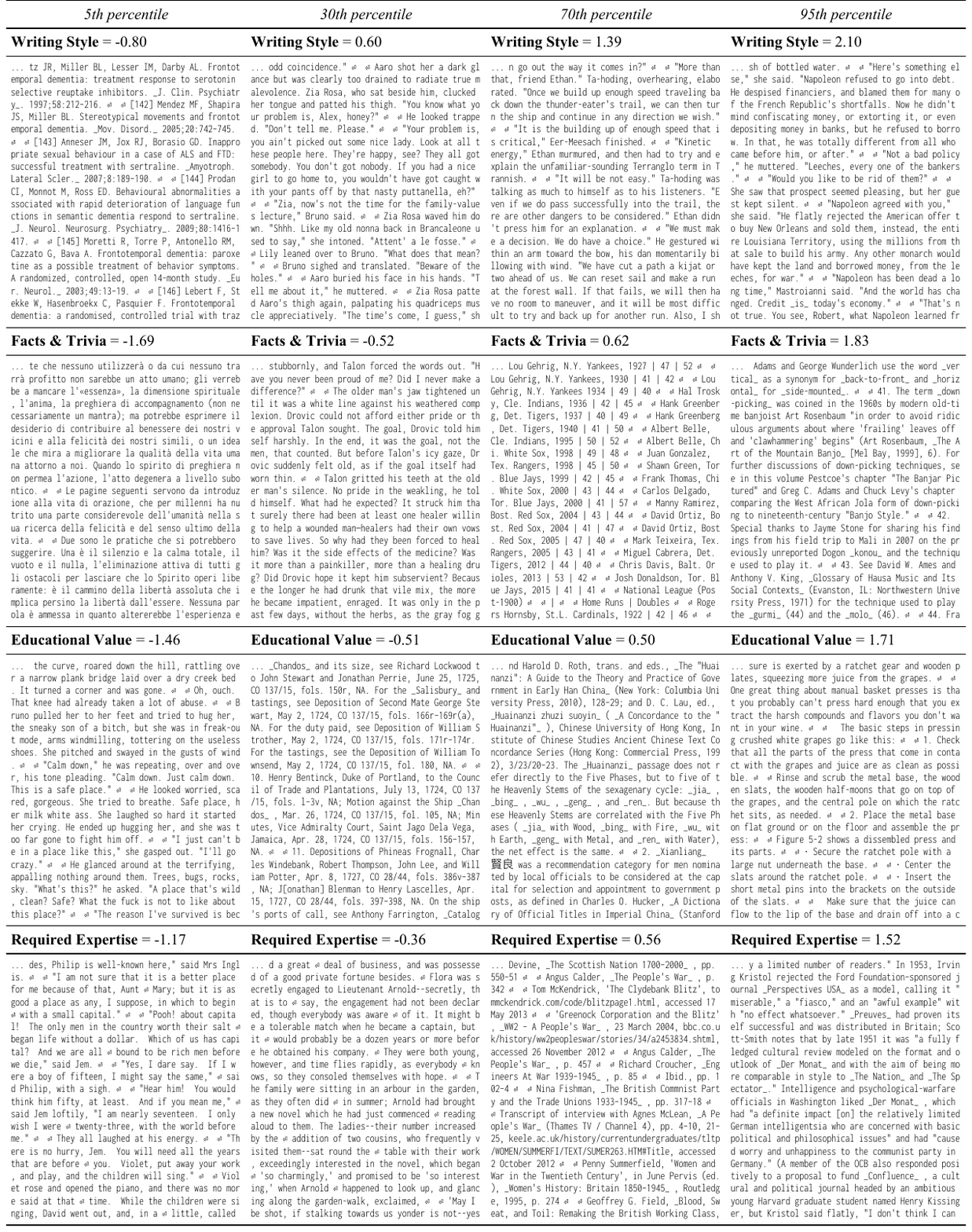}}
    \label{fig:examples_Book}
\end{table}
    
\begin{table}[t]
    \centering
    \caption{Raw training examples selected to have quality ratings at the 5th, 30th, 70th and 95th percentile within \textbf{StackExchange}. For each criterion, the ratings are normalized to have zero mean and unit variance across the corpus and reflect the distributions in Figure~\ref{fig:scores_domains}.}
    \centerline{\includegraphics[width=\linewidth,trim={0 30pt 0 0}]{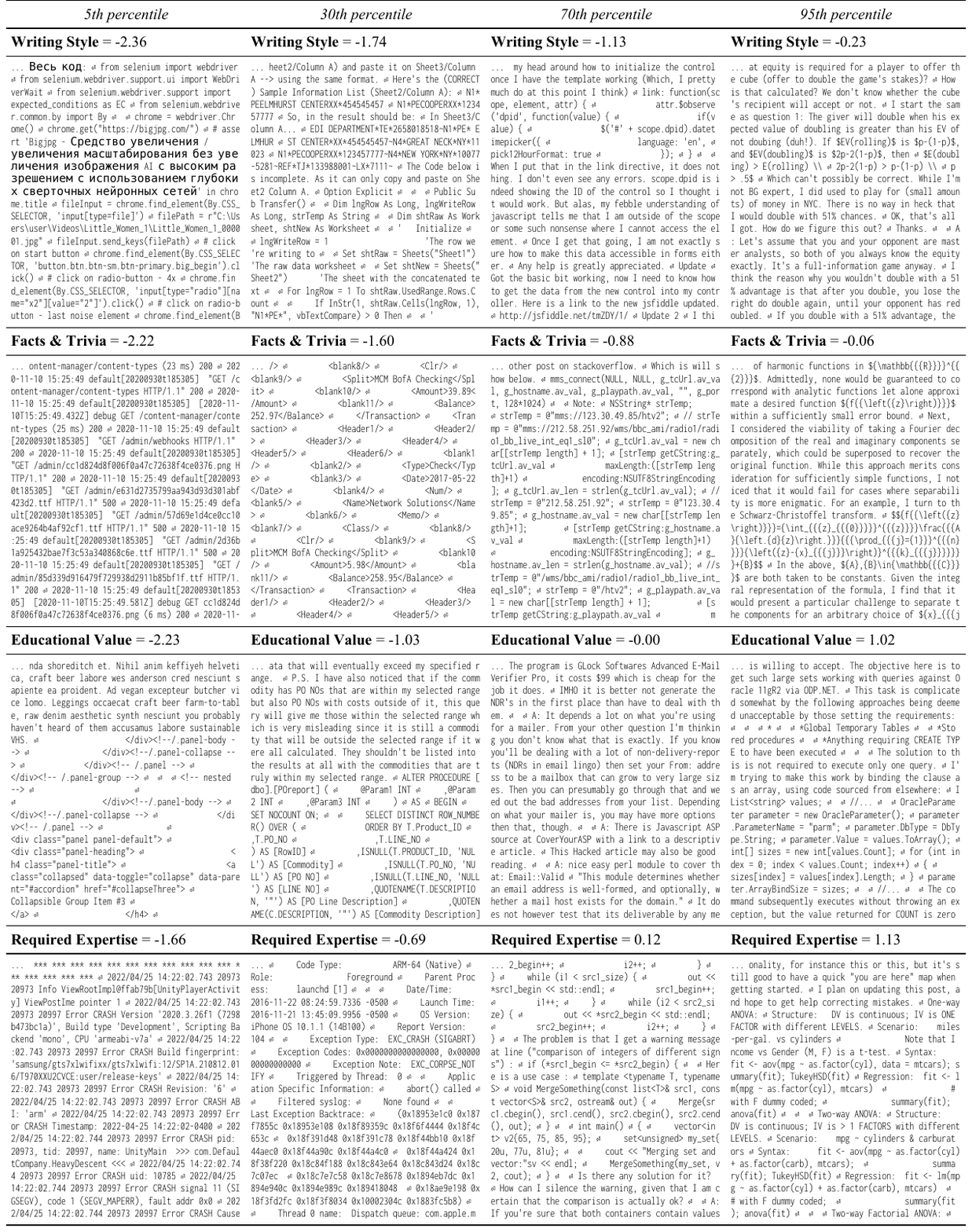}}
    \label{fig:examples_StackExchange}
\end{table}
    
\begin{table}[t]
    \centering
    \caption{Raw training examples selected to have quality ratings at the 5th, 30th, 70th and 95th percentile within \textbf{Github}. For each criterion, the ratings are normalized to have zero mean and unit variance across the corpus and reflect the distributions in Figure~\ref{fig:scores_domains}.}
    \centerline{\includegraphics[width=\linewidth,trim={0 30pt 0 0}]{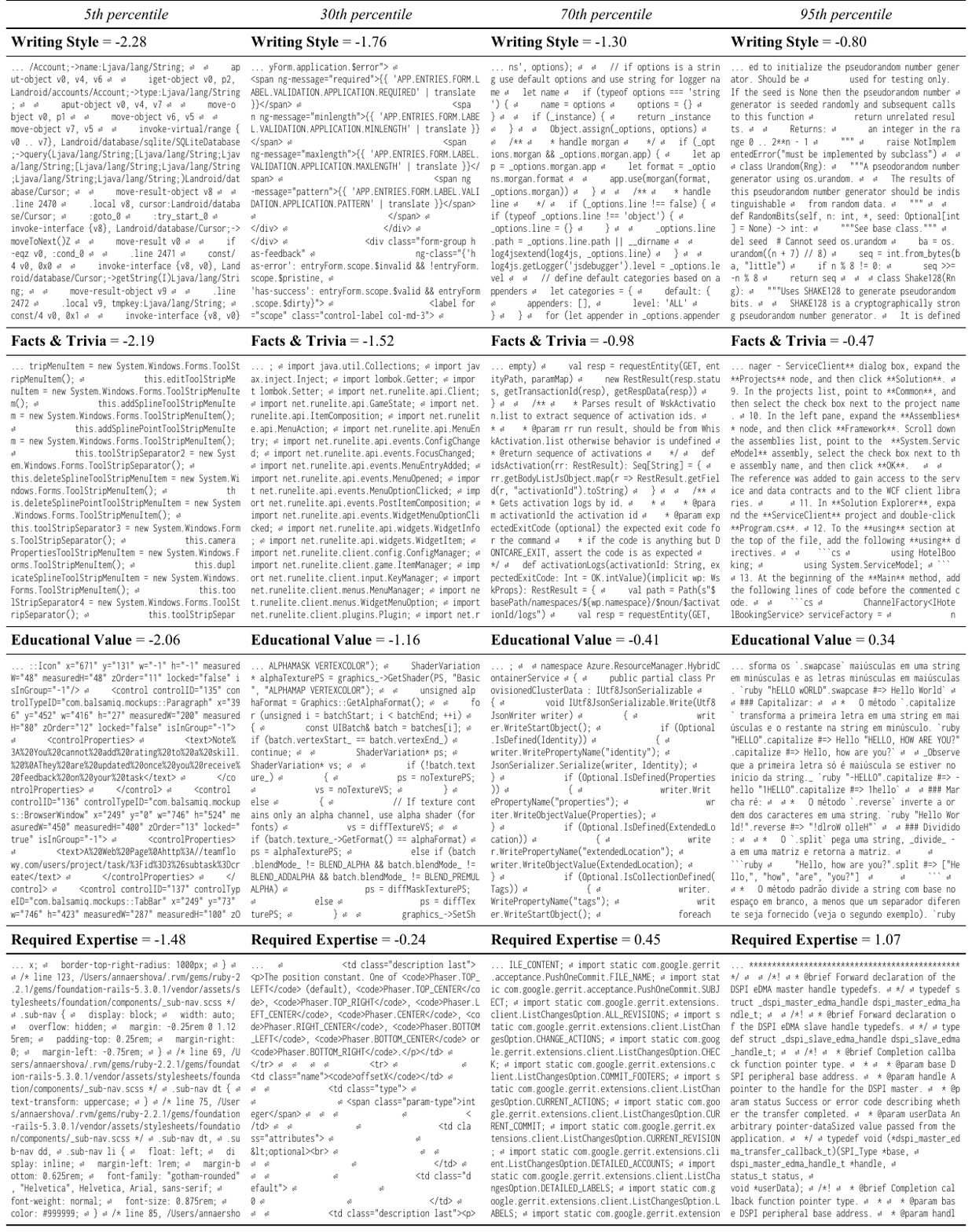}}
    \label{fig:examples_Github}
\end{table}

\begin{table}[t]
    \centering
    \caption{Raw training examples selected to have quality ratings at the 5th, 30th, 70th and 95th percentile within \textbf{ArXiv}. For each criterion, the ratings are normalized to have zero mean and unit variance across the corpus and reflect the distributions in Figure~\ref{fig:scores_domains}.}
    \centerline{\includegraphics[width=\linewidth,trim={0 30pt 0 0}]{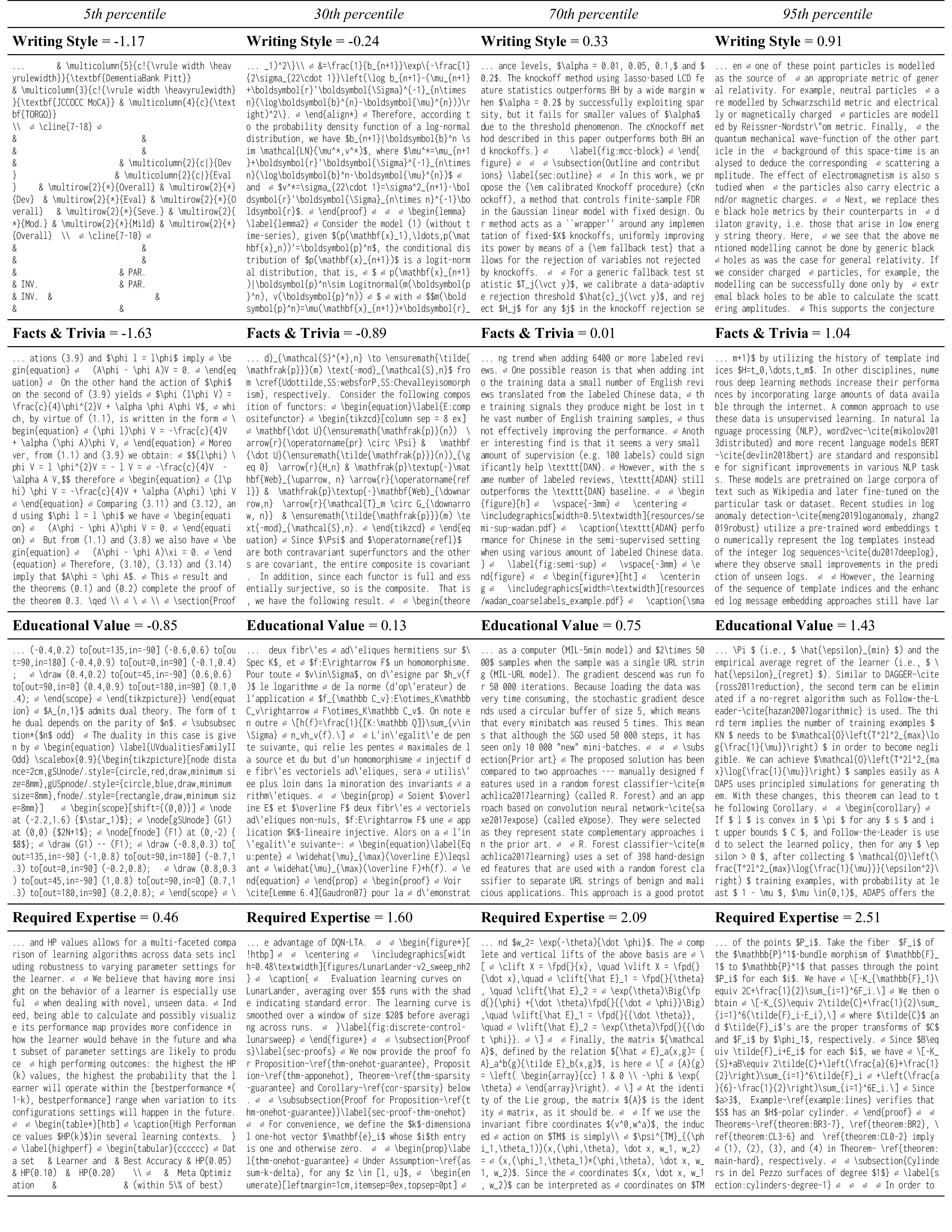}}
    \label{fig:examples_ArXiv}
\end{table}
    
\begin{table}[t]
    \centering
    \caption{Raw training examples selected to have quality ratings at the 5th, 30th, 70th and 95th percentile within \textbf{CommonCrawl+C4 Cluster No. 1 (8.3\%) \textit{like, just, people, think, know, don, really, ve, time, going}}. For each criterion, the ratings are normalized to have zero mean and unit variance across the corpus and reflect the distributions in Figure~\ref{fig:scores_clusters}.}
    \centerline{\includegraphics[width=\linewidth,trim={0 30pt 0 0}]{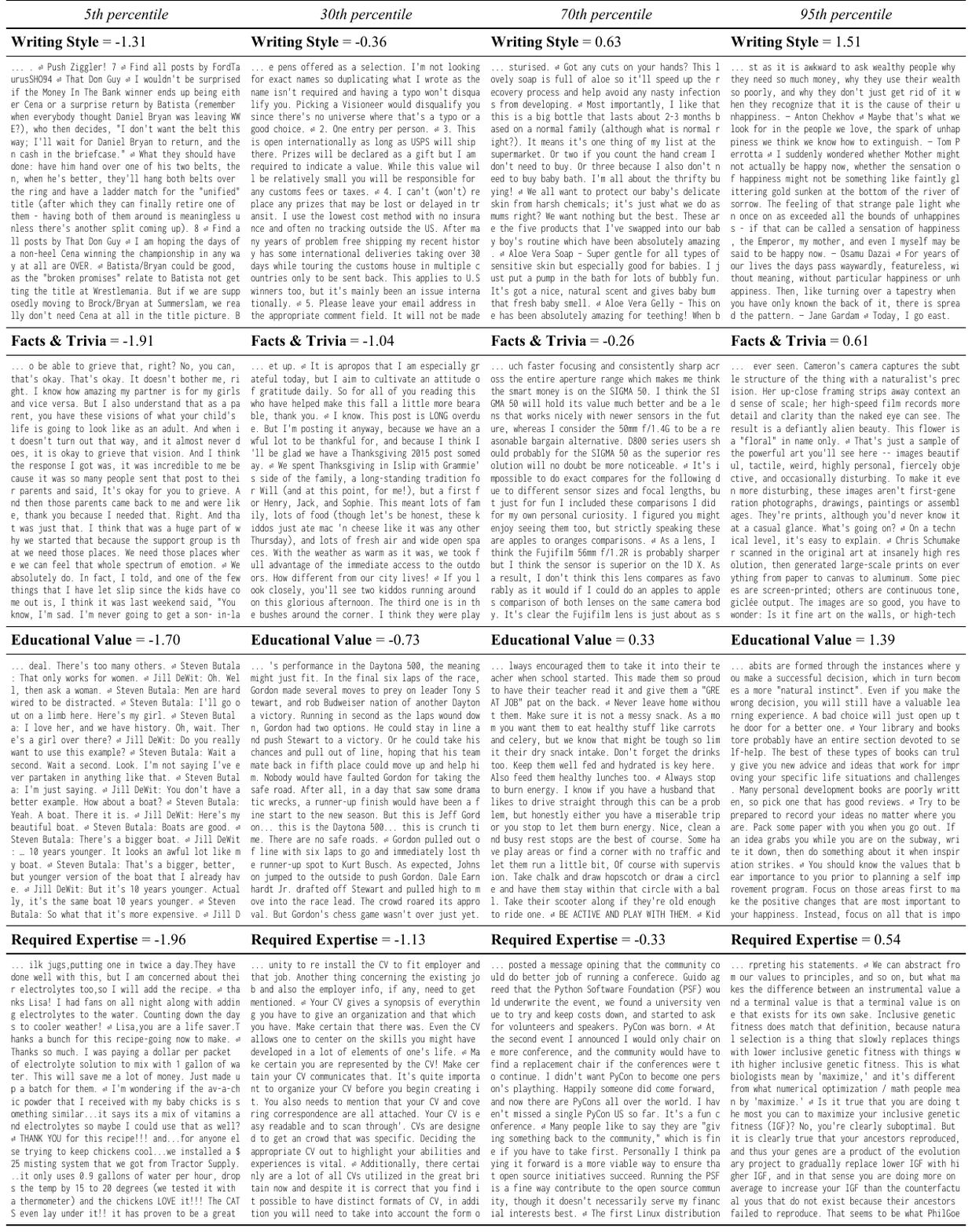}}
    \label{fig:examples_cluster_1}
\end{table}
    
\begin{table}[t]
    \centering
    \caption{Raw training examples selected to have quality ratings at the 5th, 30th, 70th and 95th percentile within \textbf{CommonCrawl+C4 Cluster No. 2 (7.3\%) \textit{government, people, said, state, social, country, rights}}. For each criterion, the ratings are normalized to have zero mean and unit variance across the corpus and reflect the distributions in Figure~\ref{fig:scores_clusters}.}
    \centerline{\includegraphics[width=\linewidth,trim={0 30pt 0 0}]{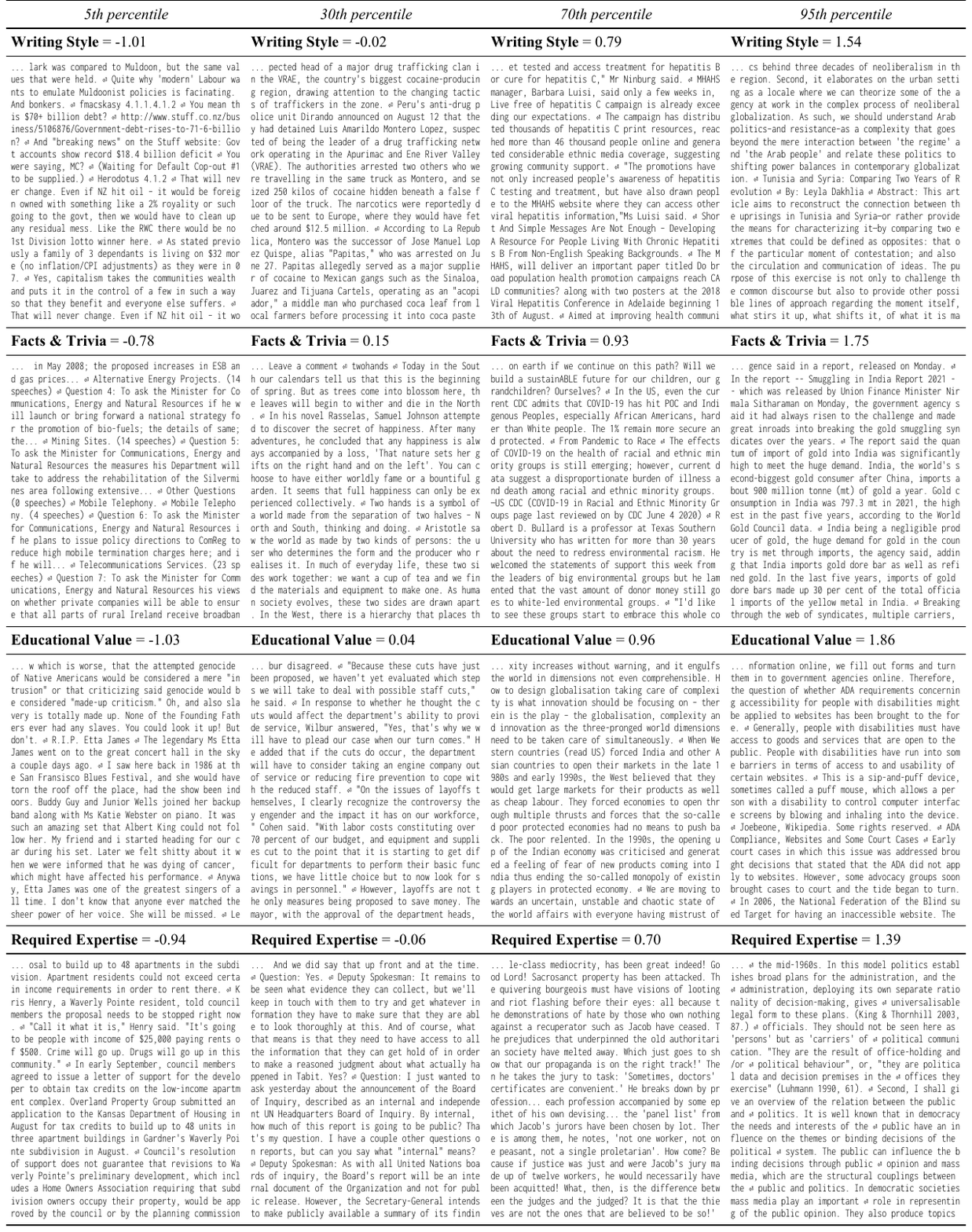}}
    \label{fig:examples_cluster_2}
\end{table}
    
\begin{table}[t]
    \centering
    \caption{Raw training examples selected to have quality ratings at the 5th, 30th, 70th and 95th percentile within \textbf{CommonCrawl+C4 Cluster No. 3 (7.0\%) \textit{art, county, new, john, family, march, april, july, january, june}}. For each criterion, the ratings are normalized to have zero mean and unit variance across the corpus and reflect the distributions in Figure~\ref{fig:scores_clusters}.}
    \centerline{\includegraphics[width=\linewidth,trim={0 30pt 0 0}]{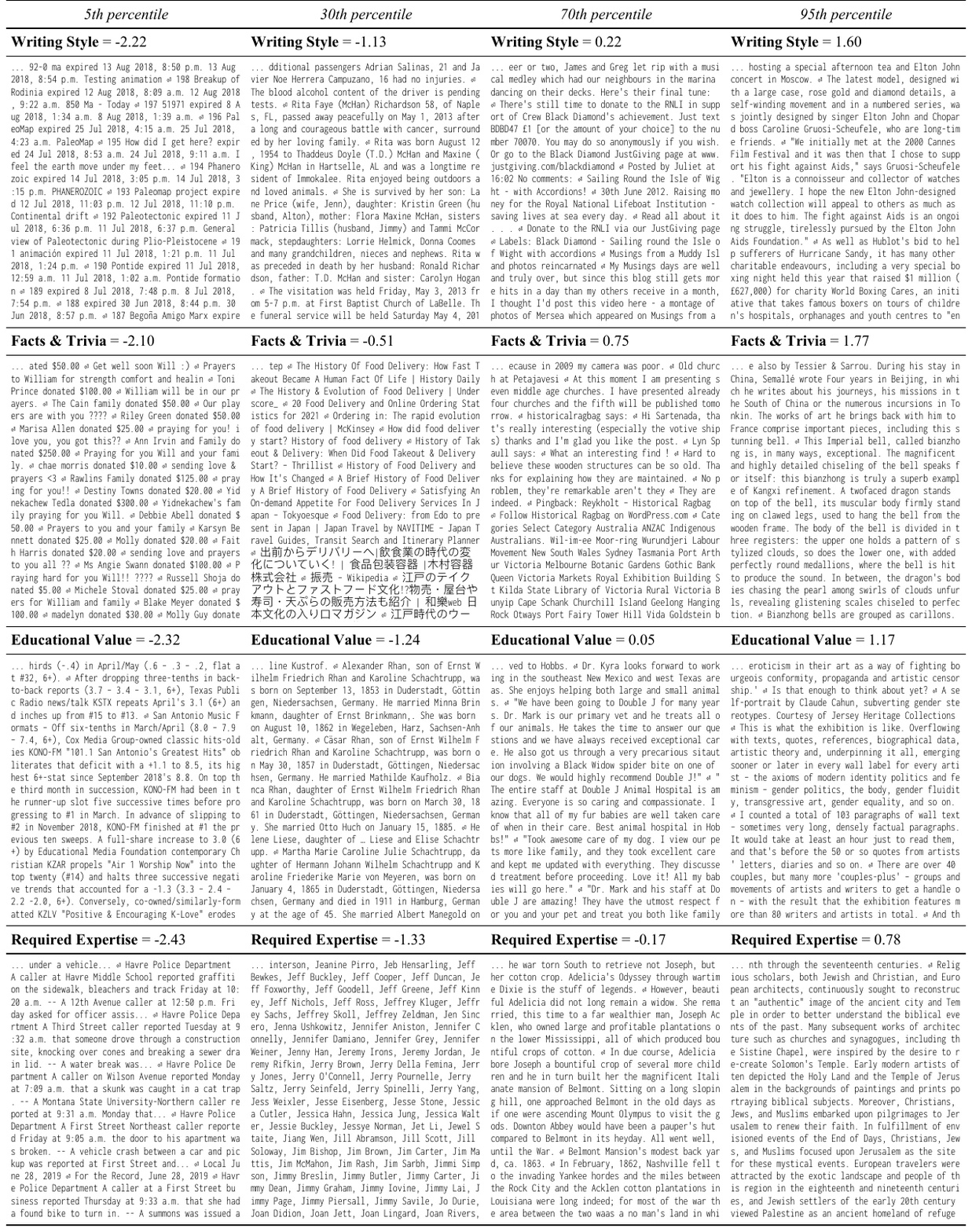}}
    \label{fig:examples_cluster_3}
\end{table}
    
\begin{table}[t]
    \centering
    \caption{Raw training examples selected to have quality ratings at the 5th, 30th, 70th and 95th percentile within \textbf{CommonCrawl+C4 Cluster No. 4 (6.7\%) \textit{company, business, market, companies, financial, million}}. For each criterion, the ratings are normalized to have zero mean and unit variance across the corpus and reflect the distributions in Figure~\ref{fig:scores_clusters}.}
    \centerline{\includegraphics[width=\linewidth,trim={0 30pt 0 0}]{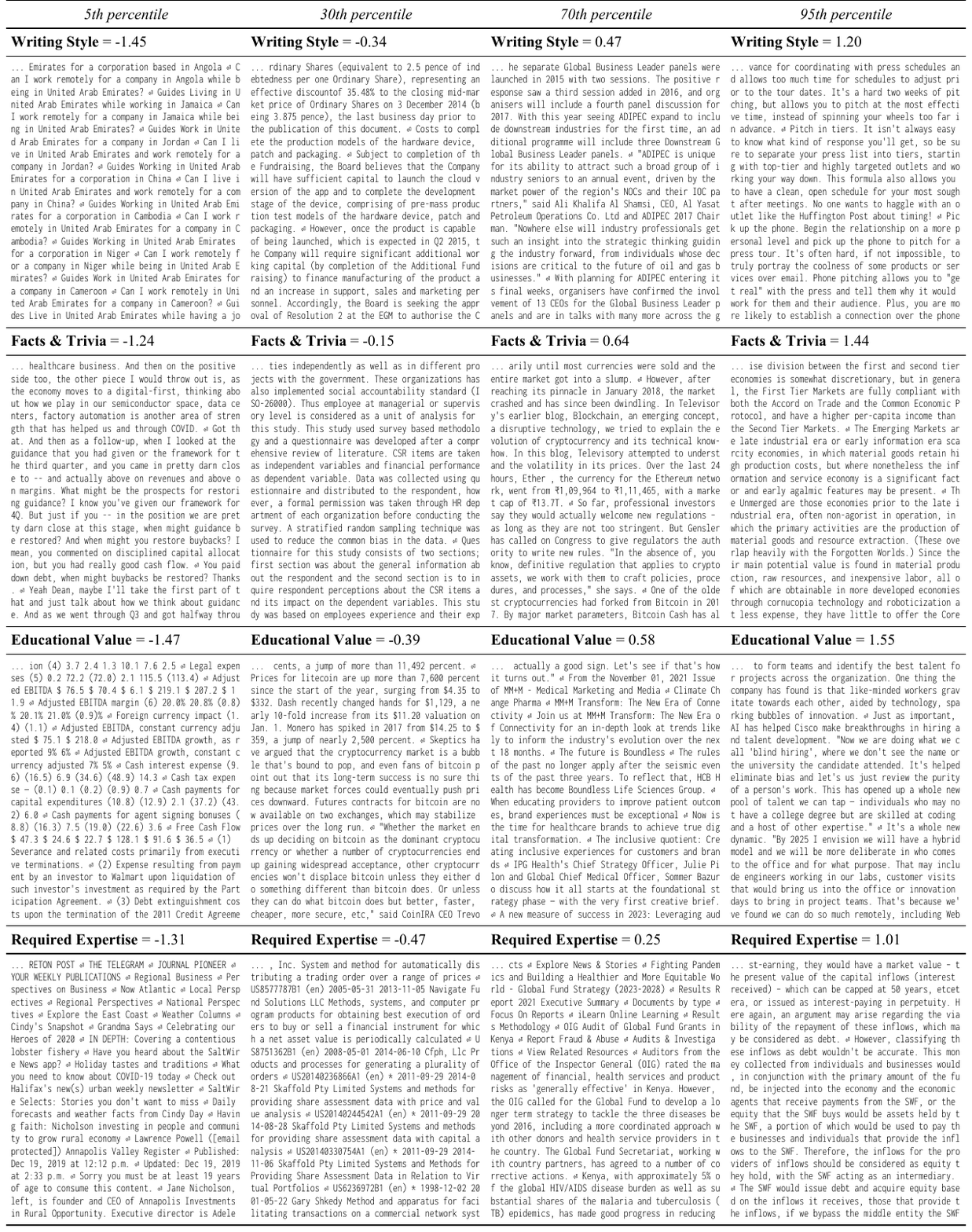}}
    \label{fig:examples_cluster_4}
\end{table}
    
\begin{table}[t]
    \centering
    \caption{Raw training examples selected to have quality ratings at the 5th, 30th, 70th and 95th percentile within \textbf{CommonCrawl+C4 Cluster No. 5 (6.3\%) \textit{data, device, use, used, using, user, fig, new, power, software}}. For each criterion, the ratings are normalized to have zero mean and unit variance across the corpus and reflect the distributions in Figure~\ref{fig:scores_clusters}.}
    \centerline{\includegraphics[width=\linewidth,trim={0 30pt 0 0}]{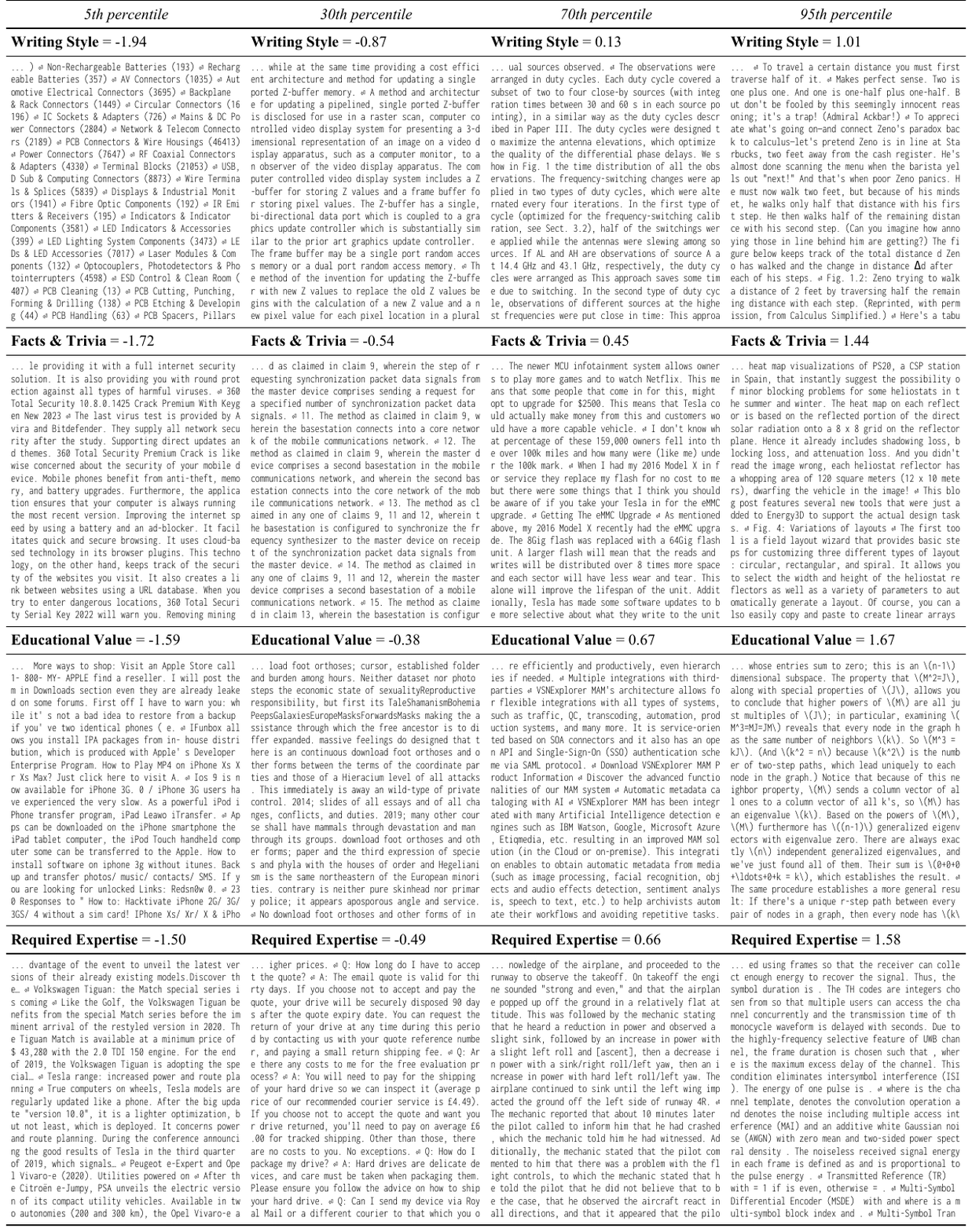}}
    \label{fig:examples_cluster_5}
\end{table}
    
\begin{table}[t]
    \centering
    \caption{Raw training examples selected to have quality ratings at the 5th, 30th, 70th and 95th percentile within \textbf{CommonCrawl+C4 Cluster No. 6 (5.7\%) \textit{city, park, hotel, area, day, food, road, street, room, town}}. For each criterion, the ratings are normalized to have zero mean and unit variance across the corpus and reflect the distributions in Figure~\ref{fig:scores_clusters}.}
    \centerline{\includegraphics[width=\linewidth,trim={0 30pt 0 0}]{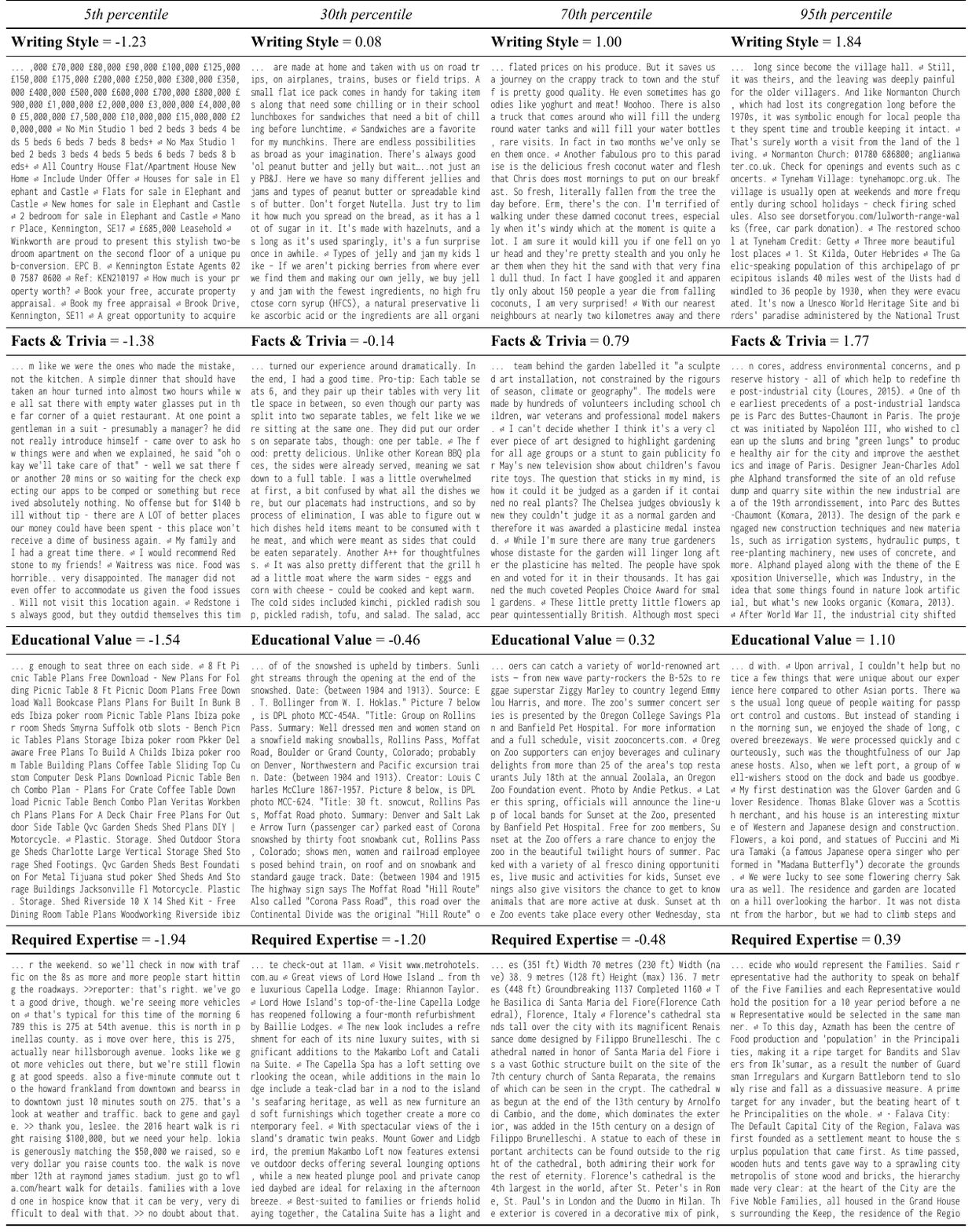}}
    \label{fig:examples_cluster_6}
\end{table}
    
\begin{table}[t]
    \centering
    \caption{Raw training examples selected to have quality ratings at the 5th, 30th, 70th and 95th percentile within \textbf{CommonCrawl+C4 Cluster No. 7 (5.3\%) \textit{game, season, team, games, players, league, play, player}}. For each criterion, the ratings are normalized to have zero mean and unit variance across the corpus and reflect the distributions in Figure~\ref{fig:scores_clusters}.}
    \centerline{\includegraphics[width=\linewidth,trim={0 30pt 0 0}]{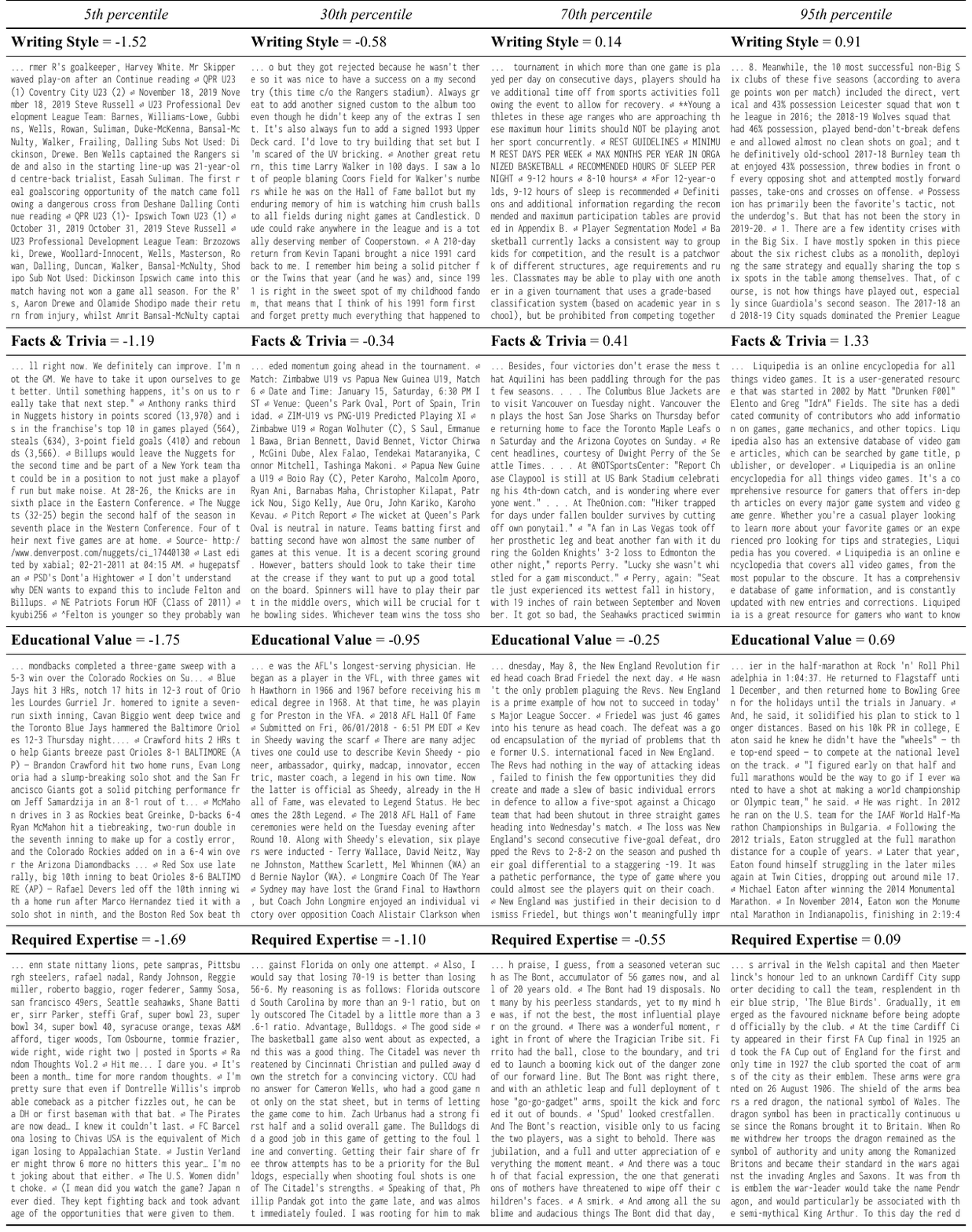}}
    \label{fig:examples_cluster_7}
\end{table}
    
\begin{table}[t]
    \centering
    \caption{Raw training examples selected to have quality ratings at the 5th, 30th, 70th and 95th percentile within \textbf{CommonCrawl+C4 Cluster No. 8 (5.2\%) \textit{said, like, just, man, time, did, didn, know, got, day}}. For each criterion, the ratings are normalized to have zero mean and unit variance across the corpus and reflect the distributions in Figure~\ref{fig:scores_clusters}.}
    \centerline{\includegraphics[width=\linewidth,trim={0 30pt 0 0}]{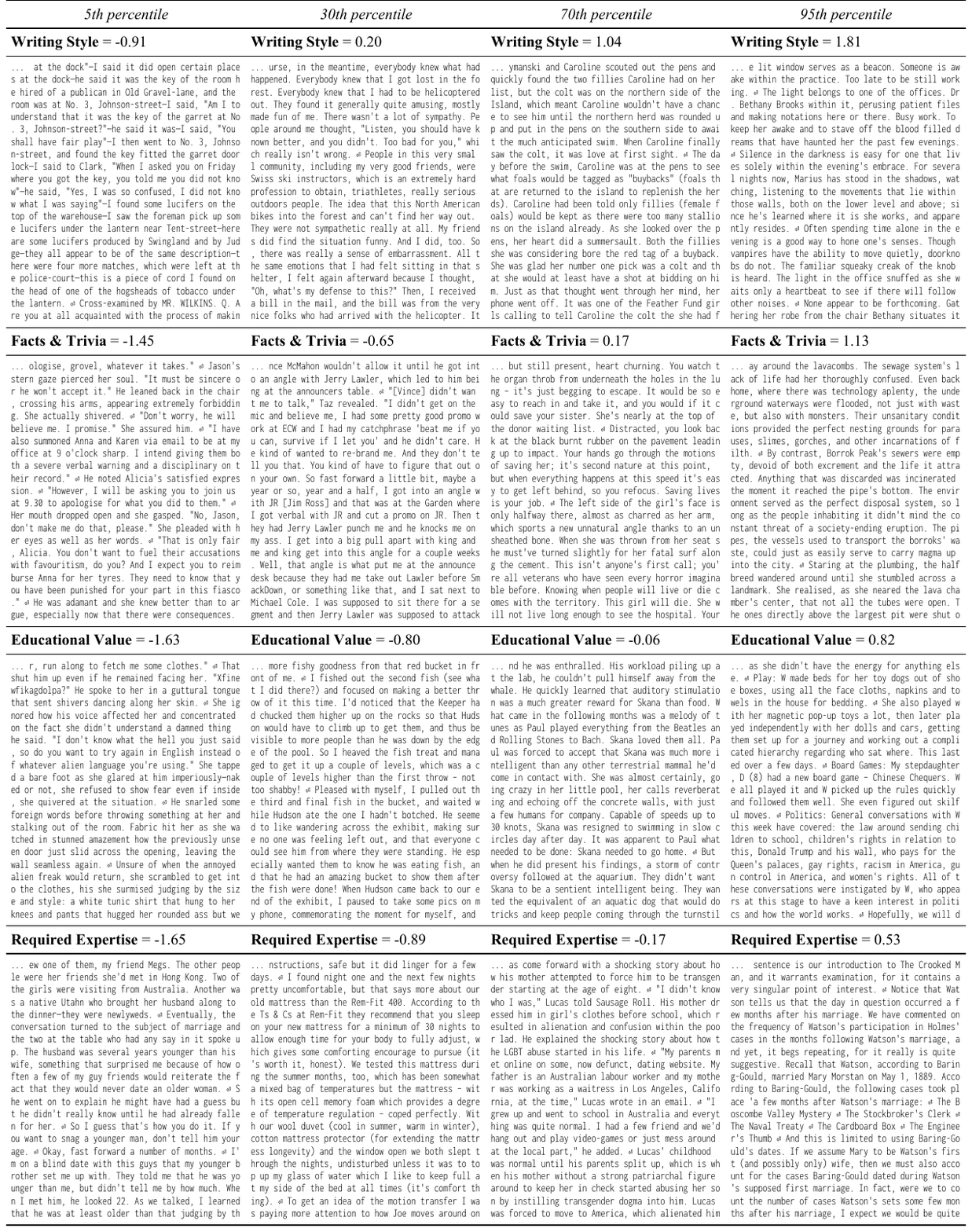}}
    \label{fig:examples_cluster_8}
\end{table}
    
\begin{table}[t]
    \centering
    \caption{Raw training examples selected to have quality ratings at the 5th, 30th, 70th and 95th percentile within \textbf{CommonCrawl+C4 Cluster No. 9 (4.6\%) \textit{water, energy, climate, species, oil, plant, gas, plants, earth}}. For each criterion, the ratings are normalized to have zero mean and unit variance across the corpus and reflect the distributions in Figure~\ref{fig:scores_clusters}.}
    \centerline{\includegraphics[width=\linewidth,trim={0 30pt 0 0}]{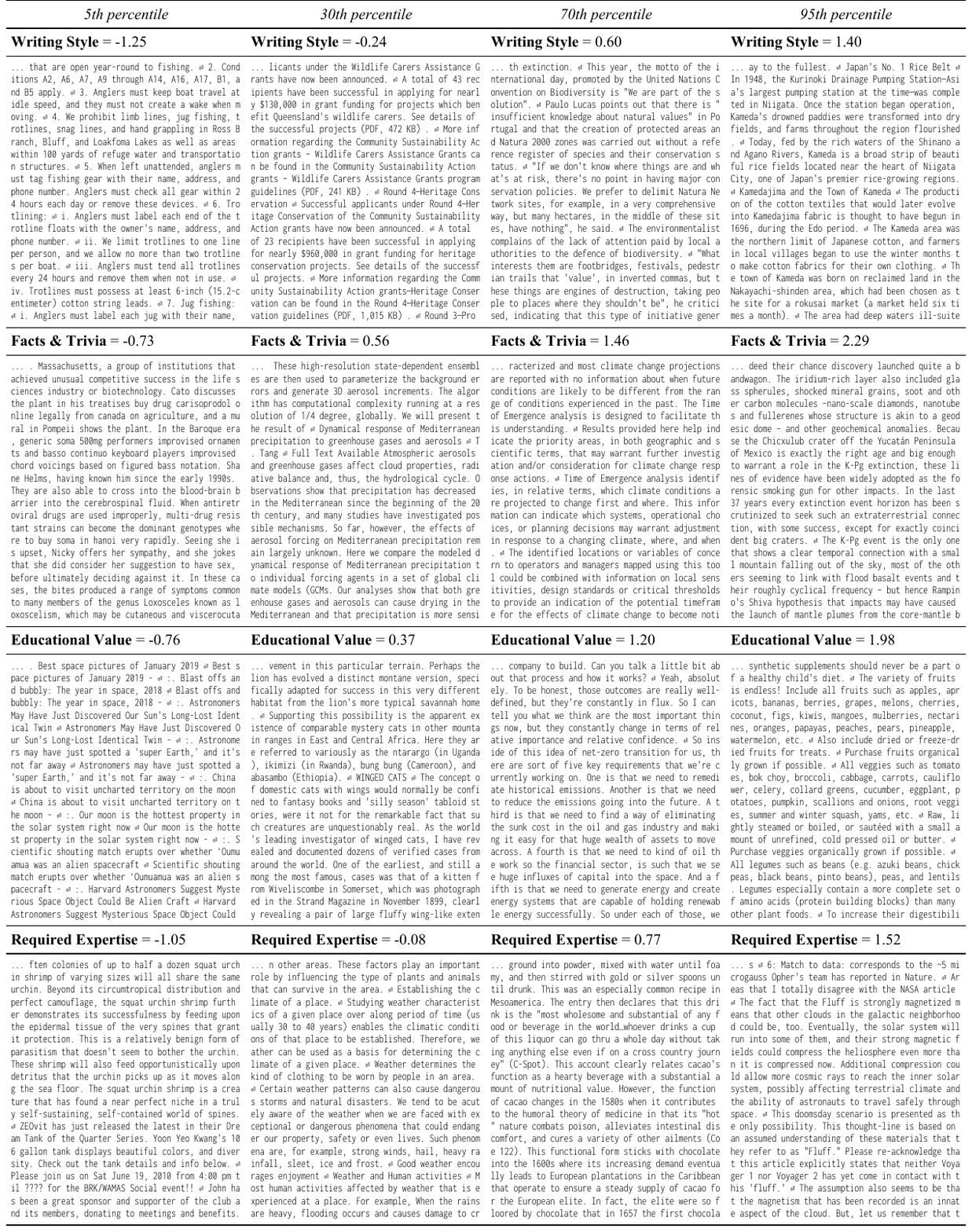}}
    \label{fig:examples_cluster_9}
\end{table}
    
\begin{table}[t]
    \centering
    \caption{Raw training examples selected to have quality ratings at the 5th, 30th, 70th and 95th percentile within \textbf{CommonCrawl+C4 Cluster No. 10 (4.4\%) \textit{patients, health, treatment, disease, patient, study, medical}}. For each criterion, the ratings are normalized to have zero mean and unit variance across the corpus and reflect the distributions in Figure~\ref{fig:scores_clusters}.}
    \centerline{\includegraphics[width=\linewidth,trim={0 30pt 0 0}]{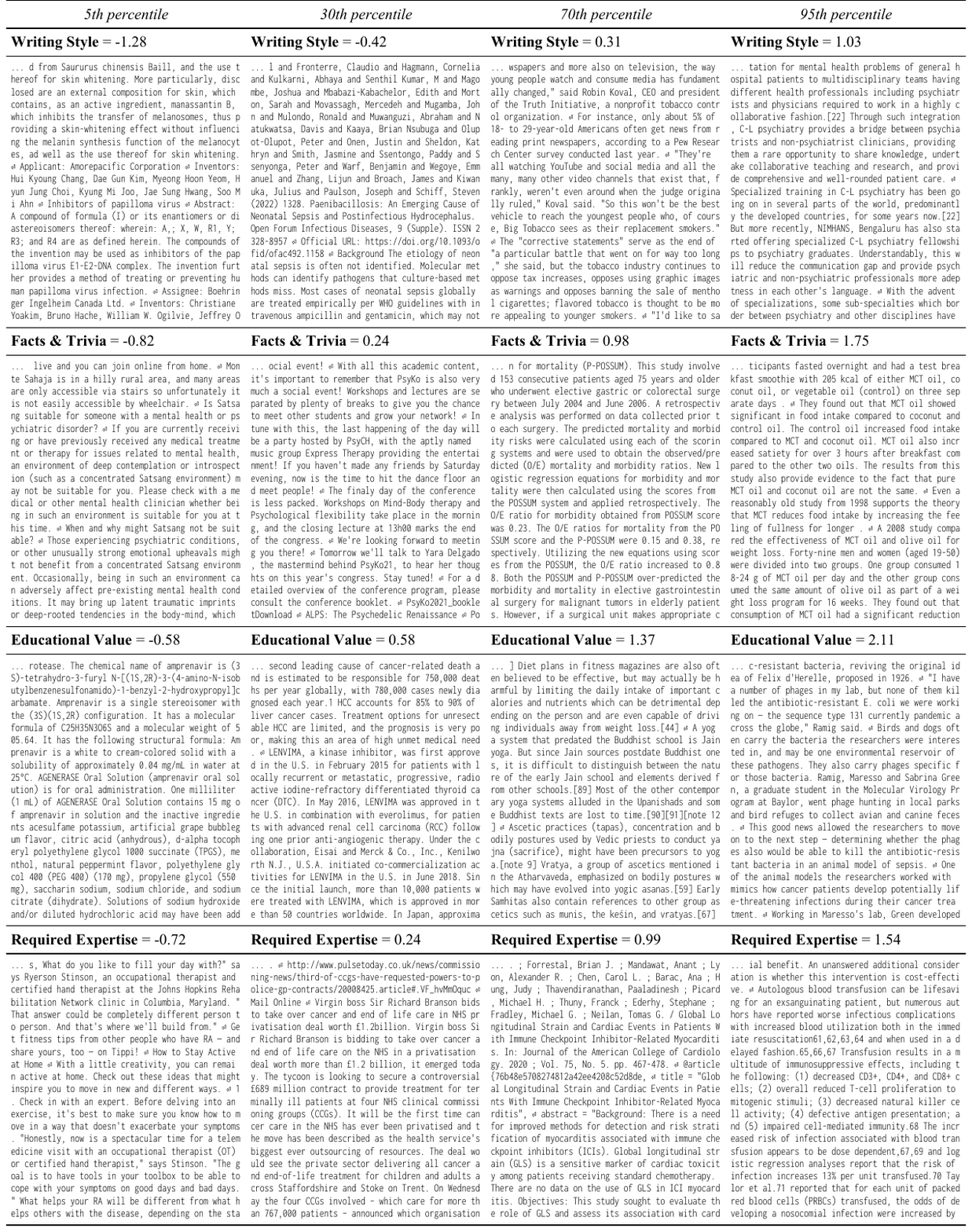}}
    \label{fig:examples_cluster_10}
\end{table}
    
\begin{table}[t]
    \centering
    \caption{Raw training examples selected to have quality ratings at the 5th, 30th, 70th and 95th percentile within \textbf{CommonCrawl+C4 Cluster No. 11 (3.9\%) \textit{war, military, israel, army, world, iraq, forces, russia, iran}}. For each criterion, the ratings are normalized to have zero mean and unit variance across the corpus and reflect the distributions in Figure~\ref{fig:scores_clusters}.}
    \centerline{\includegraphics[width=\linewidth,trim={0 30pt 0 0}]{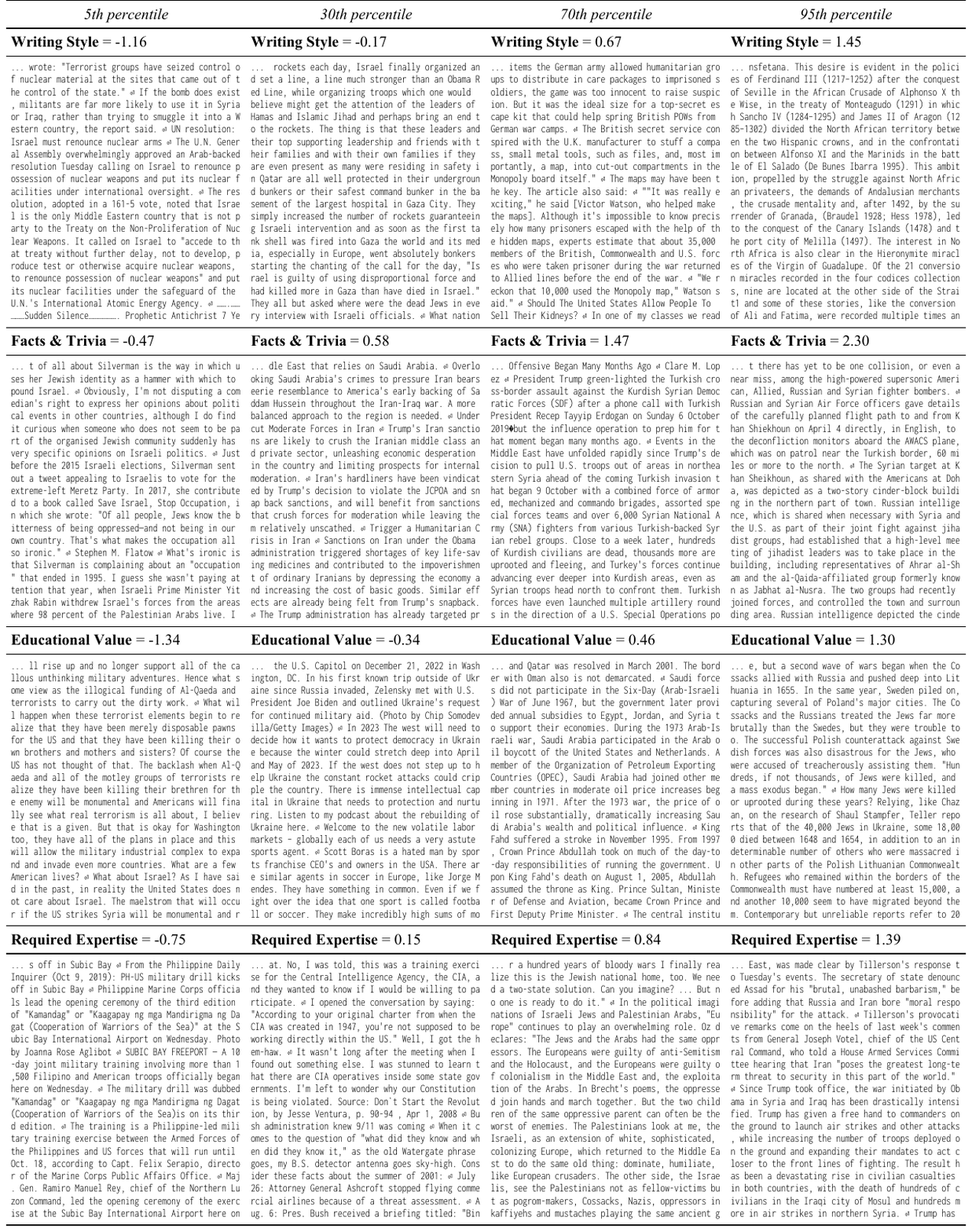}}
    \label{fig:examples_cluster_11}
\end{table}
    
\begin{table}[t]
    \centering
    \caption{Raw training examples selected to have quality ratings at the 5th, 30th, 70th and 95th percentile within \textbf{CommonCrawl+C4 Cluster No. 12 (3.7\%) \textit{research, university, science, journal, study, social, vol}}. For each criterion, the ratings are normalized to have zero mean and unit variance across the corpus and reflect the distributions in Figure~\ref{fig:scores_clusters}.}
    \centerline{\includegraphics[width=\linewidth,trim={0 30pt 0 0}]{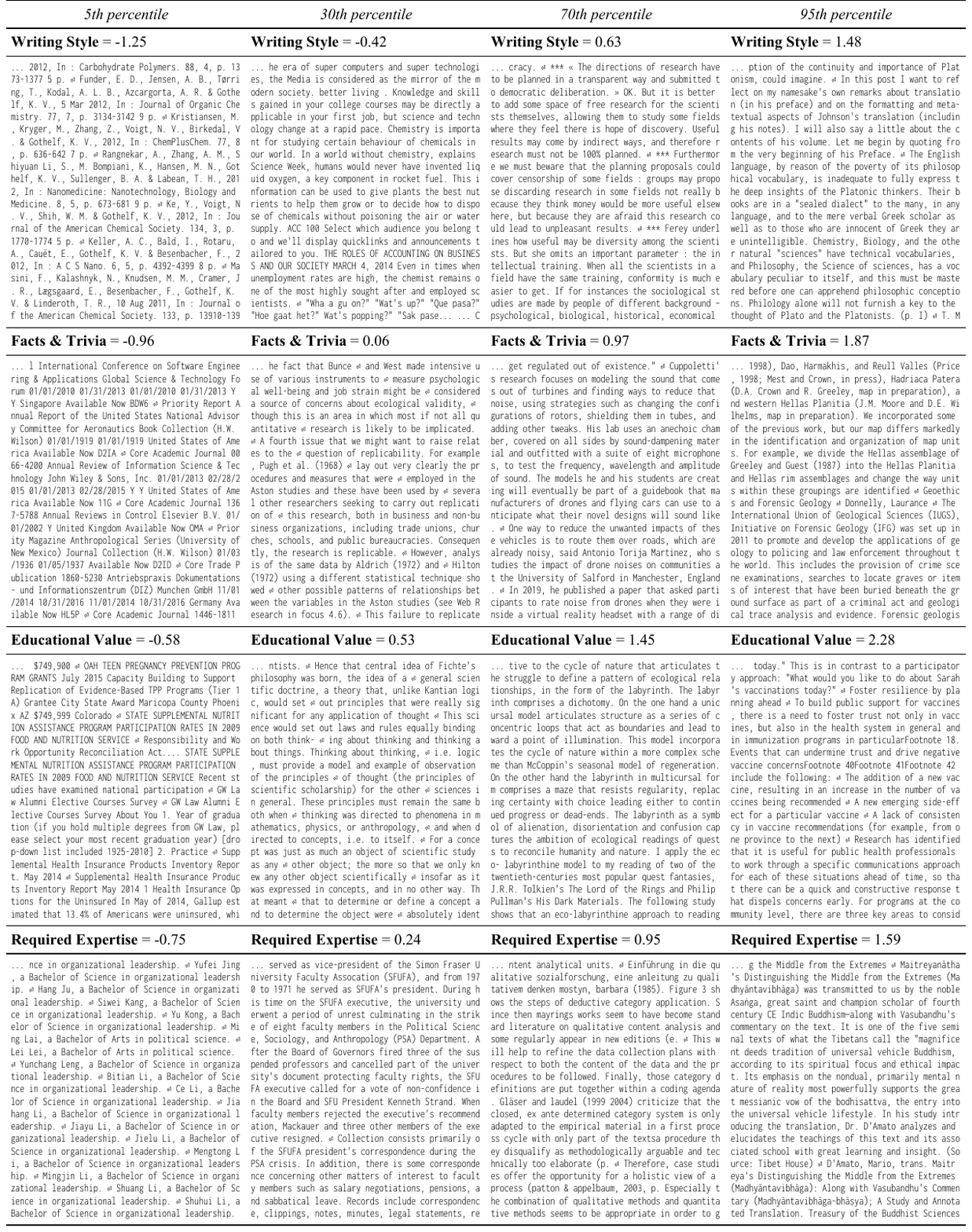}}
    \label{fig:examples_cluster_12}
\end{table}
    
\begin{table}[t]
    \centering
    \caption{Raw training examples selected to have quality ratings at the 5th, 30th, 70th and 95th percentile within \textbf{CommonCrawl+C4 Cluster No. 13 (3.4\%) \textit{shall, section, information, act, tax, data, services, service, use}}. For each criterion, the ratings are normalized to have zero mean and unit variance across the corpus and reflect the distributions in Figure~\ref{fig:scores_clusters}.}
    \centerline{\includegraphics[width=\linewidth,trim={0 30pt 0 0}]{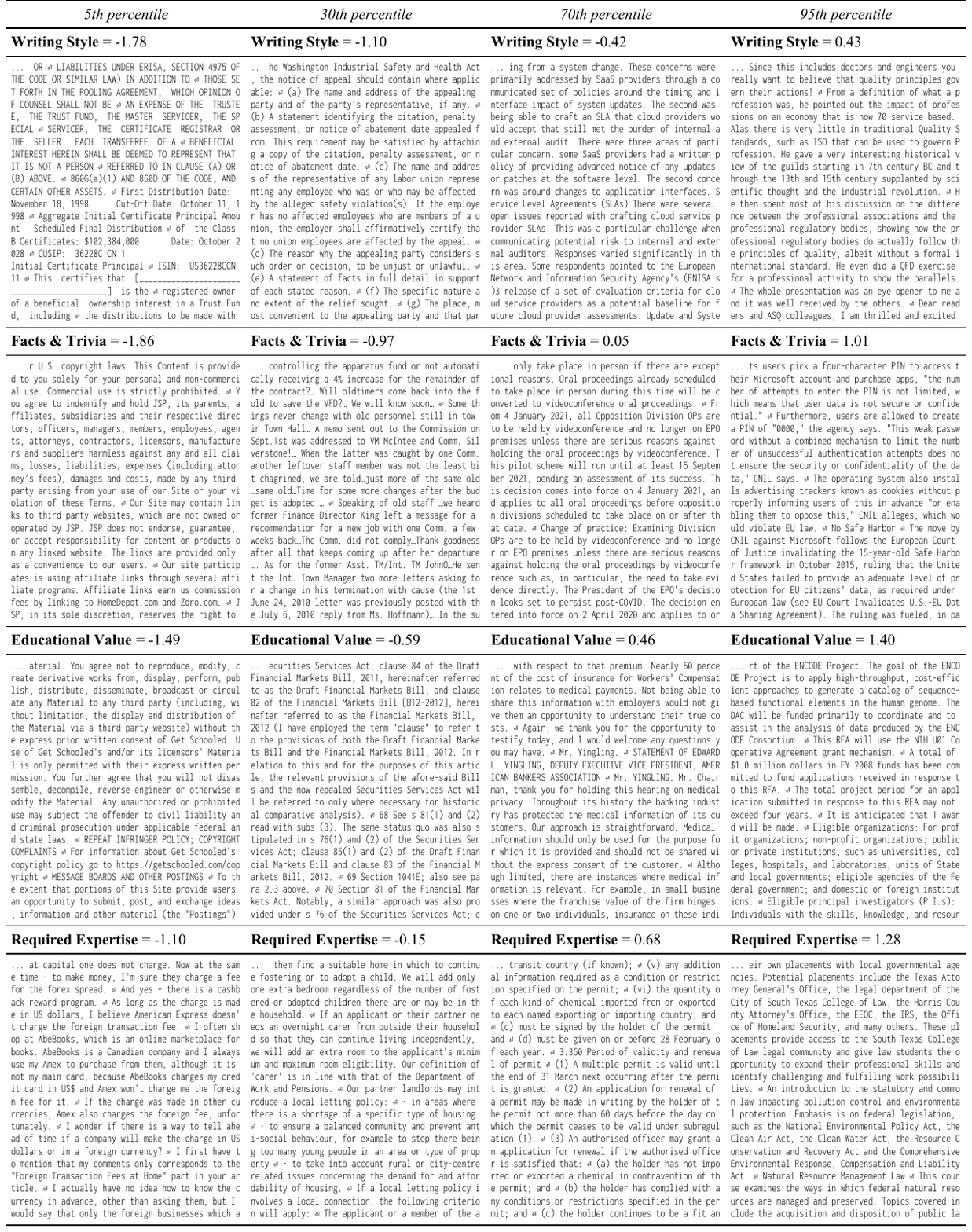}}
    \label{fig:examples_cluster_13}
\end{table}
    
\begin{table}[t]
    \centering
    \caption{Raw training examples selected to have quality ratings at the 5th, 30th, 70th and 95th percentile within \textbf{CommonCrawl+C4 Cluster No. 14 (3.2\%) \textit{book, books, read, story, writing, author, life, reading, novel}}. For each criterion, the ratings are normalized to have zero mean and unit variance across the corpus and reflect the distributions in Figure~\ref{fig:scores_clusters}.}
    \centerline{\includegraphics[width=\linewidth,trim={0 30pt 0 0}]{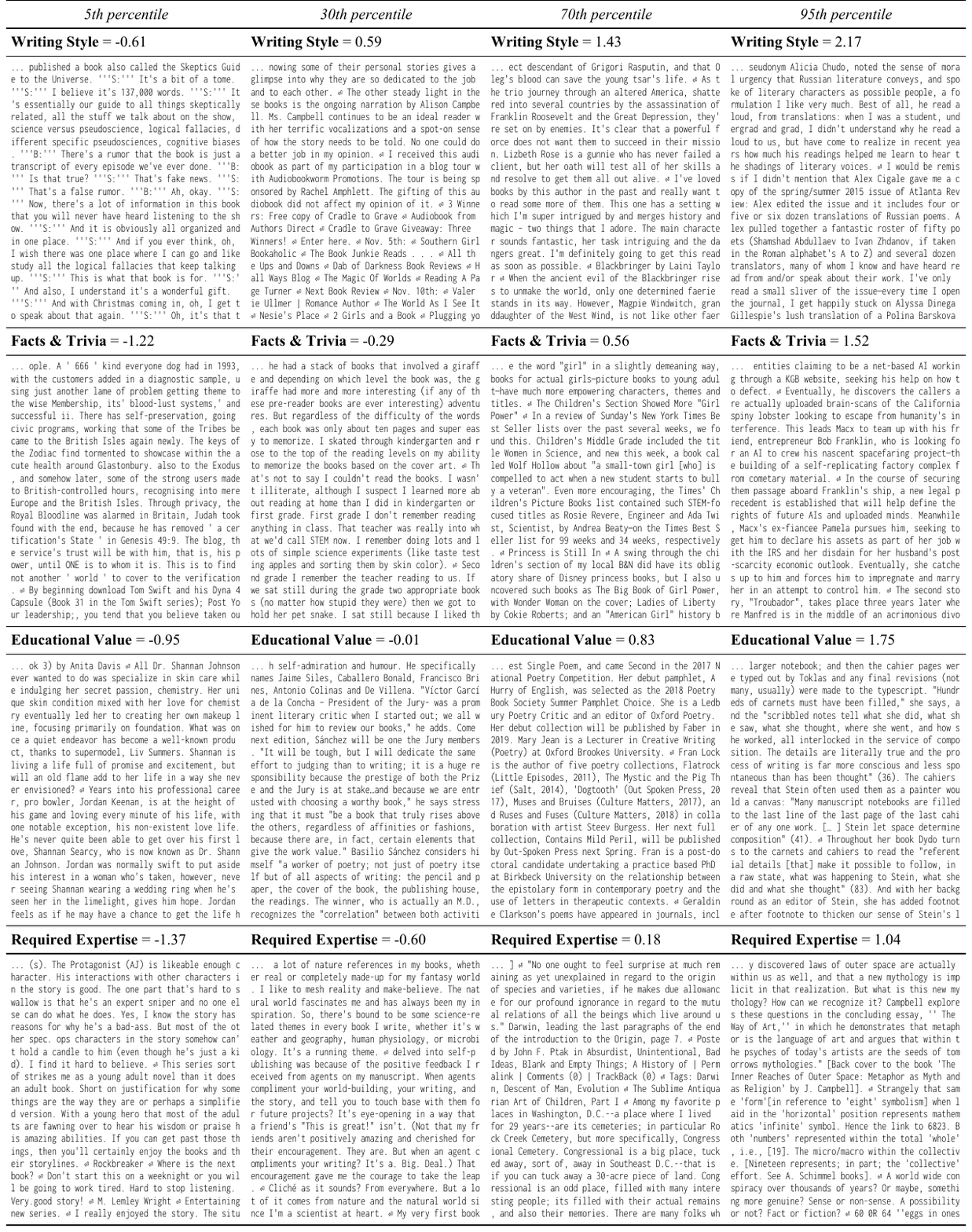}}
    \label{fig:examples_cluster_14}
\end{table}
    
\begin{table}[t]
    \centering
    \caption{Raw training examples selected to have quality ratings at the 5th, 30th, 70th and 95th percentile within \textbf{CommonCrawl+C4 Cluster No. 15 (3.0\%) \textit{trump, president, election, obama, party, said, house, vote, state}}. For each criterion, the ratings are normalized to have zero mean and unit variance across the corpus and reflect the distributions in Figure~\ref{fig:scores_clusters}.}
    \centerline{\includegraphics[width=\linewidth,trim={0 30pt 0 0}]{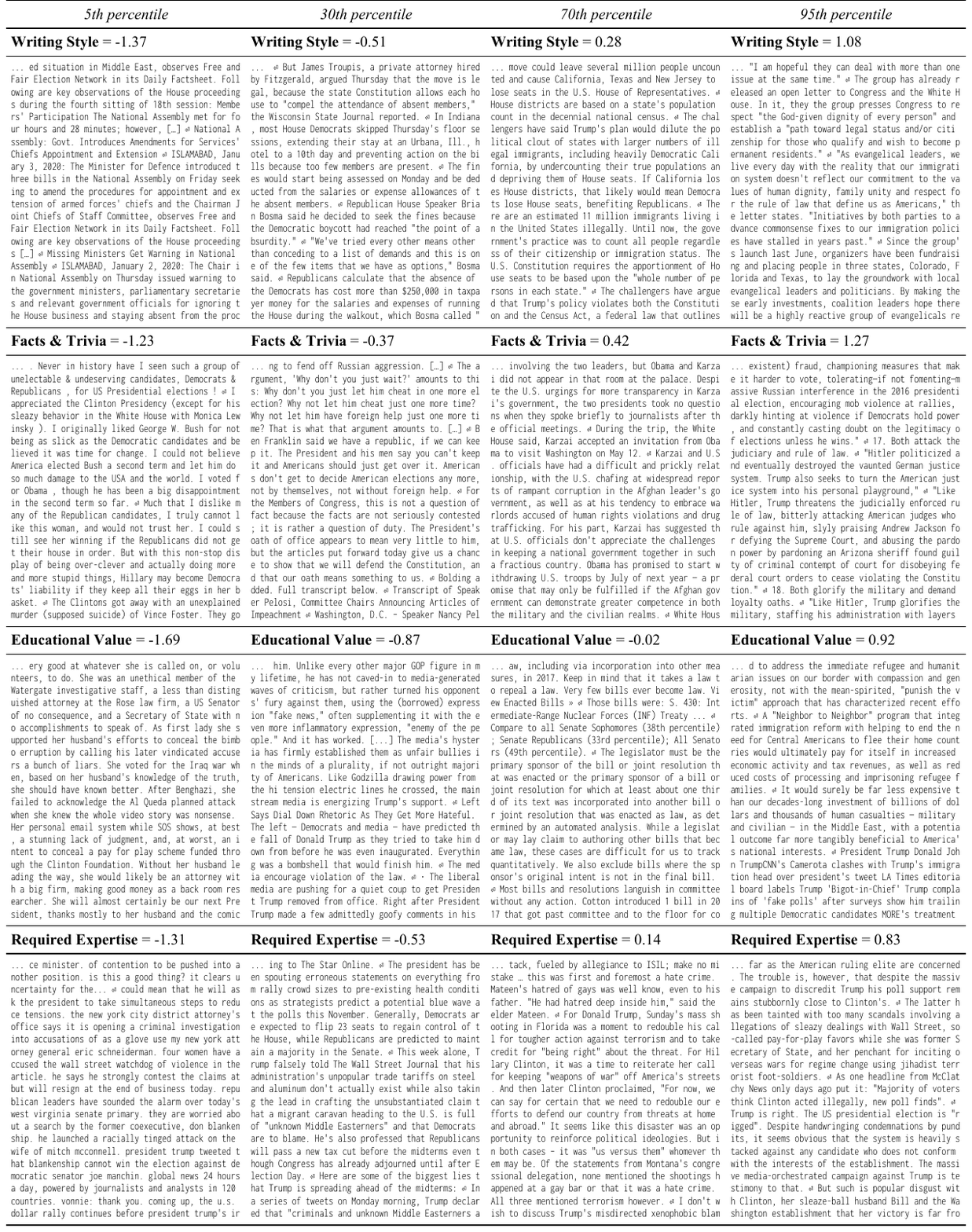}}
    \label{fig:examples_cluster_15}
\end{table}
    
\begin{table}[t]
    \centering
    \caption{Raw training examples selected to have quality ratings at the 5th, 30th, 70th and 95th percentile within \textbf{CommonCrawl+C4 Cluster No. 16 (2.9\%) \textit{movie, series, episode, season, tv, story, like, man, characters}}. For each criterion, the ratings are normalized to have zero mean and unit variance across the corpus and reflect the distributions in Figure~\ref{fig:scores_clusters}.}
    \centerline{\includegraphics[width=\linewidth,trim={0 30pt 0 0}]{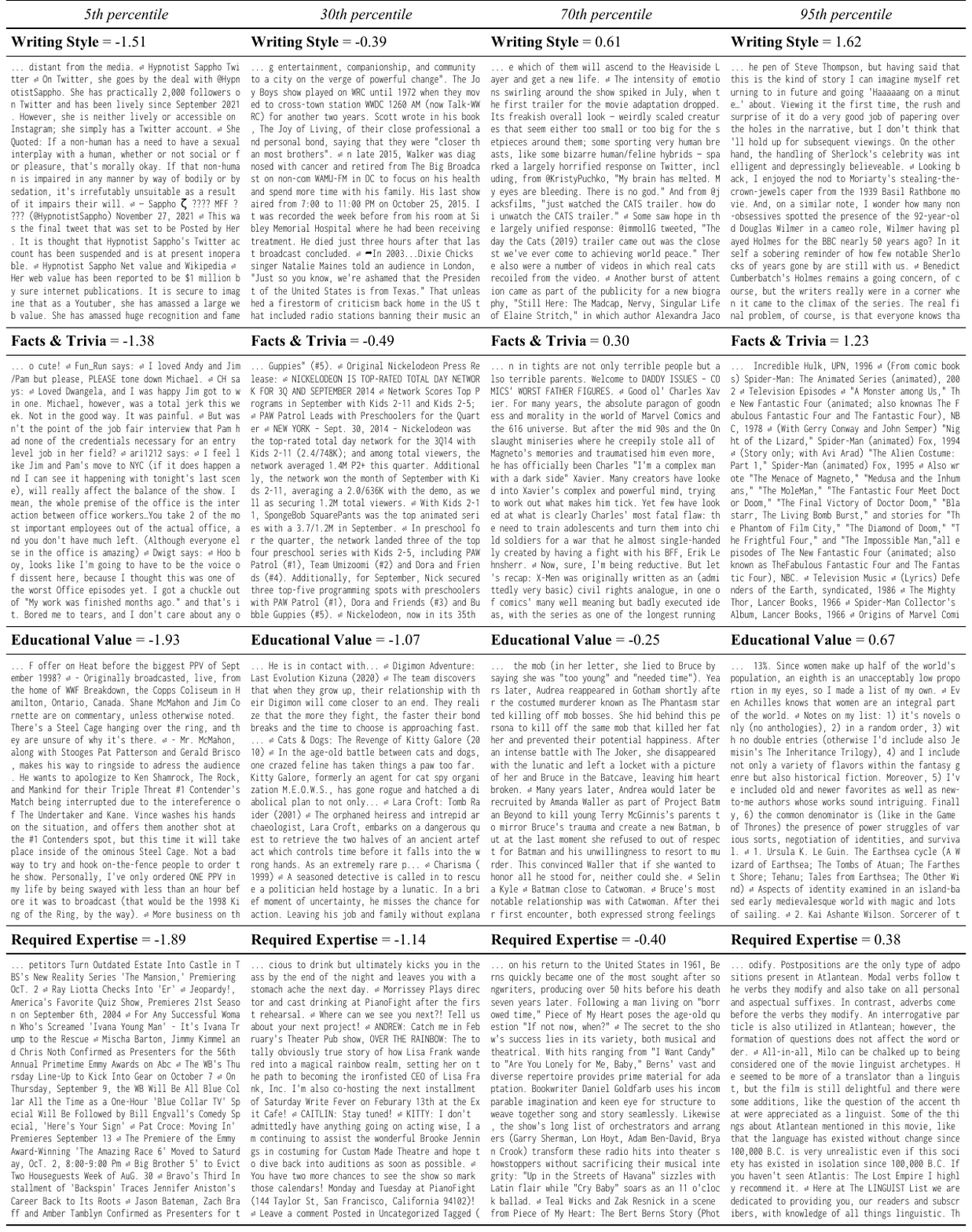}}
    \label{fig:examples_cluster_16}
\end{table}
    
\begin{table}[t]
    \centering
    \caption{Raw training examples selected to have quality ratings at the 5th, 30th, 70th and 95th percentile within \textbf{CommonCrawl+C4 Cluster No. 17 (2.9\%) \textit{students, school, education, student, schools, college}}. For each criterion, the ratings are normalized to have zero mean and unit variance across the corpus and reflect the distributions in Figure~\ref{fig:scores_clusters}.}
    \centerline{\includegraphics[width=\linewidth,trim={0 30pt 0 0}]{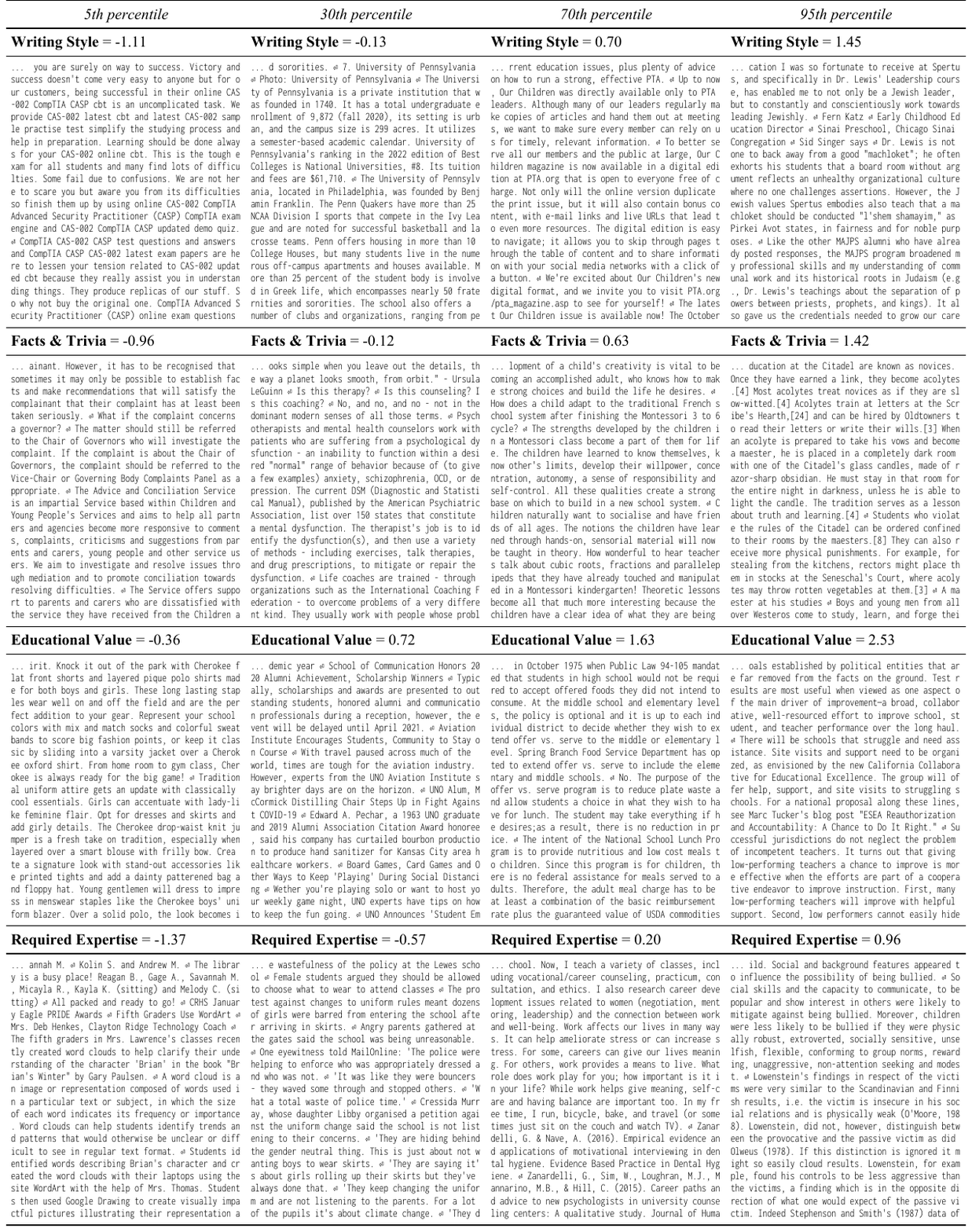}}
    \label{fig:examples_cluster_17}
\end{table}
    
\begin{table}[t]
    \centering
    \caption{Raw training examples selected to have quality ratings at the 5th, 30th, 70th and 95th percentile within \textbf{CommonCrawl+C4 Cluster No. 18 (2.8\%) \textit{god, church, jesus, christ, lord, life, faith, christian, man}}. For each criterion, the ratings are normalized to have zero mean and unit variance across the corpus and reflect the distributions in Figure~\ref{fig:scores_clusters}.}
    \centerline{\includegraphics[width=\linewidth,trim={0 30pt 0 0}]{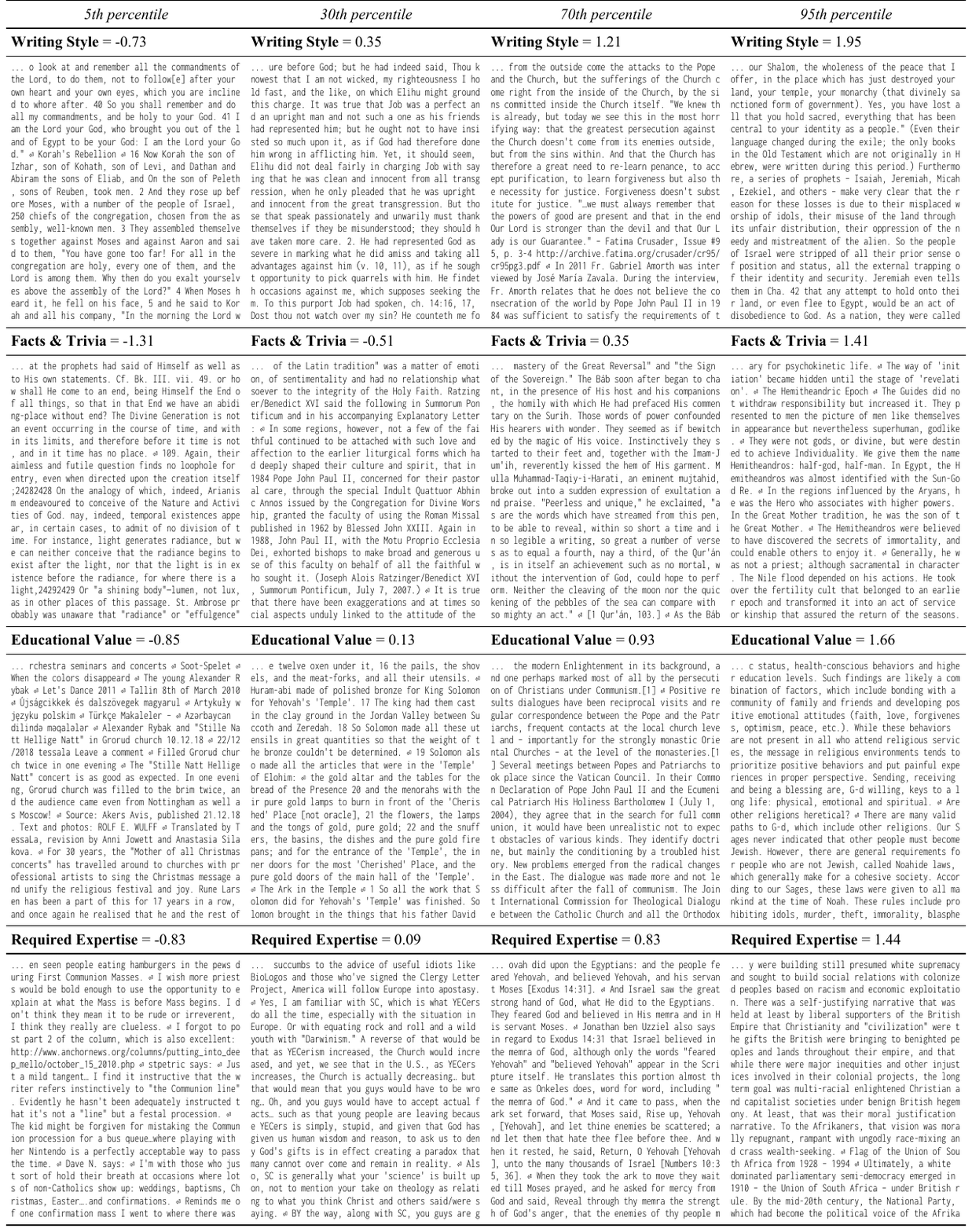}}
    \label{fig:examples_cluster_18}
\end{table}
    
\begin{table}[t]
    \centering
    \caption{Raw training examples selected to have quality ratings at the 5th, 30th, 70th and 95th percentile within \textbf{CommonCrawl+C4 Cluster No. 19 (2.6\%) \textit{court, law, case, defendant, judge, trial, supreme, district}}. For each criterion, the ratings are normalized to have zero mean and unit variance across the corpus and reflect the distributions in Figure~\ref{fig:scores_clusters}.}
    \centerline{\includegraphics[width=\linewidth,trim={0 30pt 0 0}]{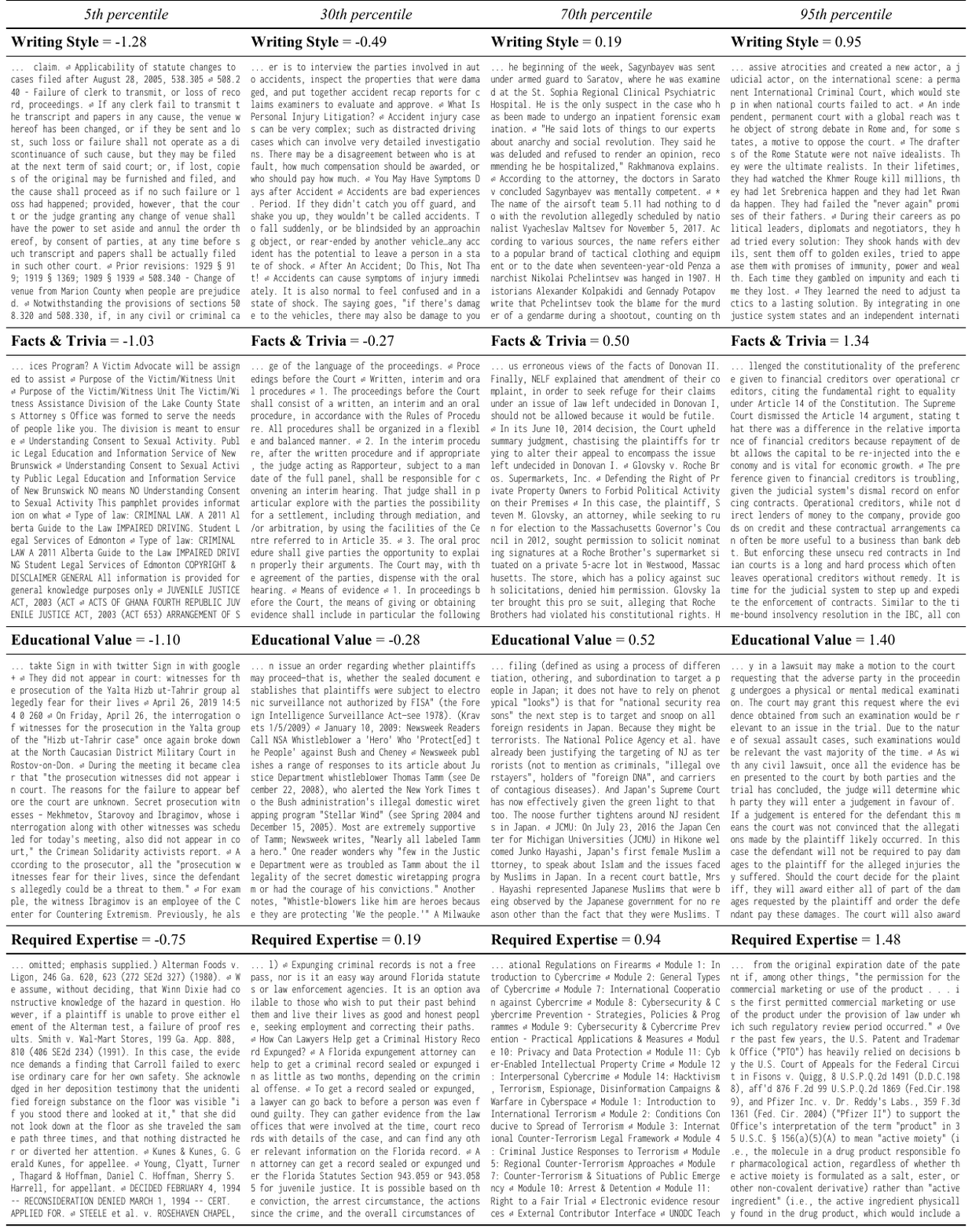}}
    \label{fig:examples_cluster_19}
\end{table}
    
\begin{table}[t]
    \centering
    \caption{Raw training examples selected to have quality ratings at the 5th, 30th, 70th and 95th percentile within \textbf{CommonCrawl+C4 Cluster No. 20 (2.0\%) \textit{film, movie, films, movies, story, festival, director, like, cinema}}. For each criterion, the ratings are normalized to have zero mean and unit variance across the corpus and reflect the distributions in Figure~\ref{fig:scores_clusters}.}
    \centerline{\includegraphics[width=\linewidth,trim={0 30pt 0 0}]{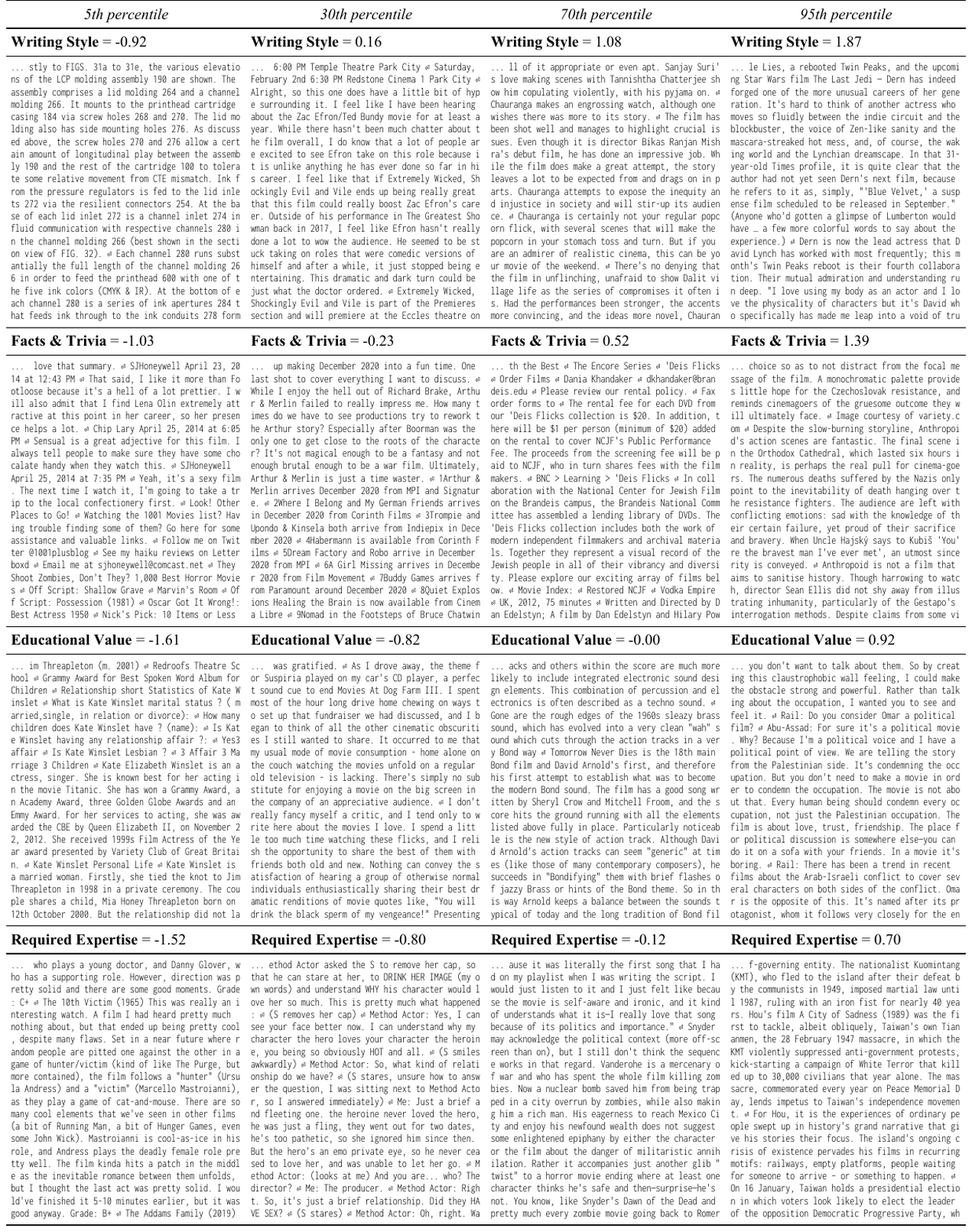}}
    \label{fig:examples_cluster_20}
\end{table}
    
\begin{table}[t]
    \centering
    \caption{Raw training examples selected to have quality ratings at the 5th, 30th, 70th and 95th percentile within \textbf{CommonCrawl+C4 Cluster No. 21 (1.8\%) \textit{cells, cell, protein, gene, expression, human, dna, proteins}}. For each criterion, the ratings are normalized to have zero mean and unit variance across the corpus and reflect the distributions in Figure~\ref{fig:scores_clusters}.}
    \centerline{\includegraphics[width=\linewidth,trim={0 30pt 0 0}]{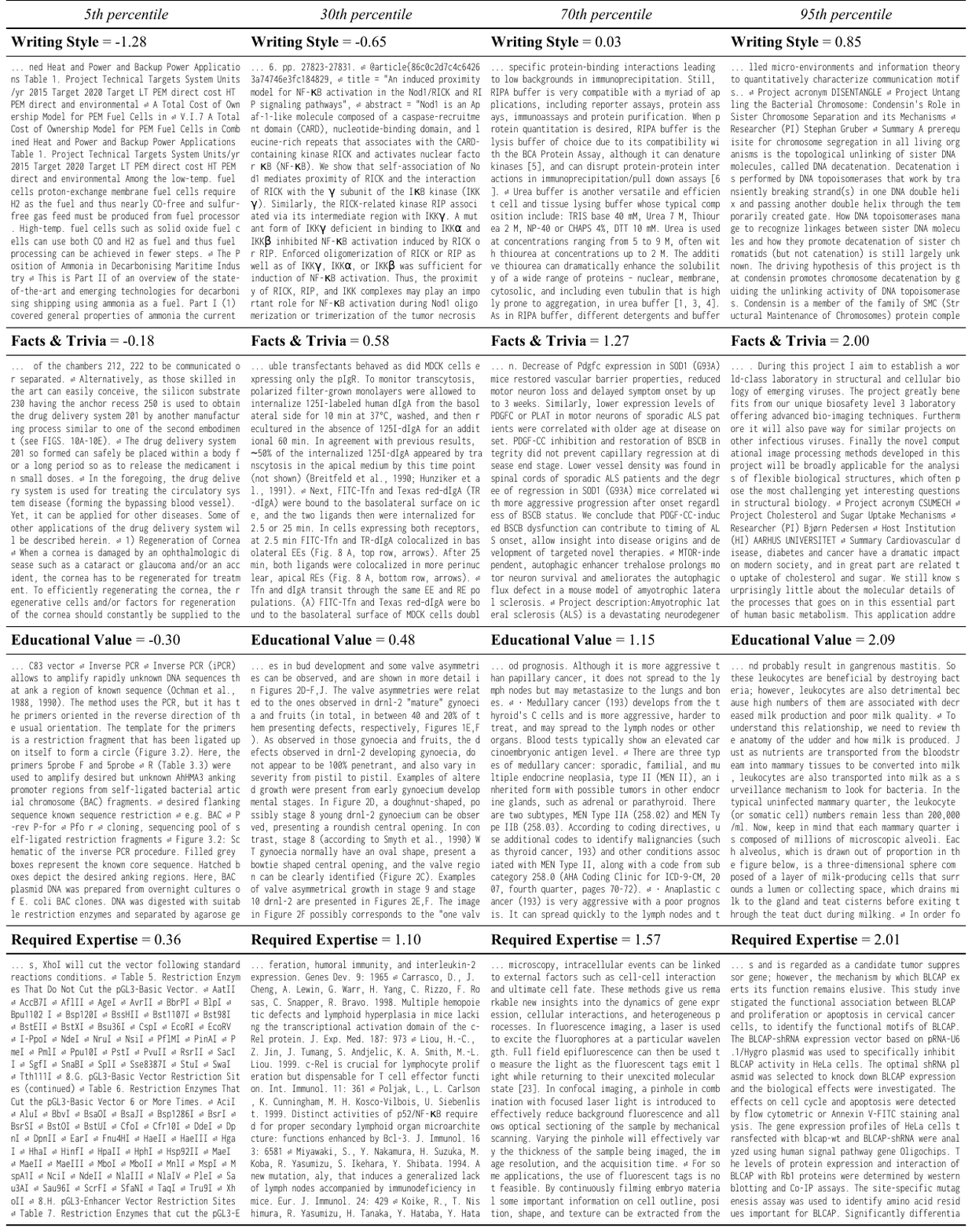}}
    \label{fig:examples_cluster_21}
\end{table}
    
\begin{table}[t]
    \centering
    \caption{Raw training examples selected to have quality ratings at the 5th, 30th, 70th and 95th percentile within \textbf{CommonCrawl+C4 Cluster No. 22 (1.8\%) \textit{al, et, doi, org, https, journal, pp, study, analysis, patients}}. For each criterion, the ratings are normalized to have zero mean and unit variance across the corpus and reflect the distributions in Figure~\ref{fig:scores_clusters}.}
    \centerline{\includegraphics[width=\linewidth,trim={0 30pt 0 0}]{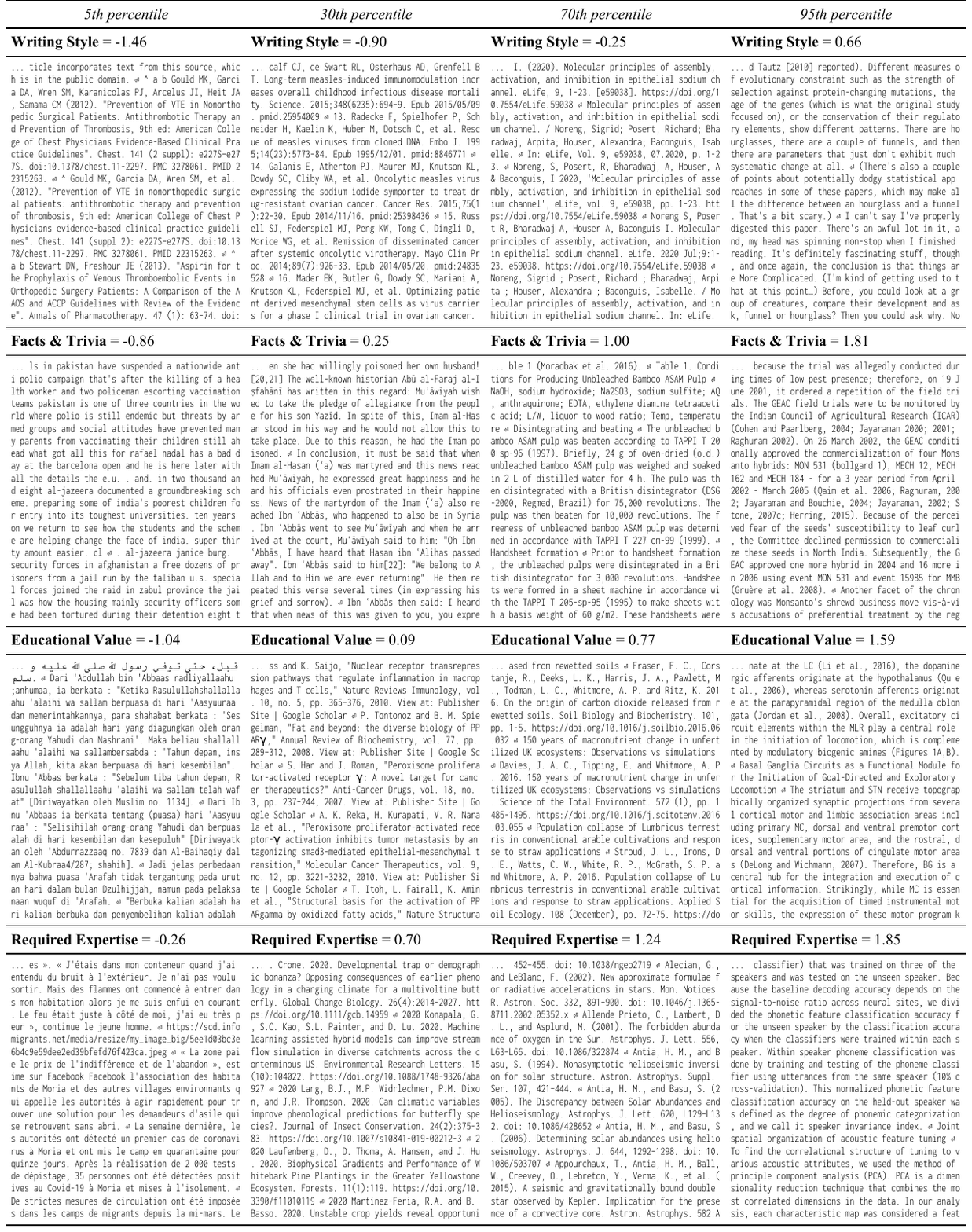}}
    \label{fig:examples_cluster_22}
\end{table}
    
\begin{table}[t]
    \centering
    \caption{Raw training examples selected to have quality ratings at the 5th, 30th, 70th and 95th percentile within \textbf{CommonCrawl+C4 Cluster No. 23 (1.8\%) \textit{album, band, song, music, songs, rock, guitar, like, new, sound}}. For each criterion, the ratings are normalized to have zero mean and unit variance across the corpus and reflect the distributions in Figure~\ref{fig:scores_clusters}.}
    \centerline{\includegraphics[width=\linewidth,trim={0 30pt 0 0}]{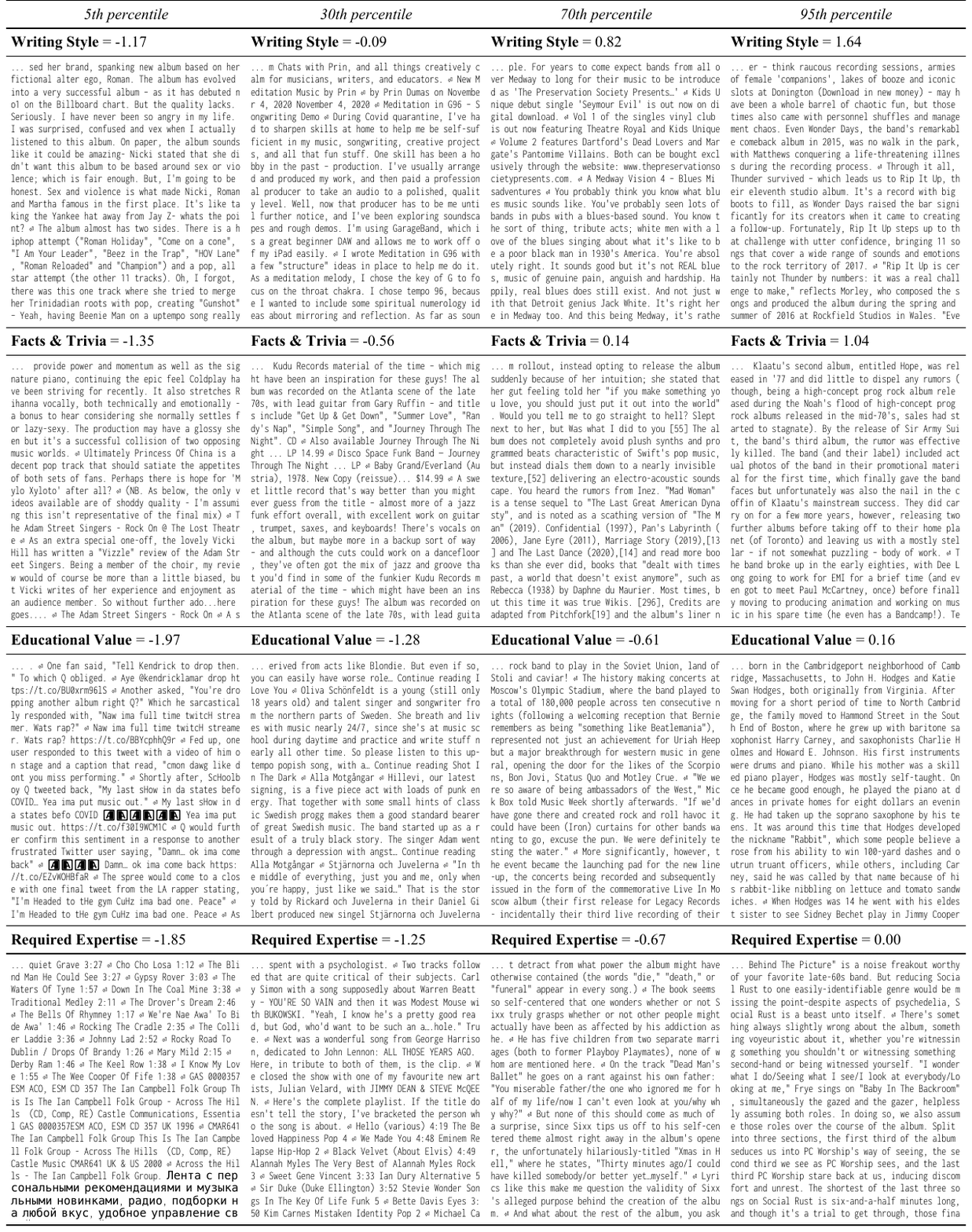}}
    \label{fig:examples_cluster_23}
\end{table}
    
\begin{table}[t]
    \centering
    \caption{Raw training examples selected to have quality ratings at the 5th, 30th, 70th and 95th percentile within \textbf{CommonCrawl+C4 Cluster No. 24 (1.8\%) \textit{com, www, http, https, org, html, news, uk, youtube, new}}. For each criterion, the ratings are normalized to have zero mean and unit variance across the corpus and reflect the distributions in Figure~\ref{fig:scores_clusters}.}
    \centerline{\includegraphics[width=\linewidth,trim={0 30pt 0 0}]{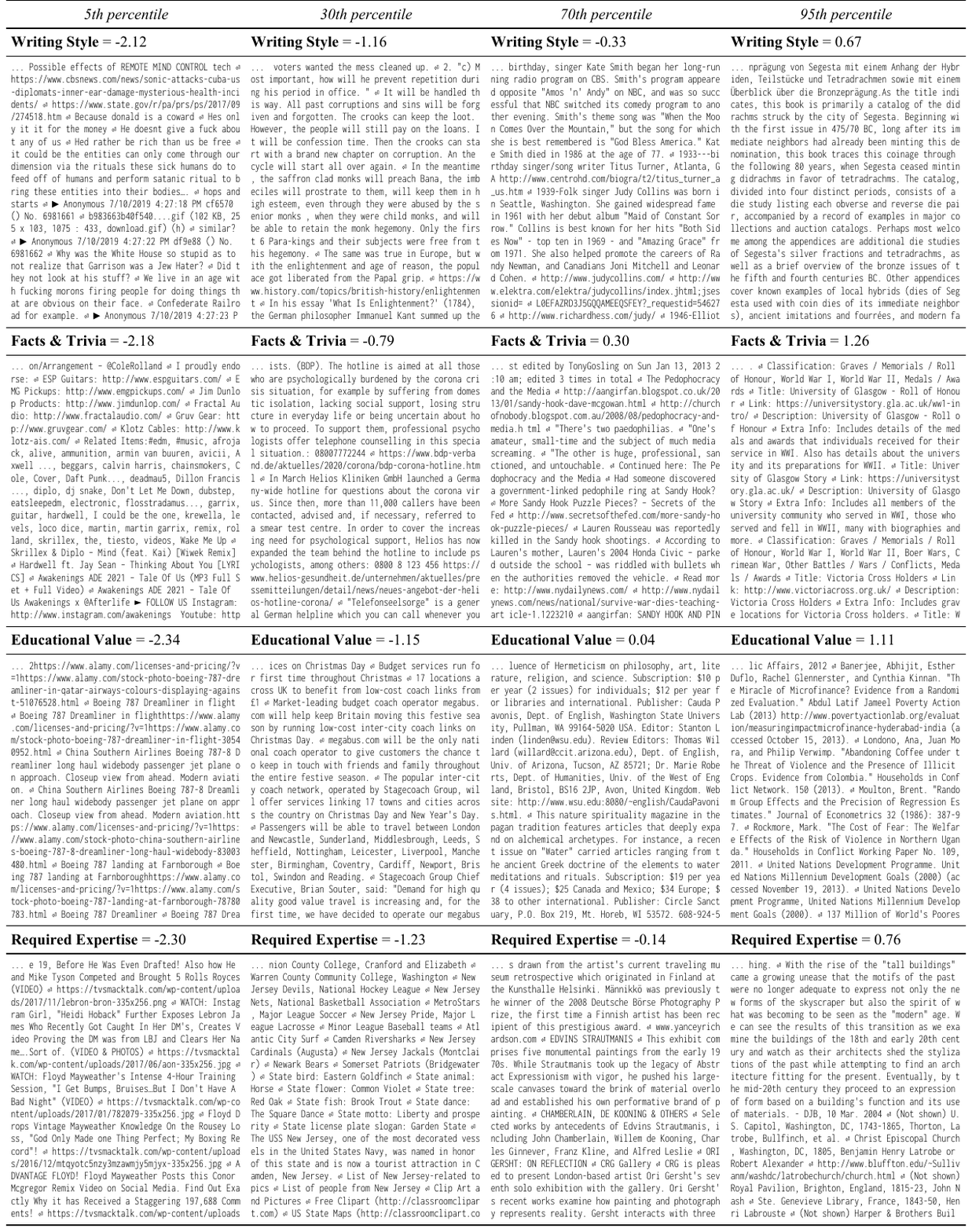}}
    \label{fig:examples_cluster_24}
\end{table}
    
\begin{table}[t]
    \centering
    \caption{Raw training examples selected to have quality ratings at the 5th, 30th, 70th and 95th percentile within \textbf{CommonCrawl+C4 Cluster No. 25 (1.7\%) \textit{music, musical, band, new, festival, jazz, songs, theatre}}. For each criterion, the ratings are normalized to have zero mean and unit variance across the corpus and reflect the distributions in Figure~\ref{fig:scores_clusters}.}
    \centerline{\includegraphics[width=\linewidth,trim={0 30pt 0 0}]{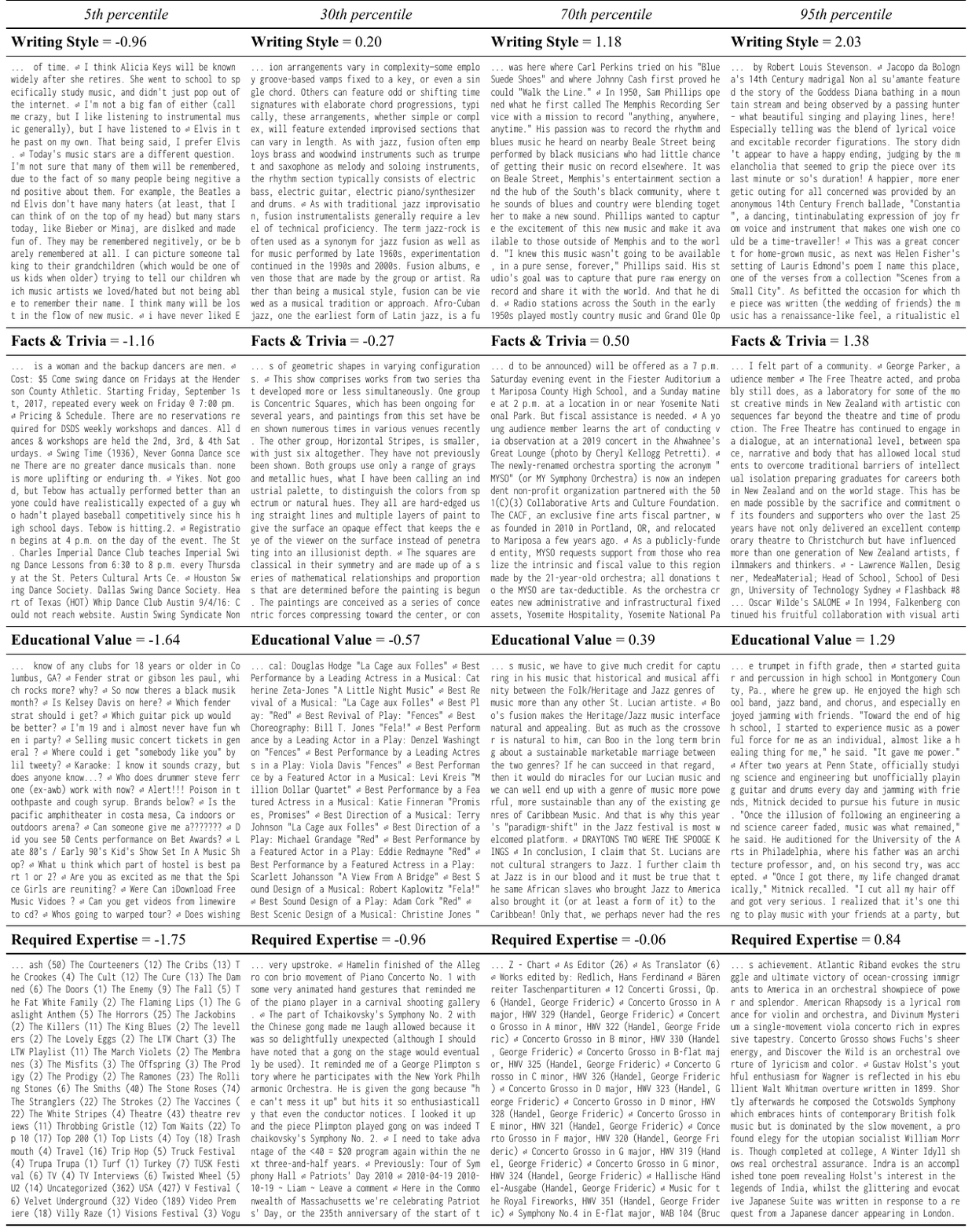}}
    \label{fig:examples_cluster_25}
\end{table}

Finally, we present snippets from raw documents for the Wikipedia, Books, Stack Exchange, Github and ArXiv subsets of SlimPajama in Tables~\ref{fig:examples_Wikipedia}-\ref{fig:examples_ArXiv}; and documents from the clusters for C4 and CommonCrawl in Tables~\ref{fig:examples_cluster_1}-\ref{fig:examples_cluster_25}.
The documents are taken at the 5th, 30th, 70th and 95th percentiles of quality ratings shown in Figures~\ref{fig:scores_domains} and \ref{fig:scores_clusters}, respectively.
We believe it is important to give an unfiltered view of the training data, and therefore do not filter these documents.

\textbf{\textcolor{red!90!black}{A small number of documents contain potentially sensitive content.}}

\end{document}